\documentclass{article}

     \PassOptionsToPackage{numbers, compress}{natbib}

\usepackage[preprint]{neurips_2020}

\usepackage[utf8]{inputenc} 
\usepackage[T1]{fontenc}    
\usepackage{hyperref}       
\usepackage{url}            
\usepackage{booktabs}       
\usepackage{amsfonts}       
\usepackage{nicefrac}       
\usepackage{microtype}      
\usepackage{enumitem}
\usepackage{kotex}

\usepackage{amsmath}        
\usepackage{multirow}       
\usepackage{multicol}       
\usepackage[ruled,vlined]{algorithm2e}   
\usepackage[dvipsnames]{xcolor} 
\usepackage{graphicx}       
\usepackage{makecell}       
\usepackage{tabularx}       
\usepackage{adjustbox}      
\usepackage{imakeidx}       
\makeindex[columns=2, intoc]

\newcommand{\wikipedia}{WIKIPEDIA}
\newcommand{\yna}{YNA}
\newcommand{\wikitree}{WIKITREE}
\newcommand{\policy}{POLICY}
\newcommand{\nsmc}{NSMC}
\newcommand{\airbnb}{AIRBNB}
\newcommand{\wikinews}{WIKINEWS}
\newcommand{\parakqc}{PARAKQC}
\newcommand{\hankyung}{The Korea Economy Daily}
\newcommand{\acrofan}{ACROFAN}

\newcommand{\modu}{MODU}
\newcommand{\CommonCrawl}{CC-100-Kor}
\newcommand{\namuwiki}{NAMUWIKI}
\newcommand{\newscrawl}{NEWSCRAWL}
\newcommand{\petition}{PETITION}

\newcommand{\ner}{NER}
\newcommand{\posdp}{DP}
\newcommand{\tc}{TC}
\newcommand{\sts}{STS}
\newcommand{\nli}{NLI}
\newcommand{\re}{RE}
\newcommand{\mrc}{MRC}
\newcommand{\dst}{DST}
\newcommand{\kluener}{KLUE-NER}
\newcommand{\klueposdp}{KLUE-DP}
\newcommand{\ynat}{YNAT}
\newcommand{\kluests}{KLUE-STS}
\newcommand{\kluenli}{KLUE-NLI}
\newcommand{\kluere}{KLUE-RE}
\newcommand{\kluemrc}{KLUE-MRC}
\newcommand{\wizard}{WoS}
\newcommand{\kluebert}{KLUE-BERT}
\newcommand{\klueroberta}{KLUE-RoBERTa}

\newcommand{\bertbase}{$\text{KLUE-BERT}_{\text{BASE}}$}
\newcommand{\mbertbase}{$\text{mBERT}_{\text{BASE}}$}
\newcommand{\xlmrbase}{$\text{XLM-R}_{\text{BASE}}$}
\newcommand{\xlmrlarge}{$\text{XLM-R}_{\text{LARGE}}$}
\newcommand{\krbertbase}{$\text{KR-BERT}_{\text{BASE}}$}
\newcommand{\koelectrabase}{$\text{KoELECTRA}_{\text{BASE}}$}
\newcommand{\robertasmall}{$\text{KLUE-RoBERTa}_{\text{SMALL}}$}
\newcommand{\robertabase}{$\text{KLUE-RoBERTa}_{\text{BASE}}$}
\newcommand{\robertalarge}{$\text{KLUE-RoBERTa}_{\text{LARGE}}$}

\title{KLUE: Korean Language Understanding Evaluation}

\author{

Sungjoon Park\textsuperscript{*}\\ 
    \small{Upstage, KAIST}\\ 
    \scriptsize{\texttt{sungjoon.park@kaist.ac.kr}}\And

Jihyung Moon\textsuperscript{*}\\ 
    \small{Upstage}\\  
    \scriptsize{\texttt{jihyung.moon@upstage.ai}}\And

Sungdong Kim\textsuperscript{*}\\ 
    \small{NAVER AI Lab}\\ 
    \scriptsize{\texttt{sungdong.kim@navercorp.com}} \And

Won Ik Cho\textsuperscript{*}\\ 
    \footnotesize{Seoul National University}\\ 
    \scriptsize{\texttt{tsatsuki@snu.ac.kr}} \And

Jiyoon Han\textsuperscript{$\dagger$}\\
    \small{Yonsei University}\\
    \scriptsize{\texttt{clinamen35@yonsei.ac.kr}} \And

Jangwon Park\\
    \scriptsize{\texttt{jangwon.pk@gmail.com}} \And

Chisung Song\\
    \scriptsize{\texttt{daydrilling@gmail.com}} \And

Junseong Kim\\
    \small{Scatter Lab}\\
    \tiny{\texttt{junseong.kim@scatterlab.co.kr}} \And

Youngsook Song\\	
    \small{KyungHee University} \\ 
    \scriptsize{\texttt{youngsoksong@khu.ac.kr}} \And

Taehwan Oh\textsuperscript{$\dagger$}\\	
    \small{Yonsei University} \\ 
    \scriptsize{\texttt{ghksl0604@yonsei.ac.kr}} \And

Joohong Lee\\
    \small{Scatter Lab}\\ 
    \scriptsize{\texttt{joohong@scatterlab.co.kr}} \And

Juhyun Oh\textsuperscript{$\dagger$}\\
    \footnotesize{Seoul National University} \\ 
    \scriptsize{\texttt{411juhyun@snu.ac.kr}} \And

Sungwon Lyu\\
    \small{Kakao Enterprise} \\
    \tiny{\texttt{james.ryu@kakaoenterprise.com}} \And

Younghoon Jeong\\
    \small{Sogang University} \\ 
    \scriptsize{\texttt{boychaboy@sogang.ac.kr}} \And

Inkwon Lee\\
    \small{Sogang University} \\ 
    \scriptsize{\texttt{md98765@naver.com}} \And

Sangwoo Seo\\
    \small{Scatter Lab} \\ 
    \tiny{\texttt{sangwoo@scatterlab.co.kr}} \And

Dongjun Lee\\
    \tiny{\texttt{humanbrain.djlee@gmail.com}} \And

Hyunwoo Kim\\
    \footnotesize{Seoul National University} \\ 
    \scriptsize{\texttt{hyunw.kim@vl.snu.ac.kr}} \And

Myeonghwa Lee\\
    \small{KAIST} \\ 
    \scriptsize{\texttt{myeon9h@kaist.ac.kr}} \And

Seongbo Jang\\
    \small{Scatter Lab} \\ 
    \tiny{\texttt{seongbo@scatterlab.co.kr}} \And

Seungwon Do\\
    \scriptsize{\texttt{seungwon.do1@gmail.com}} \And

Sunkyoung Kim\\
    KAIST \\ 
    \scriptsize{\texttt{sunkyoung@kaist.ac.kr}} \And

Kyungtae Lim\\
    \footnotesize{Hanbat National University} \\ 
    \scriptsize{\texttt{ktlim@hanbat.ac.kr}} \And

Jongwon Lee\\
    \scriptsize{\texttt{mybizzer@gmail.com}} \And

Kyumin Park\\
    \small{KAIST} \\ 
    \scriptsize{\texttt{pkm9403@kaist.ac.kr}} \And

Jamin Shin\\
    \small{Riiid AI Research} \\ 
    \scriptsize{\texttt{jshin49@gmail.com}} \And

Seonghyun Kim \\
    \scriptsize{\texttt{bananaband657@gmail.com}} \And

Lucy Park\\
    \small{Upstage} \\ 
    \scriptsize{\texttt{lucy@upstage.ai}} \And

Alice Oh\textsuperscript{**}\\	
    \small{KAIST} \\ 
    \scriptsize{\texttt{alice.oh@kaist.edu}} \And

Jung-Woo Ha\textsuperscript{**}\\
    \small{NAVER AI Lab} \\ 
    \scriptsize{\texttt{jungwoo.ha@navercorp.com}} \And

Kyunghyun Cho\textsuperscript{**}\\
    \small{New York University} \\ 
    \scriptsize{\texttt{kyunghyun.cho@nyu.edu}}

}

\makeatletter
\def\blfootnotestar{\gdef\@thefnmark{*}\@footnotetext}
\def\blfootnotedoublestar{\gdef\@thefnmark{**}\@footnotetext}
\def\blfootnotedagger{\gdef\@thefnmark{$\dagger$}\@footnotetext}
\makeatother

\begin{document}

\maketitle

\blfootnotestar{Equal Contribution. A description of each author's contribution is available at the end of paper.}
\blfootnotedoublestar{Corresponding Authors.}
\blfootnotedagger{Work done at Upstage.}

\begin{abstract}
We introduce Korean Language Understanding Evaluation (KLUE) benchmark. KLUE is a collection of 8 Korean natural language understanding (NLU) tasks, including Topic Classification, Semantic Textual Similarity, Natural Language Inference, Named Entity Recognition, Relation Extraction, Dependency Parsing, Machine Reading Comprehension, and Dialogue State Tracking. We build all of the tasks from scratch from diverse source corpora while respecting copyrights, to ensure accessibility for anyone without any restrictions. With ethical considerations in mind, we carefully design annotation protocols. Along with the benchmark tasks and data, we provide suitable evaluation metrics and fine-tuning recipes for pretrained language models for each task. We furthermore release the pretrained language models (PLM), \kluebert{} and \klueroberta{}, to help reproducing baseline models on KLUE and thereby facilitate future research.
We make a few interesting observations from the preliminary experiments using the proposed KLUE benchmark suite, already demonstrating the usefulness of this new benchmark suite. First, we find \robertalarge{} outperforms other baselines, including multilingual PLMs and existing open-source Korean PLMs. Second, we see minimal degradation in performance even when we replace personally identifiable information from the pretraining corpus, suggesting that privacy and NLU capability are not at odds with each other. Lastly, we find that using BPE tokenization in combination with morpheme-level pre-tokenization is effective in tasks involving morpheme-level tagging, detection and generation. In addition to accelerating Korean NLP research, our comprehensive documentation on creating KLUE will facilitate creating similar resources for other languages in the future. KLUE is available at \url{https://klue-benchmark.com/}.
\end{abstract}

\clearpage
\setcounter{tocdepth}{3}
\tableofcontents
\clearpage

\section{Introduction}
A major factor behind recent success of pretrained language models, such as BERT~\cite{devlin2019bert} and its variants~\cite{liu2019roberta,clark2020electra,he2020deberta} as well as GPT-3~\cite{radford2019language} and its variants~\cite{2020t5,lewis-etal-2020-bart,NEURIPS2020_1457c0d6}, has been the availability of well-designed benchmark suites for evaluating their effectiveness in natural language understanding (NLU). GLUE \cite{wang2018glue} and SuperGLUE \cite{wang2019superglue} are representative examples of such suites and were designed to evaluate diverse aspects of NLU, including syntax, semantics and pragmatics. The research community has embraced GLUE and SuperGLUE, and has made rapid progress in developing better model architectures as well as learning algorithms for NLU.  

The success of GLUE and SuperGLUE has sparked interest in building such a standardized benchmark suite for other languages, in order to better measure the progress in NLU in languages beyond English. Such efforts have been pursued along two directions. First, various groups in the world have independently created language-specific benchmark suites; a Chinese version of GLUE (CLUE \cite{xu2020clue}), a French version of GLUE (FLUE \cite{le2020flaubert}), an Indonesian variant \cite{wilie2020indonlu}, an Indic version \cite{kakwani2020indicnlpsuite} and a Russian variant of SuperGLUE \cite{shavrina2020russiansuperglue}. On the other hand, some have relied on both machine and human translation of existing benchmark suites for building multilingual version of the benchmark suites which were often created initially in English. These include for instance XGLUE \cite{liang2020xglue} and XTREME \cite{hu2020xtreme}. Although the latter approach scales much better than the former does, the latter often fails to capture societal aspects of NLU and also introduces various artifacts arising from translation.

To this end, we build a new benchmark suite for evaluating NLU in Korean which is the 13-th most used language in the world according to \cite{ethno2021} but lacks a unified benchmark suite for NLU. Instead of starting from existing benchmark tasks or corpora, we build this benchmark suite from ground up by determining and collecting base corpora, identifying a set of benchmark tasks, designing appropriate annotation protocols and finally validating collected annotation. This allows us to preemptively address and avoid properties that may have undesirable consequences, such as copyright infringement, annotation artifacts, social biases and privacy violations.

In the rest of this section, we summarize a series of decisions and principles that went behind creating KLUE.

\subsection{Summary}
In designing the Korean Language Understanding Evaluation (KLUE)\index{KLUE \\(Korean Language Understanding Evaluation)}, we aim to make KLUE; 1) cover diverse tasks and corpora, 2) accessible to everyone without any restriction, 3) include accurate and unambiguous annotations, 4) mitigate AI ethical issues. KLUE is safe to use for both building and evaluating systems, because KLUE has proactively addressed potential {\it ethical} issues. Here, we describe more in detail how these principles have guided creating KLUE from task selection, corpus selection, annotation protocols, determining evaluation metrics to baseline construction. 

\paragraph{Design Principles} 

First, let us describe each design principle in detail:

\begin{itemize}[leftmargin=*]

    \item \textit{Covering diverse tasks and corpora}: 
    To cover diverse aspects of language understanding, 
    we choose eight tasks that cover diverse domain, including news, encyclopedia, user review, smart home queries and task-oriented dialogue, and diverse style, both formal and colloquial.
    
    \item \textit{Accessible to everyone without any restriction}: 
    It is critical for a benchmark suite to be accessible by everyone for it to serve as a true guideline in evaluating and improving NLU systems. We thus use only corpora and resources that can be freely copied, redistributed, remixed and transformed for the purpose of benchmarking NLU systems.
    
    \item \textit{Obtaining accurate and unambiguous annotations}: 
    Ambiguity in benchmark tasks leads to ambiguity in evaluation, which often leads to the discrepancy between the quality of an NLU system measured by the benchmark and its true quality. In order to minimize such discrepancy,  
    we carefully design annotation guidelines of all tasks and improve them over multiple iterations, to avoid accurate annotations.

    \item \textit{Mitigating AI ethical issues}: 
    It has been repeatedly observed that large-scale language models can and often do amplify social biases embedded in text used to train them \cite{nangia-etal-2020-crows}. In order to disincentivize such behaviors, we proactively remove examples, from both unlabeled and labeled corpora, that reflect social biases, contain toxic content and have personally identifiable information (PII), both manually and automatically. Social biases are defined as  overgeneralized judgment on certain individuals or group based on social attributes (e.g., gender, ethnicity, religion). Toxic contents include insults, sexual harassment and offensive expressions. 

\end{itemize}


\paragraph{Diverse Task Selection} 

We carefully choose the following eight NLU tasks with two goals; 1) to cover as diverse aspects of NLU in Korean, and 2) to minimize redundancy among the tasks. See Table~\ref{tab:task-overview} for their formats, evaluation granularity and other properties:
\label{sec:task-selection}
 
\begin{itemize}[leftmargin=*]
    \item Topic Classification (\tc): classify a single sentence into a single class.
    \item Semantic Textual Similarity (\sts): judge the semantic similarity between two sentences.
    \item Natural Language Inference (\nli): classify whether the first sentence entails the second one.
    \item Named Entity Recognition (\ner): extract entities from a sentence.
    \item Relation Extraction (\re): predict the relationship between two entities within a sentence.
    \item Dependency Parsing (\posdp): predict the syntactic structure of a sentence.
    \item Machine Reading Comprehension (\mrc): identify an answer span within a paragraph given a question.
    \item Dialogue State Tracking (\dst): track the state of a goal-oriented dialogue.
\end{itemize}

\paragraph{Source Corpora Collection} 
We have actively sought corpora that are accessible, cover diverse domains and topics and are written in modern Korean. This active search has ended up with the following ten source corpora from which we derive task-specific corpora. These ten sources are released under CC BY(-SA) license or not considered as copyrighted work, permitting 1) derivative work, 2) redistribution, and 3) commercial use:

\begin{itemize}[leftmargin=*]
    \item News Headlines from Yonhap News Agency
    \item Wikipedia
    \item Wikinews
    \item Wikitree
    \item Policy News
    \item ParaKQC
    \item Airbnb Reviews
    \item NAVER Sentiment Movie Corpus
    \item The Korea Economics Daily News
    \item Acrofan News
\end{itemize}

Before sending a subset for annotation, we filter them to remove noisy, toxic or socially biased content, as well as PII. This is done automatically using predefined rules and machine learning models.

\begin{table}[t!]
\caption{Task Overview}
\label{tab:task-overview}
\centering
\begin{adjustbox}{width=\textwidth}
\begin{tabular}{@{}llllccll@{}}
\toprule
\textbf{Name} &
\textbf{Type} &
\textbf{Format} &
  \begin{tabular}[c]{@{}l@{}}\textbf{Eval.}\\ \textbf{Metric}\end{tabular} &
  \begin{tabular}[c]{@{}l@{}}\textbf{\#} \\ \textbf{Class}\end{tabular} &
  \begin{tabular}[c]{@{}l@{}}
    \textbf{\{|Train|,} \\  
    \textbf{~~|Dev|,}\\
    \textbf{~~|Test|\}}
    \end{tabular} &
  \textbf{Source} &
  \textbf{Style} \\ \toprule
  \begin{tabular}[c]{@{}l@{}}KLUE-TC \\ (\ynat{}) \end{tabular} &
  \begin{tabular}[c]{@{}l@{}}Topic \\ Classification\end{tabular} &
  \begin{tabular}[c]{@{}l@{}}Single Sentence \\ Classification\end{tabular} &
  \begin{tabular}[c]{@{}l@{}} Macro F1 \end{tabular} &
  7 &
  \begin{tabular}[c]{@{}l@{}} ~~45k, \\  ~~~~9k, \\  ~~~~9k\end{tabular} &
  \begin{tabular}[c]{@{}l@{}}News \\ (Headline)\end{tabular} &
  Formal \\ \midrule
KLUE-STS &
  \begin{tabular}[c]{@{}l@{}}Semantic \\ Textual \\ Similarity\end{tabular} &
  \begin{tabular}[c]{@{}l@{}}Sentence \\ Pair \\ Regression\end{tabular} &
  \begin{tabular}[c]{@{}l@{}}Pearson's $r$, \\ F1\end{tabular} &
  \begin{tabular}[c]{@{}c@{}}$[0,5]$ \\ 2 \end{tabular} &
  \begin{tabular}[c]{@{}l@{}}~11k, \\  0.5k, \\ ~~~1k \end{tabular} &
  \begin{tabular}[c]{@{}l@{}}News, \\ Review, \\ Query\end{tabular} &
  \begin{tabular}[c]{@{}l@{}}Colloquial, \\ Formal\end{tabular} \\ \midrule
KLUE-NLI &
  \begin{tabular}[c]{@{}l@{}}Natural \\ Language \\ Inference\end{tabular} &
  \begin{tabular}[c]{@{}l@{}}Sentence \\ Pair \\ Classification\end{tabular} &
  Accuracy &
  3 &
  \begin{tabular}[c]{@{}l@{}}~~25k, \\  ~~~~3k, \\  ~~~~3k\end{tabular} &
  \begin{tabular}[c]{@{}l@{}}News, \\ Wikipedia, \\ Review\end{tabular} &
  \begin{tabular}[c]{@{}l@{}}Colloquial, \\ Formal\end{tabular} \\ \midrule
KLUE-NER &
  \begin{tabular}[c]{@{}l@{}}Named \\ Entity \\ Recognition\end{tabular} &
  \begin{tabular}[c]{@{}l@{}}Sequence \\ Tagging\end{tabular} &
  \begin{tabular}[c]{@{}l@{}}Entity-level Macro F1 \\ Character-level Macro F1 \end{tabular} &
  \begin{tabular}[c]{@{}c@{}}6,\\12\end{tabular}  &
  \begin{tabular}[c]{@{}l@{}} ~~21k, \\ ~~~~5k, \\ ~~~~5k \end{tabular} &
  \begin{tabular}[c]{@{}l@{}}News, \\ Review\end{tabular} &
  \begin{tabular}[c]{@{}l@{}}Colloquial, \\ Formal\end{tabular} \\ \midrule
KLUE-RE &
  \begin{tabular}[c]{@{}l@{}}Relation \\ Extraction\end{tabular} &
  \begin{tabular}[c]{@{}l@{}}Single Sentence \\ Classification \\ 
    \tiny{(+2 Entity Spans)}
    \end{tabular} &
  \begin{tabular}[c]{@{}l@{}} Micro F1 \footnotesize{(without \textit{no\_relation})}, \\ AUPRC \end{tabular} &
  30 &
  \begin{tabular}[c]{@{}l@{}} ~~32k, \\ ~~~~8k, \\ ~~~~8k \end{tabular} &
  \begin{tabular}[c]{@{}l@{}}Wikipedia, \\ News\end{tabular} &
  \begin{tabular}[c]{@{}l@{}}Formal\end{tabular} \\ \midrule
KLUE-DP &
  \begin{tabular}[c]{@{}l@{}}Dependency \\ Parsing\end{tabular} &
  \begin{tabular}[c]{@{}l@{}}Sequence \\ Tagging \\ 
    \tiny{(+ POS Tags)}
    \end{tabular} &
  \begin{tabular}[c]{@{}l@{}}Unlabeled Attachment Score, \\ Labeled Attachment Score \end{tabular} &
  \begin{tabular}[c]{@{}c@{}}\# Words,\\38\end{tabular} &
  \begin{tabular}[c]{@{}l@{}} ~~10k, \\  ~~~~2k, \\  ~2.5k \end{tabular} &
  \begin{tabular}[c]{@{}l@{}}News, \\ Review\end{tabular} &
  \begin{tabular}[c]{@{}l@{}}Colloquial, \\ Formal\end{tabular} \\ \midrule
KLUE-MRC &
  \begin{tabular}[c]{@{}l@{}}Machine \\ Reading \\ Comprehension\end{tabular} &
  \begin{tabular}[c]{@{}l@{}}Span Prediction \end{tabular} &
  \begin{tabular}[c]{@{}l@{}}Exact Match, \\ ROUGE-W (LCCS-based F1)\end{tabular} &
  2 &
  \begin{tabular}[c]{@{}l@{}} ~~12k,\\  ~~~~8k,\\  ~~~~9k\end{tabular} &
  \begin{tabular}[c]{@{}l@{}}Wikipedia, \\ News\end{tabular} &
  Formal \\ \midrule
\begin{tabular}[c]{@{}l@{}}KLUE-DST \\ (WoS) \end{tabular} &
  \begin{tabular}[c]{@{}l@{}}Dialogue \\ State \\ Tracking\end{tabular} &
  \begin{tabular}[c]{@{}l@{}} 
    Slot-Value \\ 
    Prediction \\ 
    \end{tabular} &
  \begin{tabular}[c]{@{}l@{}}Joint Goal Accuracy \\ Slot Micro F1\end{tabular} &
  (45) &
  \begin{tabular}[c]{@{}l@{}} ~~~8k, \\ ~~~1k, \\ ~~~1k\end{tabular} &
  \begin{tabular}[c]{@{}l@{}} Task \\ Oriented \\ Dialogue\end{tabular} &
  Colloquial \\ \bottomrule
\end{tabular}
\end{adjustbox}
\end{table}

\paragraph{Considerations in Annotation} 

For each task, we annotate a subset from the source corpora. In doing so, we take into account three major considerations below:

\begin{itemize}[leftmargin=*]

\item \textit{Better reflection of linguistic characteristics of Korean}: Many existing Korean datasets were constructed as a part of multilingually aligned benchmarks, and they do not fully reflect linguistic characteristics of Korean such as agglutinative nature in named entity recognition (NER) \cite{pan-etal-2017-cross}, or tagset in part-of-speech (POS) tagging and dependency parsing \cite{mcdonald2013universal,han2020annotation}. We write and revise annotation guidelines more appropriately to the linguistic property of Korean.

\item \textit{Obtaining accurate annotations}: 
We provide crowdworkers or select participants with carefully designed annotation guidelines and improve them over multiple iterations, in order to reduce the ambiguity of annotation process as well as to mitigate known artifact issues. In particular, we often filter out examples for which annotators cannot easily agree with each other.

\item \textit{Mitigating harmful social bias and removing PII}: 
To disincentivize socially biased NLU systems \cite{bowman2021will}, we explicitly instruct annotators as well as inspectors to manually mark and/or exclude examples that are unacceptable according to our principle of ethics. Our definitions of \textit{bias} and \textit{hate speech} follow \citet{moon-etal-2020-beep}. We denote \textit{bias} as an overgeneralized prejudice on certain groups or individuals based on the following traits: gender, race, background, nationality, ethnic group, political stance, skin color, religion, disability, age, appearance, (socio-)economic status, and occupations. In the case of \textit{hate speech}, we include offensive, aggressive, insulting, or sarcastic contents. We identify a list of personally identifiable information (PII) following KISA (Korea Internet and Security Agency) guideline,\footnote{\url{https://www.kisa.or.kr/public/laws/laws2_View.jsp?cPage=1\&mode=view\&p_No=282\&b_No=282\&d_No=3}} whose information is related to a living individual based on personal information protection act of Korea.\footnote{https://www.law.go.kr/LSW//lsInfoP.do?lsiSeq=213857\&chrClsCd=010203\&urlMode=engLsInfoR\&viewCls=engLsInfoR\#0000} We do not consider public figure's name as personal information.\footnote{
See the precedent set by the Supreme Court in Korea: 대법원 2011. 9. 2. 선고 2008다42430 전원합의체 판결 available at \url{https://glaw.scourt.go.kr/wsjo/panre/sjo100.do?contId=2060159&q=2008\%EB\%8B\%A442430}.
}
\label{sec:human-check-guideline}

\end{itemize}

\paragraph{Evaluation Metrics} 

The diversity of tasks in KLUE implies that we must choose a proper set of evaluation metrics for each task carefully and separately. Here, we list the tasks and describe how we choose the evaluation metrics for each of these tasks.

\begin{itemize}[leftmargin=*]

\item \textit{KLUE-TC} (Yonhap News Agency Topic Classification (\ynat{})): 
We formulate KLUE-TC as a multi-class classification problem with seven classes. Because the headline alone is often not enough to precisely identify the proper class to which it belongs, we manually annotate and keep 70,000 headlines, for each of which there was a majority consensus on the class by the annotators. We then use the consensus classes as ground-truth classes and use macro F1 score as an evaluation metric. 

\item \textit{\kluests{}}: 
In \kluests{} the similarity between each pair of sentences is annotated with the average (real-valued) similarity rating (between 0 and 5). We measure the quality of an NLU model in two different ways. First, we use the Pearson correlation coefficient between the real-valued target and prediction. Second, we compute the F1 score after binarizing the real-valued similarity rating as in paraphrase detection.

\item \textit{\kluenli{}}: 
Similar to existing NLI datasets, such as SNLI \cite{bowman2015large} and MNLI \cite{williams2017broad}, we use classification accuracy, and this is appropriate, as we create \kluenli{} dev/test set to have a balanced class distribution.

\item \textit{\kluener{}}: 
In \kluener{}, a named entity recognizer is expected to output BIO tags and also categorize each detected entity into one of six types; person, location, organization, date, time and quantity. To account for rich morphology in Korean, we use entity-level and character-level F1 score to evaluate the quality of the detection to evaluate the recognizer's ability in determining the type of each entity.

\item \textit{\kluere{}}: 
\kluere{} is designed as a sentence classification task in which the input is a single sentence with two marked entities and the output is their relationship out of 30 types. We use two evaluation metrics. The first one is micro F1 score, considering only meaningful types (excluding no relationship), which allows us to evaluate the NLU system's ability to identify a fine-grained relationship between a pair of entities. The second one is the area under the precision-recall curve (AUPRC), which gives us a holistic view into the quality of the relation extraction model in question.

\item \textit{\klueposdp{}}: 
Following standard practice in dependency parsing, we use both unlabeled attachment score (UAS) and labeled attachment score (LAS) to evaluate a dependency parser. We annotate and use both formal and informal text (subsets from the news corpora and colloquial review corpora, respectively), which allows us to perform fine-grained analysis across multiple domains.

\item \textit{\kluemrc{}}: 
Similarly to \kluener{}, \kluemrc{} is framed as a span prediction problem. We keep character-level exact match (EM) for comparison against existing datasets, while we propose to use ROUGE-W which measures the F1 score based on the longest common consecutive subsequence (LCCS) between the ground-truth and predicted answer spans. The latter handles rich morphology of Korean as well as the former does while being more interpretable.

\item \textit{KLUE-DST} (Wizard of Seoul, \wizard{}): 
We formulate KLUE-DST as a multiple-sentence slot-value prediction task, and evaluate an NLU system using two metrics. The first metric is the joint goal accuracy which measures whether all the slots were correctly predicted, while the other metric is average F1 score. Because the former treats all examples for which not all slots were correctly filled in, it often fails to distinguish similarly performing NLU systems. We address this shortcoming by reporting both the joint goal accuracy and slot F1 score. We furthermore build it using multiple domains in order to facilitate finer-grain analysis.

\end{itemize}

\paragraph{Baselines}

In addition to creating a benchmark suite, we also build and publicly release a set of strong baselines based on large-scale pretrained language models. In due course, we pretrain and release large-scale language models for Korean ourselves, which will reduce the burden of retraining these large-scale models from individual researchers. We also use several existing multilingual pretrained language models and open-source Korean-specific models in addition to our own models, to gain further insights into the proposed KLUE benchmark. We present all the results in Table~\ref{table:main_results} and summarize a few interesting observations here. First, Korean-specific language models generally outperform multilingual models. Second, different models perform best on different tasks when controlled for their sizes; \kluebert{} performs best for \ynat{} and \wizard{}, \klueroberta{} for \kluere{} and \kluemrc{}, and \koelectrabase{} for \kluests{} and \kluenli{}. Third, as we increase the model size, \robertalarge{} ends up outperforming all the other models in all the tasks other than \kluener{}. Lastly, we observe that removing PII has minimal effect on the downstream task performances, and our tokenization scheme, morpheme-based subword tokenization, is effective in tasks involving tagging, detection and even generation at the morpheme level.

\paragraph{Task Overview} 

In Table~\ref{tab:task-overview}, we summarize the resulting eight KLUE tasks, listing important properties, such as type, format, evaluation metrics and annotated data sizes. In the rest of the paper, we will walk through the process by which each and every one of these tasks was constructed much more in detail.


\clearpage
\section{Source Corpora}
We build KLUE from scratch, instead of putting together existing datasets, which has been a common practice in setting up benchmarks. We investigate available textual resources, and document the process in order to provide better understanding on how and why we select some corpora but not others. We adopt the recently proposed documentation frameworks; \textit{datasheets} \cite{gebru2018datasheets} and \textit{data statements} \cite{bender-friedman-2018-data}. Based on these frameworks, we document and provide more information to carefully describe our protocol.

\subsection{Corpora Selection Criteria}
We consider two criteria when sourcing a set of corpora to build a source corpus from which task-specific corpora are derived and annotated. The first criterion is accessibility. As the main purpose of KLUE is to facilitate future NLP research and development, we ensure KLUE comes with data that can be used and shared as freely as possible to all. The second criterion is the quality and diversity. We ensure each example with these corpora is of certain quality by removing low-quality text and also the balance is met between formal and colloquial text within these corpora.

\paragraph{Accessibility} 

Unlike \citet{wang2019superglue, hu2020xtreme, kakwani2020indicnlpsuite}, we design KLUE to reach as broad and diverse researchers as possible by avoiding any restriction on affiliations of users as well as the purpose of its use. Furthermore, we acknowledge the rapid pace of advances in the field and allow users to reproduce and redistribute KLUE to prolong its usability as a standard benchmark of NLU. To do so, we build and release the source corpus with CC BY-SA.\footnote{
\url{https://creativecommons.org/licenses/by-sa/4.0/}
}

The source corpus, or a set of source corpora, satisfies the following conditions:
\begin{itemize}[leftmargin=*]
    \item \textbf{No restriction on the use:} 
    We allow both non-commercial and commercial use of KLUE, in order to accommodate the recent trend of fundamental research from industry labs. 

    \item \textbf{Derivatives:} 
    We allow users to freely refurbish any part of KLUE to first address any shortcomings, such as unanticipated artifacts, ethical issues and annotation mistakes, and second derive more challenging benchmarks for the future. This is similar to what has been done with SQuAD 2.0~\cite{rajpurkar2018know} which was created to include SQuAD 1.1 \cite{rajpurkar2016squad}.
    
    \item \textbf{Redistributable:} 
    We allow KLUE benchmark datasets to be distributed by anyone via any channel as long as the proper attribution is given to the original creators of KLUE. 
    We deliberately make this decision to avoid situations where only a limited and select group of researchers have a monopoly on resources, ultimately hindering the progress overall. This is in reaction to some of the existing Korean corpora which come together with restrictive policies, often preventing derivatives as well as redistribution, and are only accessible by researchers in Korea after acquiring permissions from the corpus publishers who are often public institutions in Korea. KLUE avoids such preventive policies in order to maximally facilitate the progress in Korean NLP.
\end{itemize}

Because most of the existing datasets do not meet these conditions, we curate the source corpus from scratch by considering only those resources that either come with one of the following licenses: CC0,\footnote{\url{https://creativecommons.org/publicdomain/zero/1.0/}} CC BY,\footnote{\url{https://creativecommons.org/licenses/by/4.0/}} CC BY-SA,\footnote{\url{https://creativecommons.org/licenses/by-sa/4.0/}} and other similar licenses such as KOGL Type 1,\footnote{\url{https://www.kogl.or.kr/info/license.do\#05-tab}} 
are not protected by the copyright act according to the latest copyright act in Korea,\footnote{
See \url{https://www.law.go.kr/\%EB\%B2\%95\%EB\%A0\%B9/\%EC\%A0\%80\%EC\%9E\%91\%EA\%B6\%8C\%EB\%B2\%95} for the copyright act which went effective as of Dec 8 2020.
}
or have been explicitly provided to us by copyright holders under contracts.
We end up 20 candidate corpora in total, of which subset is selected to form a source corpus set of KLUE. They are listed in Table~\ref{tab:source-corpus-all}.

\paragraph{Quality and Diversity} 

Among these 20 source corpora, we select a subset of ten corpora to form the source corpus and to build the KLUE benchmark. In doing so, we consider the following criteria; 1) the corpus should not be specific to narrow domains (diversity), 2) the corpus must be written in contemporary Korean (quality), 3) the corpus should not be dominated by contents that have privacy or toxicity concerns (quality) and 4) the corpus must be amenable to annotation for at least one of the eight benchmark tasks. Furthermore, we select the subset of corpora to cover both formal and colloquial uses. 

\paragraph{The Final Source Corpora}

Based on these criteria and decisions, we choose News Headlines, Wikipedia, Wikinews, Policy News, The Korea Economics Daily News, and Acrofan News for (relatively) formal text.\footnote{
Although Wikitree was found to include some contents that could be considered unethical, socially biased and/or of low quality in general, we include it, as Wikitree is the largest source of license-free news articles. We address these problematic contents via annotation.
} 
For more colloquial text, we use ParaKQC, Airbnb Reviews, and NAVER Sentiment Movie Corpus. These are marked bold in Table~\ref{tab:source-corpus-all}.

\begin{table}[t!]
\caption{Collected source corpora. The corpora in the first section are not protected by copyright act. Specifically, \textit{News Headlines} are not classified as a work due to their lack of creativity and \textit{Judgements} are not protected works under Article 7, Act 3. \textit{National Assembly Minutes} and \textit{Patents}, made in National Assembly, shall not apply the copyright act by Article 24, Act 2. The second section is a collection of corpora under the permissive licenses. The last section corpora, KED and Acrofan, are originally prohibited from creating derivative works, however, we release such condition by exclusive contract. For the column, \textit{Volume}, we denote \textit{Small} as corpus size under 1k, \textit{Medium} as in between 1k and 50k, and \textit{Large} as over 50k. Bold represents our final source corpora to build KLUE benchmark.}
\label{tab:source-corpus-all}
\centering
\begin{adjustbox}{width=1\textwidth}
\begin{tabular}{@{}lcccccc@{}}
\toprule
\textbf{Dataset} & \textbf{License} & \textbf{Domain} & \textbf{Style} & \textbf{\textbf{\begin{tabular}[c]{@{}c@{}}Ethical\\ Risks\end{tabular}}}  & \textbf{Volume} & \textbf{\begin{tabular}[c]{@{}c@{}}Contemporary \\ Korean\end{tabular}} \\ \midrule \medskip
\textbf{News Headlines} & \textbf{N/A} & \textbf{\begin{tabular}[c]{@{}c@{}}News   (Headline)\end{tabular}} & \textbf{Formal} & \textbf{Low} & \textbf{Large} & \textbf{o} \\ \medskip
Judgments & Public Domain & Law & Formal & Low & Large & o  \\ \medskip
\begin{tabular}[c]{@{}l@{}}National Assembly \\ Minutes\end{tabular} & Public Domain & Politics  & Colloquial & Medium & Large & o  \\ \medskip
Patents & Public Domain & Patent & Formal & Low & Large & o  \\ \midrule \medskip
\textbf{Wikipedia} & \textbf{CC BY-SA 3.0} & \textbf{Wikipedia} & \textbf{Formal} & \textbf{Low} & \textbf{Large} & \textbf{o} \\ \medskip
Wikibooks & CC BY-SA 3.0  & Book & Formal & Low & Medium & x \\ 
\multirow{2}{*}{Wikisource} & \multirow{2}{*}{CC BY-SA 3.0} & Law & \multirow{2}{*}{Formal} & \multirow{2}{*}{Low}  & \multirow{2}{*}{Medium} & \multirow{2}{*}{x} \\ \medskip
 &  & Book &  &  & \\ \medskip
\textbf{Wikinews} & \textbf{CC BY 2.5} & \textbf{News} & \textbf{Formal} & \textbf{Low} & \textbf{Small} & \textbf{o}  \\ \medskip
\textbf{Wikitree} & \textbf{CC BY-SA 2.0} & \textbf{News} & \textbf{Formal} & \textbf{Medium} & \textbf{Large} & \textbf{o}  \\ \medskip
Librewiki & CC BY-SA 3.0 & Wiki & Formal & Medium & Large & o \\ \medskip
Zetawiki & CC BY-SA 3.0 & Wiki & Formal & Medium & Large & o \\ \medskip
\textbf{Policy News} & \textbf{KOGL Type 1} & \textbf{News} & \textbf{Formal} & \textbf{Low} & \textbf{Medium} & \textbf{o}  \\ \medskip
\begin{tabular}[c]{@{}l@{}}NIKL Standard \\ Korean Dictionary\end{tabular} & CC BY-SA 2.0 & Dictionary & Formal & Low & Large & o  \\ \medskip
\begin{tabular}[c]{@{}l@{}}Open \\ Korean Dictionary\end{tabular} & CC BY-SA 2.0 & Dictionary & Formal & Low & Large & o  \\ \medskip
\textbf{ParaKQC} & \textbf{CC BY-SA 4.0}  & \textbf{\begin{tabular}[c]{@{}c@{}}Smart Home \\ Utterances\end{tabular}} & \textbf{Colloquial} & \textbf{Low} & \textbf{Medium} & \textbf{o}\\ \medskip
\textbf{Airbnb Reviews} & \textbf{CC0 1.0} & \textbf{Review} & \textbf{Colloquial} & \textbf{Medium} & \textbf{Large} & \textbf{o} \\ \medskip
\textbf{\begin{tabular}[c]{@{}l@{}}NAVER Sentiment \\ Movie Corpus (NSMC) \end{tabular}} & \textbf{CC0 1.0} & \textbf{Review} & \textbf{Colloquial} & \textbf{Medium} & \textbf{Large} & \textbf{o}  \\ \medskip
\begin{tabular}[c]{@{}l@{}}NAVER Entertainment \\ News Reviews\end{tabular} & CC BY-SA 4.0  & Review & Colloquial & High & Large & o\\ \midrule \medskip
\textbf{Acrofan News} & \textbf{\begin{tabular}[c]{@{}c@{}}CC BY-SA 4.0 \\ for \kluemrc{} by Contract\end{tabular}} & \textbf{News} & \textbf{Formal} & \textbf{Low} & \textbf{Large} & \textbf{o}  \\
\textbf{\begin{tabular}[c]{@{}l@{}}The Korea Economics \\ Daily News\end{tabular}}  & \textbf{\begin{tabular}[c]{@{}c@{}}CC BY-SA 4.0 \\ for \kluemrc{} by Contract\end{tabular}}  & \textbf{News} & \textbf{Formal} & \textbf{Low} & \textbf{Large} & \textbf{o}\\
 \bottomrule
\end{tabular}
\end{adjustbox}
\end{table}

\subsection{Selected Corpora}
Here, we describe in more detail general characteristics and potential concerns of each source corpus. We document the collection mechanisms, timeframe, domain, style, license, and background of each corpus as well.

\paragraph{News Headlines} from Yonhap News Agency (\yna). 

{\yna}\index{YNA (Yonhap News Agency)} is a dataset of news headlines from Yonhap News Agency, one of the representative news agencies in South Korea. Using news headlines does not infringe on copyrights, unlike the actual contents of news articles. We include {\yna} from 2016 to 2020 with a main purpose of using it for a single sentence classification task.

\paragraph{Wikipedia (\wikipedia)}

{\wikipedia}\index{WIKIPEDIA} is an open encyclopedia written in a formal style and has been widely used for language modeling and dataset construction across many languages, because of its high-quality and well-curated text. The Wikipedia articles in Korean are released under CC BY-SA. We use the dump of Korean Wikipedia released on December 1st, 2020.

\paragraph{Wikinews (\wikinews)}
{\wikinews}\index{WIKINEWS} implements collective journalism and provides news articles for free under CC BY, both of which are rare for news articles. Due to these properties, we include it in the source corpora despite its limited number of articles (approximately 500 of them).

\paragraph{Wikitree (\wikitree)}
{\wikitree}\index{WIKITREE} is a dataset of news articles derived from Wikitree, the first Korean social media-based news platform that started in 2010. 
Although there are concerns that the articles on Wikitree are in many cases advertisement-in-disguise or click-bait headlines and express undesirable biases,
we include {\wikitree}, as it is the only large-scale source of news articles that are freely distributed under CC BY-SA, to the best of our knowledge. It also covers a broad spectrum of topics, including politics, economics, culture and life. We use the articles published between 2016 and 2020. We conduct more thorough manual inspection of \wikitree{} is more thoroughly conducted. See Section~\ref{sec:potential_ethical_concern} for more details.

\paragraph{Policy News (\policy)}
{\policy}\index{POLICY (Policy News)} is a dataset of various articles distributed by ministries, national offices, and national commissions of South Korea. It covers statements, notices, or media notes reported by the government agencies. {\policy} is protected under the Korea Open Government License (KOGL) Type 1, which permits users to share and remix even for commercial purposes, if attribution is properly done. We include articles released up to the end of 2020.

\paragraph{ParaKQC (\parakqc)}
{\parakqc}\index{PARAKQC} is a dataset of 10,000 utterances aimed at smart home devices, consisting of 1,000 intents of 10 similar queries \cite{cho2020discourse}. It covers various topics which are probable when interacting with smart home speakers, such as scheduling an appointment and asking about the weather. {\parakqc} is available under CC BY-SA.

\paragraph{Airbnb Reviews (\airbnb)}
{\airbnb}\index{AIRBNB (Airbnb Reviews)} is a review dataset sourced from the publicly accessible portion of the Airbnb website. More specifically, we start from the existing multilingual Airbnb reviews collected and preprocessed by Inside Airbnb.\footnote{
\url{http://insideairbnb.com/get-the-data.html}
} 
We identify a subset of reviews written in Korean from this multilingual Airbnb corpus, using regular expressions. Reviews are from hosts and guests who have completed their stays. \airbnb{} is available under CC0.

\paragraph{NAVER Sentiment Movie Corpus (\nsmc)}
{\nsmc}\index{NSMC (NAVER Sentiment Movie Corpus)} is a movie review dataset scraped from NAVER Movies.\footnote{
\url{https://movie.naver.com/movie/point/af/list.nhn}
} 
The reviews are written by online users. Each review comes with both the textual content and the binary sentiment label. There are 200,000 reviews in total. 
The numbers of positive and negative reviewers are balanced. \nsmc{} is available under CC0.

\paragraph{Acrofan News (\acrofan)}
\acrofan{}\index{ACROFAN (Acrofan News)} is a corpus consisting of news articles released by ACROFAN. Most articles are press release-like in that they often introduce new products or events of companies. The formats and styles are quite templated, although the articles cover a broad set of categories including automobiles, IT, startups, big companies, energy, beauty and fashion. 
We obtain the permission and use of the articles from ACROFAN for KLUE. We include news articles published between Dec 2020 and Jan 2021.

\paragraph{The Korea Economics Daily News (\hankyung)}

{\hankyung}\index{The Korea Economy Daily} is a news corpus consisting of articles from the Korea Economics Daily owned by Hankyung corporation. Korea Economics Daily is a newspaper that mainly covers economic issues, but also publishes various topics such as politics, culture and IT topics. The owner of the Korea Economics Daily and we have entered a contract to use news articles published between Jan 2013 and Dec 2015, provided by the Hankyung corporation, as a part of KLUE. This allows us to ensure high-quality, well-curated news articles are included in KLUE. We release \hankyung{} under CC BY-SA, with the condition that these articles are used for the purpose of machine learning research.

\subsubsection{Potential Concerns}
\label{sec:potential_ethical_concern}

Based on the ten selected corpora above, we list up and discuss some of the concerns here. Some concerns are focused on the quality of data, while the others are more societal and ethical. 

\paragraph{Toxic Content} 

Although news articles, such as those from \yna{}, \wikinews{}, \wikitree{}, \policy{}, \acrofan{}, and \hankyung{} are better written and curated than user-generated contents, such as online reviews, these articles nevertheless may reflect some of the biases possessed by journalists and editors. In particular, our manual inspection has revealed that \wikitree{} contains more of potentially problematic patterns than the other news sources, due to the incentive structure that incentivizes articles that are more widely shared and clicked more on social media. This is especially true with headlines of these articles, and we thus refrain from using the headlines from \wikitree{} when constructing TC. We also do not use the article contents from \wikitree{} for MRC, as articles in whole often exaggerate and emphasize sensational aspects of stories. We however use sentences sampled from \wikitree{} when building other task-specific corpora, as they are often complete and well-formed. We discard any problematic sentences via annotation.

Unlike news articles, online reviews have higher potential to contain toxic content, although such tendency varies from one corpus to another. Due to its peer-reviewing system, \airbnb{} rarely contains reviews that are deemed toxic. \nsmc{} on the other hand contains comments that could be considered offensive toward movies, their casts, and their directors. As there is a Korean hate speech dataset on review domains~\cite{moon-etal-2020-beep}, we first filter out toxic content with a detector trained on the dataset. Then we discard problematic sentences via the annotation procedure.

All utterances of \parakqc{} are carefully created based on a pre-defined annotation guideline \cite{cho2020discourse}. This largely prevents toxic content from entering the corpus.

\paragraph{Personally Identifiable Information (PII)}

Private information is any information that can be used to identify an individual who is not considered a public figure.\footnote{
See the precedent set by the Supreme Court in Korea: 대법원 2011. 9. 2. 선고 2008다42430 전원합의체 판결 available at \url{https://glaw.scourt.go.kr/wsjo/panre/sjo100.do?contId=2060159&q=2008\%EB\%8B\%A442430}.
} 
It includes for instance names, social security numbers, telephone numbers and bank account numbers. 

In the case of news articles, due to their nature of describing social events, they often contain PII such as names and addresses. This is less so with online reviews, as they are often about public figures, such as actors, actresses and directors, as we observe in \nsmc{}. We however notice that the reviews in \airbnb{} contain the names of hosts and/or guests as well as their addresses, which must be carefully handled. 

Some of the artificially generated utterances in \parakqc{} do contain names. It is however our understanding that these are mostly fictional, meaning that they are unlikely to be truly private information.

\subsection{Preprocessing}
Because these source corpora come from various sources with varying levels of quality and curation, we carefully preprocess them even before deriving a subset for each downstream task. In this section, we describe our preprocessing routines which are applied after splitting each document within these corpora into sentences using the Korean Sentence Splitter (KSS) v2.2.0.2.\footnote{
\url{https://github.com/hyunwoongko/kss}
}
The proprocessing routines below are {\it in addition to} manual inspection and filtering during the annotation stage of each KLUE task.

\paragraph{Noise Filtering}

We  remove noisy and/or non-Korean text from the selected source corpora. 
We first remove hashtags (e.g., \#JMT), HTML tags (e.g., <br>), bad characters (e.g., U+200B (zero-width space), U+FEFF (byte order mark)), empty parenthesis (e.g., ()), and consecutive blanks. We then filter out sentences with more than 10 Chinese or Japanese characters. For the corpora derived from news articles, we remove information about reporters and press, images, source tags as well as copyright tags (e.g., copyright by ©).

\paragraph{Toxic Content Removal}

In order to avoid introducing undesire contents and biases into KLUE, we use a number of automatic tools to remove various undesirable sentences from the source corpora. Using the Korean hate speech dataset~\cite{moon-etal-2020-beep}, we train a gender bias\footnote{
\url{https://huggingface.co/monologg/koelectra-base-v3-gender-bias}
} 
and a hate speech detector.\footnote{
\url{https://huggingface.co/monologg/koelectra-base-v3-hate-speech}
} 
We discard a sentence which was predicted to exhibit gender bias with the predictive score of at least $0.5$.  We also discard a sentence if it was deemed to be hate speech, with the predictive score of $0.9$ or above. The thresholds are manually determined for each corpus. This approach work well for online text, such as reviews, because the Korean hate speech dataset was constructed using online reviews. It however does not work well for more formal text, such as found in news articles, based on which we decide against using this strategy on \hankyung{}, \acrofan{}, and \yna{}.

\paragraph{PII Removal}

To mitigate potential privacy issues, we get rid of sentences that contain private information. We detect such sentences using regular expressions that match email addresses, URL and user-mentioning keywords, such as `@gildong'.

\label{sec:corpus-preprocessing}

\subsection{Task Assignment}
We use these source corpora to build the datasets for the seven KLUE tasks, except for the \dst{}. \dst{} is built from simulated dialogues by crowdworkers and does not require access to offline text. For each downstream task, we use a subset of the source corpora, as described below:

\begin{itemize}[leftmargin=*]
    \item Topic Classification (\tc{}): We use \yna{}, which has been widely studied for a single sentence topic classification task.
    
    \item Semantic Textual Similarity (\sts{}): We use {\airbnb}, {\policy}, and {\parakqc} to include diverse semantic contexts. Intent queries and topic information of \parakqc{} are useful when generating semantically related sentence pairs. 
    
    \item Natural Language Inference (\nli{}): Following MNLI \cite{williams2017broad}, we use multiple sources to construct \nli{}. We use \wikitree{}, \policy{}, \wikinews{}, \wikipedia{}, \nsmc{} and \airbnb{}.
    
    \item Named Entity Recognition (\ner{}): Due to the nature of \ner{}, we must build a corpus in which (named) entities frequently appear. We thus use \wikitree{} and \nsmc{}, which enables us to include both formal and informal writing styles.
    
    \item Relation Extraction (\re{}): 
    We use {\wikipedia}, {\wikitree} and {\policy}. These corpora tend to have long complete sentences with the names of public figures and their relationships to various organizations.
    
    \item Dependency Parsing (\posdp{}): We balance formal and colloquial writing styles, while ensuring most of sentences from selected corpora are complete.
    We end up using \wikitree{} and \airbnb{}. We choose \airbnb{} over \nsmc{}, because the former has better-formed sentences.
    
    \item Machine Reading Comprehension (\mrc{}): To provide informative passages, we use \wikipedia{}, \hankyung{}, and \acrofan{}. 
\end{itemize}


\clearpage
\section{KLUE Benchmark}
\label{tasks}

The goal of KLUE is to provide high quality evaluation datasets and suitable automatic metrics to test a system’s ability to understand Korean language. We provide comprehensive details on how we construct our 8 benchmark datasets. We document 1) background of source corpus selection, 2) annotation protocol, 3) annotation process, 4) dataset split strategy, and 5) design process of the metrics. In the annotation process, we guide workers to identify texts containing potential ethical issues. See Section~\ref{sec:human-check-guideline} for our definitions on bias, hate, and PII.

\subsection{Topic Classification (\tc)}
In topic classification (TC)\index{TC (Topic Classification)}, the goal is to train a classifier to predict the topic of a given text snippet. Topic classification datasets typically consist of news or Wikipedia articles and their predefined categories, because the categories often represent topics \cite{zhang2015character}. 

We include TC in our KLUE benchmark, as inferring the topic of a text is a key capability that should be possessed by a language understanding system. As a typical single sentence classification task, other NLU benchmarks such as CLUE \cite{xu2020clue} and IndicGLUE \cite{kakwani2020indicnlpsuite} also contain TNEWS and News Category Classification. For Korean, no dataset has been proposed for the task, which motivates us to construct the first Korean topic classification benchmark. 

In this task, given a news headline, a text classifier must predict a topic which is one of \{politics, economy, society, culture, world, IT/science, sports\}. We formulate TC as single sentence classification task following previous works and use macro-F1 score as an evaluation metric.

\subsubsection{Dataset Construction}
Our TC benchmark is constructed in three stages. First, we collect headlines and their corresponding categories, then we annotate the topics without looking at the categories and we finalize the dataset by defining its split into training, development and test splits considering the publication date and term appearances.

\paragraph{Source Corpora}
We collect news headlines from online articles distributed by Yonhap News Agency (YNA), the largest news agency in Korea. Specifically, we collect the headlines of the published articles from January 2016 to December 2020 from Naver News.\footnote{\url{https://news.naver.com/}} These articles belong to one of the following seven sections: politics, economy, society, culture, world, IT/science, and sports. To balance the data across the different sections, we randomly sample 10,000 articles from each section, except for the sports and IT/science section. We collect 9,000 sports articles and 11,000 IT/science articles.

Unlike other benchmarks such as TNEWS in CLUE \cite{xu2020clue} or AG News \cite{zhang2015character}, we exclude contents of the articles to avoid infringement of copyright. Since the contents are protected as copyrighted work, we cannot freely use them without permission. Headlines, on the other hand, are not considered copyrighted work based on a legal precedent \cite{kcc2009copyright}.

\paragraph{Annotation Protocol}
The headline of each article may not reflect all of the main content, such that the \textit{topic} of the headline may be different from the original news section of the article. To address this gap between the headline and the corresponding article, we manually annotate the topics of the headlines.

We use SelectStar,\footnote{\url{https://selectstar.ai/}} a crowdsourcing platform in Korea, to annotate topics of the headlines. For each headline, three annotators label topics independently from each other. Each annotator picks at most three topics in the order of relevance among the seven categories. For precise annotation, we also present \textit{key terms} of each topic to annotators. The terms are subsections of corresponding topics in NAVER news platform as shown in Table~\ref{tab:ynat-stats}.

An annotator may choose \textit{unable-to-decide} if the headline does not contain sufficient information to identify the appropriate categories. Such an example is ``Youngsoo Kim awards an appreciation plaque''. There is no clue about who ``Youngsoo Kim'' is nor why he is awarding the appreciation plaque, in this headline. 

We request the workers to report any headline that includes personally identifiable information (PII), expresses social bias, or is hate speech. We discard the reported headline after manually reviewing them.

\paragraph{Annotation Process}
We run a pilot study to select workers, before commencing the main annotation process. We exclude workers who have continuously failed to assign a topic or have failed to agree with the other workers during the pilot stage. As a result, 13 workers have passed this stage of pilot study.

In the main annotation, the 13 selected workers labeled topics for all 70,000 headlines. During the annotation, they reported 650 headlines are including potential PIIs (0.93\%), 194 toxic contents (0.28\%), and 2,515 \textit{unable-to-decide}s (3.59\%). We first exclude such invalid 2,953 headlines. The sum of the three type of problematic headlines are larger than the total value because of the intersection among them. After filtering them, 67,047 headlines remain.

We look at agreements between three annotators in valid headlines. We consider each of the first relevant topics chosen by three annotators. In 40,359 (60.5\%) headlines, all three annotators agree to a single topic. 23,353 (34.8\%) had two majority votes, and the other 3,155 (4.7\%) did not reached to agreement. To make the headlines classified to a single topic, we remove the others, leaving 63,892 headlines.

We examine the second and third relevant topics within an annotator. For 48,885 (69.8\%) of headlines, three annotators did not choose any second and third most relevant topic. Only 5,088 (7.3\%) of headlines have the second topic in three annotators. We thus assume that headlines are sufficiently represented by the first relevant topics within an annotator.

We thus keep only a single topic for each headline, selected by at least two annotators out of three. The annotator agreement on the resulting 63,892 headlines is fairly high (Krippendorff's $\alpha=0.713$) \cite{Krippendorff2011ComputingKA}. 

\paragraph{Final Dataset}
We partition the final dataset, named \ynat{} (Yonhap News Agency dataset for Topic classification), into train, development, and test sets. 
We split the dataset based on the publication date. We include headlines published after 2020 in the development and test sets, while those published before 2020 in the training set. To prevent TC models attending specific keyword to classify the headlines, we also include headlines containing terms that have not appeared in the train set in the development and test set. As shown in the Table~\ref{tab:ynat-stats}, train, development, and test sets consist of 45,678, 9,107, and 9,107 examples, respectively.

\begin{table}[t]
\centering
\caption{The final statistics of {\ynat} (KLUE-TC), provided with the key terms of each category.}
\label{tab:ynat-stats}
\begin{tabular}{@{}llcccc@{}}
\toprule
\textbf{Topic}     & \textbf{Key Terms} & \textbf{|Train|} & \textbf{|Dev|} & \textbf{|Test|} & \textbf{Total}  \\ \midrule \medskip
Politics   & \begin{tabular}{@{}l@{}}{\small Blue House, Ministry, Parliament, North Korea}\\\small{Political parties, Defense, Diplomacy}\end{tabular} &  7,379   & 750   & 722    & 8,851  \\ \medskip
Economy    & \begin{tabular}{@{}l@{}}{\small Stock, Finance, Industry Enterprise, Real estate}\end{tabular} &  6,118   & 1,268 & 1,348  & 8,734  \\ \medskip
Society    & \begin{tabular}{@{}l@{}}{\small Education, Labor, Journalism}\\ {\small Environment, Human rights, Food and drugs} \end{tabular} &  5,133  & 3,740 & 3,701  & 12,574 \\ \medskip
Culture    & \begin{tabular}{@{}l@{}}{\small Health, Transportation, Leisure, Hot places, Fashion}, \\ {\small Beauty, Performance, Exhibition, Books, Weather} \end{tabular} &  5,751   & 1,387 & 1,369  & 8,507  \\ \medskip
World      & \begin{tabular}{@{}l@{}}{\small Asia/Australia, America, Europe, Middle East/Africa} \end{tabular} &  8,320   & 776   & 835    & 9,931  \\ \medskip
IT/Science & \begin{tabular}{@{}l@{}}{\small Mobile, IT, Internet Social media, Communication} \\ {\small Computer, Game, Scientific journalism} \end{tabular} &  5,235   & 587   & 554    & 6,376  \\
Sports     & \begin{tabular}{@{}l@{}}{\small Baseball, Basketball, Volleyball, E-sports} \end{tabular} &  7,742   & 599   & 578    & 8,919  \\ \midrule
\textbf{Total}      & &\textbf{45,678}  & \textbf{9,107} & \textbf{9,107}  & \textbf{63,892} \\ \bottomrule
\end{tabular}
\end{table}

\subsubsection{Evaluation Metric}
\label{sec:tc-metric}

The evaluation metric for \ynat{} is macro F1 score. Macro F1 score is defined as the mean of topic-wise F1 scores, giving the same importance to each topic. Topic-wise F1 score weights recall and precision equally.

\subsubsection{Related Work}
Although many topic classification datasets have been proposed in various languages, we are not aware of any public TC benchmark in Korea. AG News \cite{zhang2015character}, a widely used benchmark for topic classification in English, consists of more than a million of news articles collected from the news search engine ComeToMyHead,\footnote{
More information available in \url{http://groups.di.unipi.it/~gulli/AG_corpus_of_news_articles.html}}
and categorizes articles into four sections: world, sports, business, and science/technology. More recently, a number of TC benchmark datasets in languages other than English were proposed.
IndicGLUE~\cite{kakwani2020indicnlpsuite} includes News Genre Classification in Indian languages, in which the goal is to classify a news article or news headline into seven categories; entertainment, sports, business, lifestyle, technology, politics, and crime.
TNEWS from CLUE \cite{xu2020clue} is a news topic classification task in Mandarin and consists of 73K titles with 15 news categories, published in Toutiao.

Since a large language model fine-tuned on TC benchmark can closely reach 100\% accuracy as in IndicGLUE~\cite{kakwani2020indicnlpsuite}, some researchers focus on making challenging TC benchmark to leave a room for improvement. CLUE \cite{xu2020clue} filters easy examples in TNEWS by using 4-fold cross-validation, and then randomly shuffle and split the dataset. Instead of designing our benchmark artificially more difficult, we reflect how topic classification is done in practice even a baseline model reaches to good performance with relatively easy examples in our benchmark.

\subsubsection{Conclusion}
We introduce \ynat\index{YNAT (YNA Topic Classification, KLUE-TC)}, the first Korean topic classification benchmark. The benchmark includes 63,892 news headlines classified to a single hand-labeled topic among 7 categories. We assume each headline has only a single topic, but it could be formulated as multi-label classification. We thus open the second and third relevant topic annotations. Also, URLs for each headlines are accompanied for future work if metadata is needed. If some of them requires permission to use, one should contact to the agency. We expect YNAT to serve as a simple and basic NLU task compared to others in KLUE.

\clearpage
\subsection{Semantic Textual Similarity (\sts)}
Semantic textual similarity (STS)\index{STS (Semantic Textual Similarity)} is to measure the degree of semantic equivalence between two sentences. We include STS in our benchmark because it is essential to other NLP tasks such as machine translation, summarization, and question answering. Like STS \cite{cer-etal-2017-semeval} in GLUE \cite{wang2018glue}, many NLU benchmarks include comparing semantic similarity of text snippets such as semantic similarity \cite{xu2020clue}, paraphrase detection \cite{wang2018glue, kakwani2020indicnlpsuite}, or word sense disambiguation \cite{shavrina2020russiansuperglue, le2020flaubert}. 

We formulate STS as a sentence pair regression task which predicts the semantic similarity of two input sentences as a real value from 0 (no meaning overlap) to 5 (meaning equivalence). A model performance is measured by Pearson's correlation coefficient following the evaluation scheme of STS-b \cite{cer-etal-2017-semeval}. We additionally binarize the real numbers into two classes with a threshold score 3.0 (paraphrased or not), and use F1 score to evaluate the model. 

\subsubsection{Dataset Construction}

\paragraph{Source Corpora} 
To diversify domain and style of source corpora, we collect sentences from {\airbnb} (colloquial review), {\policy} (formal news), and {\parakqc} \cite{cho2020discourse} (smart home utterances). We carefully match them to sentence pairs.

For each corpus, we design a sampling strategy of sentence pairs to uniformly cover all range of the similarity scores. Without a sophisticated strategy, simple random sampling and matching sentence to pairs would result in a majority of the score zero. To alleviate this skewness, \textit{potentially} similar and less similar sentences are separately paired by using various methods. For instance, if two descriptions are depicting the same image or headlines referring to the same event, they are likely to be similar because of the additional information. Otherwise, they would not be similar \cite{agirre2015semeval}. Inspired from these, we use available additional information to pair sentences as similar or not. If not available, we use round-trip translation (RTT) to obtain the similar pairs and \textit{greedy sentence matching} for the less similar pairs.

We specify the strategy for {\parakqc} where the intent of each sentence is available. All sentences are queries for a smart home domain and their intent are shared among some queries. For example, ``\textit{How's the weather today in Seoul?}'' and ``\textit{You know what the weather is like in Seoul today?}'' share the same intent which is asking ``The weather of Seoul today''. We pair two sentences with the same intent as similar pairs and different intent as the less similar. Note that even the less similar pairs share topic to avoid making too many mutually dissimilar pairs. 

For {\airbnb} and {\policy}, we cannot find meaningful metadata to estimate similarity between sentences. So we adopt RTT technique using NAVER Papago\footnote{\url{https://papago.naver.com/}} to generate the similar sentence pairs, since RTT is known to yield sentences with slightly different lexical representation while preserving the core meaning of the original sentence. We set English as an intermediate language. We choose a honorific option when translating back to Korean because the option tends to preserve the meaning of the sentences empirically. For less similar pairs, we first compute ROUGE \cite{lin2004rouge} of all possible sentence pairs, by assuming the higher score correlates with higher semantic similarity.\footnote{This might be replaced to any other similarity measures.} Then we draw a pair with the largest score from all possible pairs and the draw is repeated over remaining pairs until all of sentences are matched. As it progresses, the score declines as the number of remaining pairs becomes smaller, producing less similar pairs. We summarize this process as \textit{greedy sentence matching} (GSM), as presented in  Algorithm~\ref{sts:algo1}.

\begin{algorithm}[H]
\SetAlgoLined
\KwResult{Set of sentence pairs SET in a corpus C}
 Prepare corpus C, Let SET = []\;
 \While{size of C $\geq$ 2} {
  1. Choose a random sentence S from C\;
  2. Find a sentence T where ROUGE(S, T) is maximized and T $\in$ C\textbackslash\{S\}\;
  3. Remove \{S, T\} from C\;
  4. Add matched pair \{(S, T)\} to SET
  }
 \caption{Pseudocode of our greedy sentence matching (GSM) in AIRBNB and POLICY.}
 \label{sts:algo1}
\end{algorithm}

\paragraph{Annotation Protocol} 

We modify the original annotation guide used in SemEval-2015 \cite{agirre2015semeval}. It suggests chunking both sentences and compares similarity in chunk-level (e.g., NP, verb chain, PP, etc.). Then an annotator should sum up their judgement to sentence-level similarity. However, we could not directly apply the guide because chunking is highly challenging in Korean. In chunking, tokenization and morpheme-level decomposition of words are required, but they are difficult and even not deterministic in some cases \cite{park2020empirical}. 
We thus guide an annotator to evaluate the similarity without chunking and stick to sentence-level comparison. 

We give crowdworkers additional cues what is \textit{important} or \textit{unimportant} for sentence-level similarity evaluation.  \textit{Important} content indicates the main idea in a sentence. If it is a declarative sentence, its providing facts, explanation, or information is the main idea. For an interrogative and imperative sentence, conveying a request or command is important. In exclamatory sentence, feelings or opinion is the main content \cite{jen2000activity}. Other components than these \textit{important} contents are regarded as \textit{unimportant}. For example, they are auxiliary verbs or function words which affect its nuance or politeness. An annotator should score the similarity as follows:

\begin{itemize}[leftmargin=*]
    \item 5: Two sentences are equivalent in terms of \textit{important} and \textit{unimportant} content.
    \item 4: Two sentences are closely equivalent. Some \textit{unimportant} content differ.
    \item 3: Two sentences are roughly equivalent. \textit{Important} content are similar to each other, but difference between \textit{unimportant} content is not ignorable. 
    \item 2: Two sentences are not equivalent. \textit{Important} content are not similar to each other, only sharing some \textit{unimportant} contents.
    \item 1: Two sentences are not equivalent. \textit{Important} and \textit{unimportant} content are not similar to each other. Two sentences only share their topics.
    \item 0: Two sentences are not equivalent. They are not sharing any \textit{important} and \textit{unimportant} contents and even topics.
\end{itemize}

We also guide crowdworkers to consider the context of sentences. If it significantly affects distinguishing the meaning of two sentences, the score should be low. For example, let two sentences contain important information `check-in' such as ``Check-in was done by someone other than the host.'' and ``Check-in was done by someone.'' In the latter sentence, `someone' might be the host. Since we lose information by dropping `other than the host' from the former, difference of meaning between the two sentence is not ignorable. We score this pair to 3. Furthermore, if the former sentence is compared to `Check-out was done by someone other than the host.', \textit{important} information differ so we give score 2.

\paragraph{Annotation Process}
We recruit workers from SelectStar,\footnote{\url{https://selectstar.ai/}} a crowdsourcing platform in Korea and familiarize them to our annotation protocol. We run pilot annotation to select qualified workers. If a crowdworker's judgement is frequently disagreed against that of other workers, the person is excluded from the main annotation process. As a result, 19 out of the initial 20 workers participate in the main annotation. After removing the sentence pairs used in the pilot, we use 14,869 pairs for the main annotation, consisting of 7,375 for {\airbnb}, 2,956 for {\policy}, and 4,538 for {\parakqc}. 7 different workers labeled all sentence pairs independently. 

We average 7 labels for each sentence pair and remove outliers following \citet{agirre-etal-2016-semeval, cer-etal-2017-semeval}. First, we filter out annotators showing Pearson's correlation < 0.80 or Krippendorff’s alpha < 0.20 (nominal)  \cite{Krippendorff2011ComputingKA} with others' annotations. We exclude two annotators with this criteria so all sentence pairs have annotations from at least five people. Lastly, similarity score is rounded up to the first decimal place. 

A few more filtering schemes are applied. First, we drop 14 pairs whose annotations are showing larger than 2 standard deviation. Those pairs might contain ambiguous expressions interpreted in various ways, or misannotations. Second, we ask workers to report the sentences including translation error or misinformation caused by RTT. We inspect the reported sentences and remove 418 sentence pairs. Third, we drop sentences involving ethical issues. Workers report the pairs if they are including any kind of hate speech, social bias, and potential personally identifiable information (PII). 1,213 sentence pairs were additionally removed after inspection. As a result, we have 13,224 sentence pairs in total. We report inter-annotator agreement (IAA) by using Krippendorff’s alpha instead of Pearson's correlation because 7 annotators (or less) differ by pairs. The annotator agreed to each other's annotations. (Krippendorff’s alpha (interval) = 0.85).

\begin{figure}[ht] 
\centering
\includegraphics[width=\textwidth]{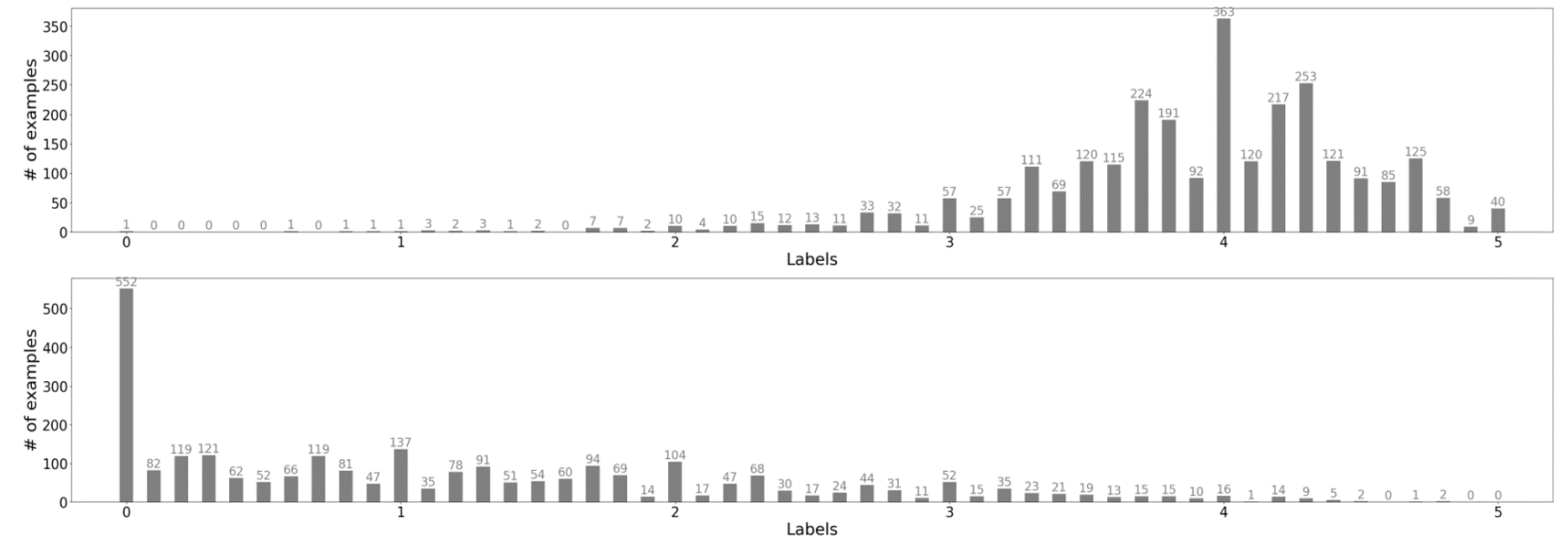}
\caption{Label distributions generated by RTT (top) and GSM (bottom) in {\airbnb}.}
\label{fig:RTT_GSM}
\end{figure}

We observe the distribution of similarity score annotations differ between the \textit{potentially} similar sentence pairs and the less similar pairs. Figure \ref{fig:RTT_GSM} illustrates label distributions generated by RTT (top) and GSM (bottom) in {\airbnb}. As expected, RTT pairs tend to show high similarity (from 3 to 5) while GSM pairs are considered less similar (from 0 to 3). Note that the number of GSM pairs scored 0 is high even we employ similarity-based matching. Similar tendencies are observed in {\policy} and {\parakqc}. By combining two distributions, we manage to obtain various sentence pairs in terms of similarity scores.

\begin{figure}[ht] 
\centering
\includegraphics[width=\textwidth]{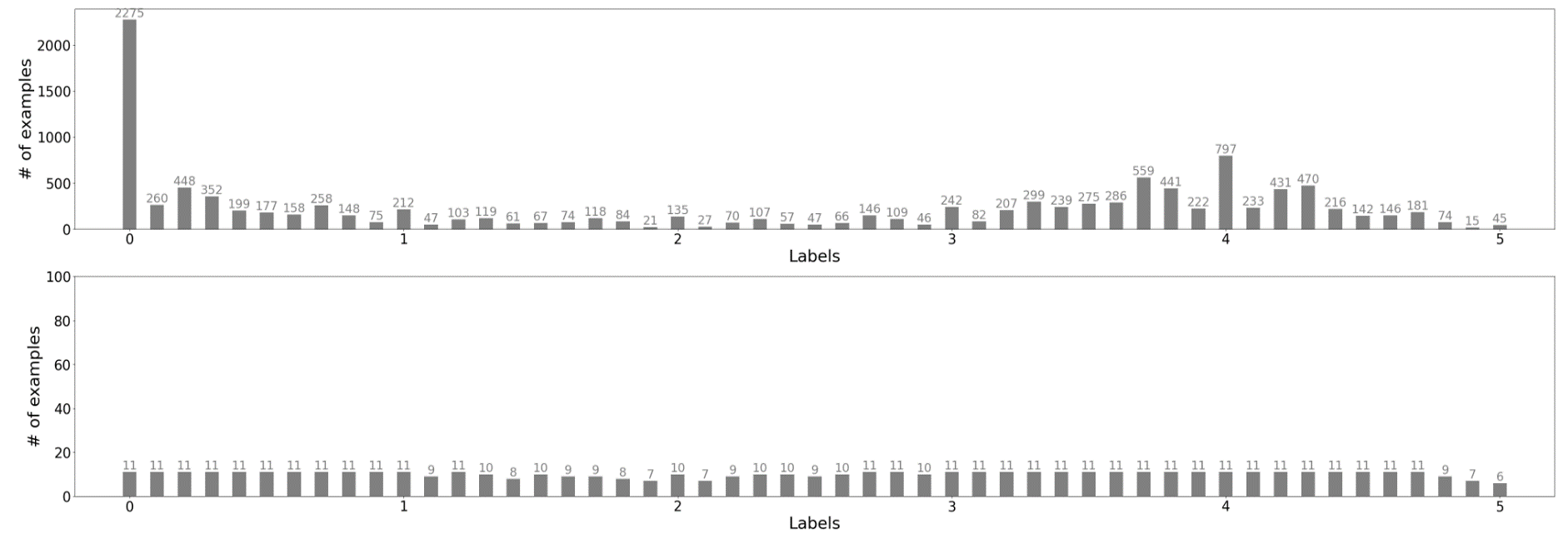}
\caption{Similarity score distribution of the train (top) and dev (bottom) set. The scores of dev set is close to uniform distribution across range 0$-$5. The scores are rounded to the first decimal place.}
\label{fig:sts_eval_dist}
\end{figure}

\paragraph{Final Dataset} We collect 13,224 sentence pairs and corresponding similarity scores. We split them to training, development, and test sets, considering the distribution of the scores. Even if we carefully sampled the pairs, the overall score distribution is not uniform across 0$-$5 as shown in Figure \ref{fig:RTT_GSM}. However, we prefer uniform distribution at least in evaluation (development and test) set, in order to prevent evaluation bias toward a specific score. We therefore construct the evaluation set having approximately uniform distribution as shown in Figure \ref{fig:sts_eval_dist}. To this end, we divide the score range 0$-$5 to 51 bins, rounding up to the first decimal place of every scores. We try to balance the number of pairs across bins. Since some of them have small number of pairs, we try to fit all number of the pairs close to that number. 

We also consider word overlap between sentences in each pair for evaluation set. Since larger word overlap might indicate higher semantic similarity, we try to reduce pairs satisfying such tendency to prevent the model from predicting similarity simply using word overlap. The overlap is measured by morpheme-level Jaccard distance by using MeCab \cite{kudo2006mecab}. We choose the pairs with the least word overlap from score 3$-$5, and the pairs with most word overlap from the rest. Such pairs are prioritized to be included to every bins in the dev and the test sets.
 
We split the evaluation set with 1:2 ratio to construct the dev and the test sets, resulting in 519 and 1,037 pairs, respectively. The rest 11,668 pairs comprise the train set. Detailed numbers for each corpus are presented in Table \ref{tab:sts_final_dataset}. For all the sets, we balance the ratio between source corpora with that of the original pairs. Additionally, the scores are binarized with a threshold 3.0 same as paraphrase detection task.

\begin{table}[ht!]
\centering
\caption{Statistics for {\kluests}. The first three columns provide the number of examples in train, dev, and test sets of each source corpus and the final data.}
\label{tab:sts_final_dataset}
\begin{tabular}{@{}lcccc@{}}
\toprule
\textbf{Source}           & \textbf{|Train|} & \textbf{|Dev|} & \textbf{|Test|} & \textbf{Total} \\ \midrule
\airbnb  & 5,371    & 255   & 510    & 6,136  \\
\policy    & 2,344    & 132   & 264    & 2,740  \\
\parakqc  & 3,953    & 132   & 263    & 4,348  \\ \midrule
\textbf{Overall}                   & \textbf{11,668}   & \textbf{519}  & \textbf{1,037}   & \textbf{13,224} \\ \bottomrule
\end{tabular}
\end{table}

\subsubsection{Evaluation Metrics}
\label{sec:sts-metric}

The evaluation metrics for \kluests{} is 1) Pearson's correlation coefficient (Pearson' $r$), and 2) F1 score. Pearson's $r$ is a measure of linear correlation between human-labeled sentence-similarity scores and model predicted scores, adopted in STS-b \cite{cer-etal-2017-semeval}. Since our dev and test set have a balanced score distribution, the coefficient correctly gives the magnitude of the relationship. F1 score is adopted to measure binarized results (\textit{paraphrased} / \textit{not paraphrased}). Specifically, our F1 reports results for the \textit{paraphrased} class. 

\subsubsection{Related Work}
Measuring similarity between sentences is a fundamental natural language understanding problem so that closely related to various NLP applications. Because of its importance, STS is included in various NLU benchmarks \cite{wang2018glue, xu2020clue}. To facilitate research in this area, many shared tasks have been held and annotated corpora are released \cite{agirre2015semeval,agirre-etal-2016-semeval,cer-etal-2017-semeval}. Typically, they cover multiple text domains such as question pairs, image descriptions, news headlines, annotated with a real value from 0 (no meaning overlap) to 5 (meaning equivalence).

Recently, \citet{ham-etal-2020-kornli} introduces a machine-translated Korean STS benchmark. This is a translation of \citep{cer-etal-2017-semeval} in GLUE, which contains around 8,600 sentence pairs in total. All examples are solely relying on machine translation, and sentence pairs in evaluation (dev and test) set are further post-edited by human. However, corresponding labels were not adjusted to translated meanings. Lack of re-labeling process would be problematic because Korean speakers would judge the similarity between them differently.

If similarity labels are binarized by a certain threshold, STS also could be seen as paraphrase detection task such as Microsoft Research Paraphrase Corpus (MRPC) \cite{dolan2005automatically}, Quora Question Pairs (QQP) \cite{wang2018glue}, or PAWS \cite{zhang-etal-2019-paws} and PAWS-X \cite{yang2019paws}. Thus we additionally binarize our ground truths and predictions, reporting binary classification performance to see how well a model performs in paraphrase detection.

In paraphrase detection, \citet{cho2020discourse} presents a benchmark that includes the human-generated queries for smart home, where ten paraphrase sentences are grouped together to make up a total of 1,000 groups. The granularity of scale is from 0 to 5, but the semantic similarity is judged only with attributes such as topic (smart home, weather, etc.) and speech act (question, prohibition, etc.), which does not consider other details such as nuance and syntactic structure because it lacks direct human judgement of similarity. PAWS-X \cite{yang2019paws} provides a translated version of PAWS \cite{zhang-etal-2019-paws} of Korean. Like KorSTS, the train split is machine-translated and its dev and test splits are human-translated, and corresponding labels are preserved without human inspection. There are also paraphrase corpora provided by government-funded institutions such as National Institute of Korean Language (NIKL) \cite{nikl2020corpora}, but it simply provides human-generated and machine-paraphrased sentences with limited accessibility.

\subsubsection{Conclusion}
We create the first human-annotated Korean STS benchmark, \kluests\index{KLUE-STS}, that covers multiple domains and styles with free accessibility to everyone. The similarity score annotation process is specially designed to capture the characteristics of the Korean language. Covering the expressions from various domains, our benchmark is expected to be a useful resource for further research, beyond serving as a benchmark. Our benchmark helps to develop numerous models established on STS resources, such as SentenceBERT \cite{reimers2019sentence}.

\clearpage
\subsection{Natural Language Inference (\nli)}
The goal of natural language inference (NLI)\index{NLI (Natural Language Inference)} is to train a model to infer the relationship between the \textit{hypothesis} sentence and the \textit{premise} sentence. Given a \textit{premise}, an NLI model determines if \textit{hypothesis} is true (entailment), false (contradiction), or undetermined (neutral). The task is also known as recognizing textual entailment (RTE) \cite{dagan2005pascal}.

Understanding entailment and contradiction between sentences is fundamental to NLU. NLI datasets are also included in various NLU benchmarks such as GLUE~\cite{wang2018glue} and superGLUE \cite{wang2019superglue}, and they are valuable as training data for other NLU tasks~\cite{conneau2017supervised, phang2018sentence, reimers2019sentence}. 

We formulate NLI as a classification task where an NLI model reads each pair of \textit{premise} and \textit{hypothesis} sentences and predicts whether the relationship is entailment, contradiction, or neutral. We use the classification accuracy to measure the model performance.

\subsubsection{Dataset Construction}

We construct {\kluenli} by using a collection method similar to that of SNLI \cite{bowman2015large} and MNLI \cite{williams2017broad}. First, we collect premise sentences from existing corpora. Then for each premise sentence, we ask one annotator to generate three new hypothesis sentences, one for each of the three relationship classes. Then for each pair of premise and hypothesis sentences, we ask four additional annotators to label the relationship for validation. 
We follow the criteria proposed by \citet{williams2017broad} to describe the three labels to the annotators. For both hypothesis generation and pair validation, we recruit workers from SelectStar,\footnote{\url{https://selectstar.ai/}} a Korean crowdsourcing platform.

\paragraph{Source Corpora for Premise Sentences}


We use six corpora for the set of premise sentences: \wikitree, \policy, \wikinews, \wikipedia, {\nsmc} and {\airbnb}. They cover diverse topics and writing styles of contemporary Korean. \wikitree, \policy{} and {\wikinews} are news articles and {\wikipedia} is a crowd-sourced encyclopedia, all of which are written in formal Korean. {\nsmc} and {\airbnb} consist of colloquial reviews in the domains of movies and travel, respectively.  

From the six corpora, we extract 10,000 premises with which we elicit hypotheses. A valid premise should satisfy three conditions. First, premise is a proposition, a declarative sentence to which we can assign a truth value, excluding mathematical formulae and lists. Second, a premise must include at least one predicate, and the predicate can be of diverse types such as states (e.g., be, believe, know), activities (e.g., play, smile, walk), achievements (e.g. realize, reach, break), and accomplishments (e.g. eat, build, paint). Third, the length of a premise should be from 20 to 90 characters including whitespace.

\paragraph{Annotation Protocol for Hypothesis Generation}
We show annotators a premise and ask them to write three hypotheses that correspond to each label. This allows us to collect nearly equal number of the (premise, hypothesis) pairs for each labels. We maintain the outline of the criteria as follows: 
\begin{itemize}[leftmargin=*]
    \item ENTAILMENT: The hypothesis is necessarily true given the premise is true 
    \item CONTRADICTION: The hypothesis is necessarily false given the premise is true 
    \item NEUTRAL: The hypothesis may or may not be true given the premise is true
\end{itemize}

We are aware of the annotation artifacts coming from human writing-based hypothesis generation. Sentence length and explicit lexical patterns are highly associated with certain classes. Neutral sentences tend to be the longest among all classes, since workers can produce neutral hypothesis simply by introducing additional phrase or clause not stated in the premise. Negations such as ``no'', ``never'' and ``nothing'' are often accompanied with the class CONTRADICTION \cite{gururangan2018annotation, poliak-etal-2018-hypothesis}. 

Despite the concerns of such artifacts, we stick to such a writing-based annotation procedure.
Compared to automatic pipelines to collect hypotheses, human writing yields higher quality data and is still an effective protocol \cite{vania-etal-2020-asking}. 
We focus on ways to encourage annotators to avoid injecting trivial patterns. We prepare guidelines with specific \textit{Do}s and \textit{Don't}s, and rigorously train the workers in advance. 
To minimize annotation artifacts, we instruct the annotators to write sentences with similar lengths across the classes, refrain from inserting certain lexical items repeatedly, and use as diverse strategies as possible when making inferences.  

Specifically, we provide detailed guidelines for hypothesis generation together with examples. We encourage annotators to create hypotheses that exhibit diverse linguistic phenomena, in terms of 1) lexical choice, 2) syntactic structures and 3) world knowledge. In the case of lexical choice, our guideline suggests annotators use synonyms/antonyms, hypernyms/hyponyms, and auxiliary particles. To introduce various syntactic structures, we provide several syntactic transformation strategies such as word scrambling, voice alteration, and causative alternation. Methods like subject/object swapping or passivization is motivated by existing NLI data augmentation strategies \cite{min-etal-2020-syntactic, glockner-etal-2018-breaking}. We also encourage using expressions that reflect world knowledge such as time, quantity and geography in order to create a dataset grounded to the real world. 

There are a few more details in the guideline. We instruct annotators to maintain the writing style of the premise to create a balanced dataset in terms of the style as well. We also instruct them to skip sentences that are difficult to understand either due to the ungrammaticality or the complexity of the content. They are also instructed to skip and report sentences that contain ethical issues such as hate speech, social bias, or personally identifiable information. We examine all reported sentences and make final decisions whether to include the sentences in the dataset.

\paragraph{Annotation Protocol for Label Validation}
Crowdworkers annotate the relations of the resulting premise-hypothesis pairs for validation. For each of the pairs created, we ask four crowdworkers to supply a single label among (ENTAILMENT, CONTRADICTION, NEUTRAL). This yields a total of five labels per pair, including the initial label intended by the annotator who wrote the hypothesis sentence. For each validated sentence pair, we assign a gold label representing the majority of three or more votes out of five.

\paragraph{Annotation Process} 
For hypothesis generation, we go through a pilot phase where we iteratively update the guidelines and train the workers. During the pilot, we find writing a semantically unacceptable sentence or introducing a demonstrative pronoun not used in the premise could be potential problems. Since they might alter the intended label, we ask workers to avoid writing such sentences. The number of workers for this part of the annotation process is 11.

We then validate the relation labels for every pair. We go through a pilot phase, starting with 2,604 applicants in the pilot, then select 684 who passed the test to participate in the validation step. With 138 workers dropping out, the final number of workers is 546.

Validation results are summarized in Table \ref{tab:nli_validation_statistics}. They suggest that our writing protocol is effective in producing a high quality corpus. The rate of unanimous gold labeled examples in {\kluenli} is 18\% higher than  SNLI and MNLI. The higher the rate of such examples, the clearer the relationship between the generated hypothesis sentences and the original premise sentences. Individual annotator's agreement with the gold label and the author's label are also higher than SNLI and MNLI, and almost all pairs receive the gold label. Only a few sentence pairs (0.53\%) lack the gold label, and we remove those before finalizing our dataset.

\begin{table}[t!]
\normalsize
\centering
\caption{Summary of validation statistics for {\kluenli} compared to SNLI and MNLI \cite{williams2017broad}. We call the label intended by the original annotator in writing the hypothesis "author's label." Consensus among three out of five annotators is "gold label."}
\label{tab:nli_validation_statistics}
\begin{tabular}{@{}lccc@{}}
\toprule
\textbf{Statistics}                         & \textbf{SNLI}    & \textbf{MNLI}    & \textbf{\kluenli} \\ \midrule
Unanimous Gold Label     & 58.30\% & 58.20\% & \textbf{76.29\%}     \\ \midrule
Individual Label = Gold Label     & 89.00\% & 88.70\% & \textbf{92.63\%}     \\
Individual Label = Author’s Label & 85.80\% & 85.20\% & \textbf{90.92\%}     \\ \midrule
Gold Label = Author’s Label       & 91.20\% & 92.60\% & \textbf{96.76\%}     \\
Gold Label $\neq$  Author’s Label & 6.80\%  & 5.60\%  & \textbf{2.71\%}      \\
No Gold Label (No 3 Labels Match) & 2.00\%  & 1.80\%  & \textbf{0.53\%}      \\ \bottomrule
\end{tabular}
\end{table}

\paragraph{Final Dataset} The final dataset consists of 30,998 sentence pairs that are divided into train/development/test sets. Table~\ref{tab:nli_final_dataset} shows the basic statistics of the dataset. As observed in SNLI and MNLI, our premise sentences also tend to be longer than the corresponding hypothesis sentences. This is because workers generally use partial information of a premise to write a hypothesis. 

Note that we deliberately form the development and test sets in a way to 1) contain balanced source styles and 2) disincentivize models exploiting annotation artifacts. The development and the test set each contains 3,000 sentence pairs.

\begin{table}[ht!]
\normalsize
\centering
\caption{Statistics for {\kluenli}. The first three columns provide the number of sentence pairs in train, dev, and test sets. \textit{Avg Len Prem} and \textit{Avg Len Hyp} are the mean character counts of premise and hypothesis sentences, respectively.}
\begin{tabular}{@{}lcccccc@{}}
\toprule
\textbf{Source}           & \textbf{|Train|} & \textbf{|Dev|} & \textbf{|Test|} & \textbf{Total} & \textbf{Avg Len Prem} & \textbf{Avg Len Hyp} \\ \midrule
\wikitree  & 3,838    & 450   & 450    & 4,738  & 52.81             & 26.86                     \\
\policy    & 3,833    & 450   & 450    & 4,733  & 56.73             & 32.93                     \\
\wikinews  & 3,824    & 450   & 450    & 4,724  & 64.17             & 29.11                     \\
\wikipedia & 3,780    & 450   & 450    & 4,680  & 57.45             & 23.70                     \\ \midrule
\nsmc      & 4,899    & 600   & 600    & 6,099  & 27.48             & 21.49                     \\
\airbnb    & 4,824    & 600   & 600    & 6,024  & 24.28             & 18.65                     \\ \midrule
\textbf{Overall}                   & \textbf{24,998}   & \textbf{3,000}  & \textbf{3,000}   & \textbf{30,998} & \textbf{47.15}             & \textbf{25.46}                     \\ \bottomrule
\end{tabular}
\label{tab:nli_final_dataset}
\end{table}

To maintain consistency of style in development and test sets, we include in each set 60\% formal and 40\% colloquial sentences. We sample 450 sentences each from formal text {\wikitree}, {\policy}, {\wikinews}, {\wikipedia}, and 600 sentences each from colloquial text {\nsmc}, {\airbnb}.

To prevent our NLI benchmark from incentivizing a model that predicts a label using a spurious cue in the hypothesis, we first fine-tune the KLUE-RoBERTa-base model using only the hypothesis sentences with their corresponding labels. If the model finds no clue between the hypothesis and the label, the predicted probability scores for each label should be uniform (i.e., one-third ($\frac{1}{3}$) when classified 3-way). Assuming that such score distribution is ideal, we prefer the pairs for development/test sets whose hypothesis-only model's predictions are closest to the ideal. We compute the distance between the prediction and the ideal using cross entropy. To preserve the intact sets of a premise and its three hypotheses, we calculate the mean distance of each set. We extract the sets whose mean distance is among the lowest 20\%, and randomly split them into dev and test sets.

Our idea can be viewed as an extension of pointwise mutual information (PMI). PMI between each hypothesis word ($w$) and class label ($c$) has been used to discover the association of the word with each class \cite{gururangan2018annotation, vania-etal-2020-asking}. If PMI is expanded to the sentence-level association, the metric provides a similar measure to the hypothesis-only model prediction probability as below.
$$
\text{PMI}(w, c) = \log{\frac{P(w,c)}{P(w) P(c)}}  = \log{\frac{P(c|w)P(w)}{P(w)P(c)}} = \log{\frac{P(c|w)}{P(c)}} \propto P(c|w)
$$

To measure human performance and examine whether {\kluenli} test set improves upon KorNLI \cite{ham-etal-2020-kornli} test set, a machine-translation of the XNLI \cite{conneau2018xnli} test set, we conduct a round of human evaluation. We employ four native Korean undergraduates who major in Korean linguistics and did not participate in the {\kluenli} construction process. We randomly sample 100 sentence pairs from {\kluenli} test set and ask the workers to annotate them. We check the agreement of their annotations with the given gold label. We do the same on the subset of the KorNLI test set, to examine whether the human-elicited dataset improves the quality of the dataset. The results are shown in Table~\ref{tab:nli_kor_diff}.

For KorNLI, 38\% of the sentence pairs have responses from all four annotators that match with the gold labels. There are 18\%, 18\%, and 16\% of sentences, respectively, when three, and two, and one response match with the gold label. 10 pairs do not match with the gold label. On the other hand, {\kluenli} shows much higher agreement with the given gold label. All annotators agree with the gold label in 71\% of the pairs, and 95\% obtain at least three agreements. Furthermore, only 258 out of 400 (64.50\%) individual annotations are the same as the gold label in KorNLI. Again, {\kluenli} shows better agreement with the gold labels. 360 (91.00\%) annotations are the same as the gold label.

These numbers in annotation quality of {\kluenli} are better than KorNLI as well as SNLI and MLNI. In KorNLI, annotators often report that they do not quite understand at least one of the two sentences or choose NEUTRAL because it is difficult to distinguish the semantic relationships of the sentences. Although the distribution of the gold label is uniform (respectively 33, 33, and 34\% of entailment, contradiction, and neutral sentences), the label chosen most frequently by the annotators is NEUTRAL (56.75\% on average). There are 26\% of cases where the gold labels are different from the majority vote by the annotator. These results suggest that the annotators struggle to grasp the logical semantic relationship of KorNLI sentences.

\begin{table}[ht!]
\centering
\caption{Statistics for human evaluation results of KorNLI and {\kluenli}. We compare the labels of four annotators with gold labels of korNLI and {\kluenli} test data.}
\begin{tabular}{@{}lcc@{}}
\toprule
\textbf{Statistics} & \textbf{KorNLI} & \textbf{\kluenli}\\ \midrule
Unanimous Gold Label (4 Agree)    & 38.00\% & \textbf{71.00\%} \\
3 Agree with Gold Label	& 18.00\% & 24.00\% \\
2 Agree with Gold Label	& 18.00\% & 3.00\% \\ 
1 Agrees with Gold Label	& 16.00\% & 2.00\% \\ 
0 Agrees with Gold Label	& 10.00\% & 0.00\% \\ \midrule
Individual Label = Gold Label     & 64.50\% & \textbf{91.00\%} \\ \midrule
No Gold Label (No 3 Labels Match) & 4.00\%  & \textbf{0.00\%} \\
Majority Vote $\neq$ Gold Label & 26.00\%  & \textbf{0.00\%} \\ \bottomrule
\end{tabular}
\label{tab:nli_kor_diff}
\end{table}

On the other hand, for {\kluenli}, there is no case where none of the four responses matches the gold label. Considering the cases where more than two of the responses match the gold label, there is a 98\% chance of the gold label to be re-selected as the majority tag. Compared to KorNLI, we can see that {\kluenli} is a much more reliable dataset. This result also confirms that the headroom of our current best model (accuracy: 89.77\%) is still there, given that the human accuracy, represented by the majority tag, is 98\%.

\subsubsection{Evaluation Metric}
\label{sec:nli-metric}
The evaluation metric for \kluenli{} is accuracy, following SNLI \cite{bowman2015large} and MNLI \cite{williams2017broad}. Accuracy measures how well a classifier correctly identifies the results. The class labels are almost equally distributed, thus higher accuracy will correctly represent performances of a model.

\subsubsection{Related Work}


Recognizing Textual Entailment (RTE)~\cite{dagan2005pascal} is a task similar to NLI and was introduced in a series of textual entailment challenges. In the RTE task, two sentences are given, and the model decides whether the meaning of one sentence can be entailed from the other sentence. In earlier RTE 1–3, the task is binary, `ENTAILMENT' and `NO ENTAILMENT'. In RTE 4-5, a new class `UNKNOWN' is introduced, and the task is formulated as a three-way classification.

Two major datasets for NLI in English are Stanford Natural Language Inference (SNLI)~\cite{bowman2015large} and Multi-Genre Natural Language Inference (MNLI)~\cite{williams2017broad}. Hypothesis sentences in SNLI and MNLI are labeled ENTAILMENT, CONTRADICTION, or NEUTRAL. SNLI is two orders of magnitude larger than the RTE corpora, made from 570,152 image captions in Flickr30k ~\cite{young2014image}. MNLI premise sentences are derived from 10 different sources, covering a wider range of styles, degrees of formality, and topics.

Most of the existing NLI datasets are in English, including SNLI and MNLI, and one common approach for constructing NLI datasets in other languages is to translate the existing English corpora to the language of interest. \citet{conneau2018xnli} provides XNLI (Cross-lingual natural language inference) by employing professional translators to translate the development and test sets of MNLI into 15 languages. One main concern of the translation-based approach is whether the relation of the original sentence pair is maintained in the process. \citet{conneau2018xnli} find some translated pairs lose the initial semantic relationship, validated by human annotators who re-annotate a sample of the dataset. The result demonstrates that human translations cause 2\% misannotations given the 85\% correct examples in the MNLI and 83\% in XNLI.

Motivated by the fact that Korean is not included in XNLI, KorNLI \cite{ham-etal-2020-kornli} is introduced. KorNLI~\cite{ham-etal-2020-kornli} is a translation of existing English corpora whose train set is created through machine translation of training sets of SNLI and MNLI, and the development and test sets through machine translation of development and tests sets of XNLI and post-editing by professional translators. Although \citet{ham-etal-2020-kornli} also investigate the data manually and acknowledge some incorrect examples after the translation, no human validation process is performed to quantify the observation and leave analyzing such errors to future work. Moreover, even with post-editing, there are some sentences that are either unnatural in terms of syntactic structure or word choice.

Many studies have been proposed based on SNLI and MNLI; however, SNLI and MNLI are known to have annotation artifacts \cite{gururangan2018annotation, poliak-etal-2018-hypothesis}. Annotation artifacts are the product of certain types of annotation strategies and heuristics naturally arising from the crowdsourcing process. Such artifacts are problematic as they may lead models to adopt heuristics rather than to actually learn the relationship. 

There have been some efforts to reduce annotation artifacts in NLI. \citet{vania-etal-2020-asking} experiment with two fully automated protocols for creating premise-hypothesis pairs, but find that the methods yield poor-quality data and mixed results on annotation artifacts. OCNLI~\cite{hu-etal-2020-ocnli} enhance writing-base protocol with some interventions to control the bias: encouraging writers to use diverse ways of making inference, and putting constraints on overused words. Despite partial effects on reducing negators, the explicit constraint gives rise to other words of correlation, and the final OCNLI dataset exhibit similar level of hypothesis-only test scores to most benchmark NLI datasets.

\subsubsection{Conclusion}
Our new dataset, {\kluenli}\index{KLUE-NLI}, is the first resource constructed upon naturally occurring Korean sentences. {\kluenli} represents diverse linguistic phenomena, writing style, degree of formality and contents that are most natural and suitable for Korean. The premise sentences of our dataset come from six Korean corpora, and the hypothesis sentences are written by well-trained workers. 

By keeping the writing-based protocol and thoroughly training workers based on detailed guidelines, we improve upon the existing NLI datasets in the reliability of the labels. {\kluenli} shows much higher inter-annotator agreement rate than both the MNLI and the translation-based Korean dataset, KorNLI. The gap between the human performance scores of {\kluenli} and KorNLI also provides evidence that {\kluenli} is currently the optimal Korean NLI dataset.

Beyond its main purpose as an NLI benchmark dataset, we hope {\kluenli} will be a useful resource for future NLU research, as English dataset such as MNLI and SNLI are extended~\cite{conneau2017supervised, phang2018sentence, reimers2019sentence}. 

\clearpage
\subsection{Named Entity Recognition (\ner)}
The goal of named entity recognition (NER)\index{NER (Named Entity Recognition)} is to detect the boundaries of named entities in unstructured text and classify the types. An entity can be series of words that refers to the person, location, organization, time expressions, quantities, monetary values.

Since NER is an important for application fields like syntax analysis, goal-oriented dialog system, question and answering chatbot and information extraction, various NLU benchmarks contains NER datasets \cite{wilie2020indonlu, kakwani2020indicnlpsuite, liang2020xglue, hu2020xtreme}. Despite the rise of necessity of NER datasets in various domains and styles, there are few existing Korean NER datasets to cover such need. Therefore, we annotate corpora including web texts that can be applied to real-word applications.

In {\kluener}, a model should detect the spans and classify the types of entities included in an input sentence. The six entity types used in {\kluener} are person, location, organization, date, time, and quantity. They are tagged via character-level BIO (Begin-Inside-Outside) tagging scheme, and thus we evaluate a model’s performance using entity-level and character-level F1 score.

\subsubsection{Dataset Construction}

\paragraph{Source Corpora}

To incorporate both formal and informal writing styles, we use two corpora, {\wikitree} and {\nsmc} for annotation. {\wikitree} is a news article corpus and thus contains formal sentences with many entity types, which suits well as a source corpus for NER. {\nsmc} includes colloquial reviews of  movies or TV shows. Since the texts in {\nsmc} are user-generated comments, they contain errata and non-normalized expressions, along with emojis and slang. Such a noisy dataset will help broaden the application field of NER models.

The preprocessing of the two corpora is performed differently considering the characteristics of each corpus. For {\wikitree}, since the news articles are mainly composed of well-written sentences, we simply split the articles into sentences. In contrast, the web texts from {\nsmc} are written in the style of spoken language with blurry sentence boundaries. As each review is generally quite short and the sentences consisting it are on the same topic, we use each review as a single unit of input. In addition, the sentences that contain hate speech or socially biased terms are removed manually. For both corpora, we remove sentences longer than 400 characters. 

For efficient annotation, we perform pseudo-labeling with a pretrained model. The model is trained with BERT-CRF using a publicly available dataset KMOU-NER corpus,\footnote{\url{https://github.com/kmounlp/NER}} to support fast and accurate entity tagging for annotators. We also filter out the sentences with no pseudo-labeled entity assuming they do not include any of the entities. Remaining sentences account for about 80\% in {\wikitree} and 41\% in {\nsmc}, leaving a total of 36,515 sentences.

\paragraph{Annotation Protocol}
We use six entity types for {\kluener} annotation: PS (Person), LC (Location), OG (Organization), DT (Date), TI (Time), and QT (Quantity). The description of each entity type is as follows.

\begin{itemize}[leftmargin=*,noitemsep]
    \item PS (Person): Name of an individual or a group
    \item LC (Location): Name of a district/province or a geographical location
    \item OG (Organization): Name of an organization or an enterprise
    \item DT (Date): Expressions related to date/period/era/age
    \item TI (Time): Expressions related to time
    \item QT (Quantity): Expressions related to quantity or number including units
\end{itemize}

We employ the above sets following the convention of two existing tag sets: Korean Telecommunications Technology Association (TTA) NER guidelines\footnote{
\url{https://committee.tta.or.kr/data/standard_view.jsp?nowPage=2&pk_num=TTAK.KO-10.0852&commit_code=PG606}
} and  MUC-7 \cite{chinchor-1998-overview}. TTA guideline is a standardized NER tagging scheme for Korean language and we follow the names and the definitions of its entity types. Among the 15 entity types of TTA, we select our six types that correspond with tagsets used in MUC-7 (DATE, LOCATION, MONEY, ORGANIZATION, PERCENT, PERSON and TIME). As MONEY and PERCENT types are included in QT (QUANTITY) type from TTA set, we instead adopt an entity type QT.  

In the case of entities with multiple possible entity types, instead of assigning a unique tag for all use cases, we determine their tags based on the context. One example is \textit{Cine21}, which, in Korean, can either refer to the name of a magazine or the publisher of the magazine. In a sentence like ``'I bought a Cine21 from a bookstore and read it page by page,'' `Buy something from a bookstore' and `read page by page' are properties regarding media (magazine), rather than an organization; thus we do not assign an OG tag. 

We guide crowdworkers to report if the text for annotation does not meet certain conditions. For example, texts consisting of multiple sentences, texts that are not in a sentence form, a fragment, and a simple sequence of nouns are discarded. Workers are also required to report sentences that include hate speech and various biases in tagging process. 

In terms of personally identifiable information, we cannot simply drop or pseudonymize the information because the very task of NER often requires the specific information of proper nouns such as person names (PS). In order to minimize the loss of sentences, we inspect through the sentences after the annotation process. We investigate the sentences that include PS tags, and keep the ones that contain the name of public figures that appear in Korean search engines.\footnote{Daum: \url{http://search.daum.net/search?nil_suggest=btn&nil_ch=&rtupcoll=&w=tot&m=&f=&lpp=&q=\%C0\%CE\%B9\%B0\%B0\%CB\%BB\%F6} / Naver: \url{https://people.search.naver.com/}} Other sentences are removed if it has potential privacy issues. 

\paragraph{Annotation Process}
51 qualified crowdworkers recruited by a Korean crowdsourcing platform, DeepNatural\footnote{\url{https://deepnatural.ai/}} participate in the annotation process. The qualification is given when passing a pilot entity tagging test. Then two linguists check whether the crowdworkers' annotations are correct or not. We find some erroneous annotations remaining even after validation. Therefore, six NLP researchers manually correct the annotation errors.

During the annotation process, 5,354 sentences are dropped by workers due to their inadequacy. 118 sentences are dropped due to the privacy issue, and 35 sentences are removed after the inspection by the researchers because all annotations are false positives. A total of 5,507 sentences are dropped in the inspection process, resulting in 31,008 sentences.

\begin{table}[t!]
\normalsize
\centering
\caption{Statistics for {\kluener}.}
\begin{tabular}{@{}lcccc@{}}
\toprule
\textbf{Source}           & \textbf{|Train|} & \textbf{|Dev|} & \textbf{|Test|} & \textbf{Total} \\ \midrule
\wikitree  & 11,435    & 2,534   & 2,685    & 16,664  \\
\nsmc  & 9,573    & 2,466   & 2,315    & 14,354  \\ \midrule
\textbf{Total}                   & \textbf{21,008}   & \textbf{5,000}  & \textbf{5,000}   & \textbf{31,008} \\ \bottomrule
\end{tabular}
\label{tab:ner_final_dataset}
\end{table}

\begin{table}[t!]
\centering
\caption{Entity-wise statistics for {\kluener}. Note that the numbers in parentheses denote the number of types. The total number does not match Table~\ref{tab:ner_final_dataset} since this table does not remove duplication.}
\begin{tabular}{@{}lcccc@{}}
\toprule
\textbf{Source}           & \textbf{|Train|} & \textbf{|Dev|} & \textbf{|Test|} & \textbf{Total} \\ \midrule
PS    & 14,453 (5,428)  & 4,418 (2,706)  & 4,830 (3,063)  & 23,289 (7,124)  \\
LC    & 6,663 (2,068)   & 1,649 (896)    & 2,064 (1,130)  & 9,961 (2,650)   \\
OG    & 8,491 (3,008)   & 2,182 (1,291)  & 2,514 (1,579)  & 12,855 (3,796)  \\
DT    & 8,029 (1,608)   & 2,312 (835)    & 2,498 (933)    & 12,653 (2,060)  \\
TI    & 2,020 (573)     & 5,45 (268)     & 579 (316)      & 3,110 (730)     \\
QT    & 11,717 (3,628)  & 3,151 (1,763)  & 3,827 (2,369)  & 18,019 (4,776)  \\ \midrule
\textbf{Total}  & \textbf{51,373 (16,313)} & \textbf{14,257 (7,759)} & \textbf{16,312 (9,390)} & \textbf{79,887 (21,136)} \\ \bottomrule
\end{tabular}
\label{tab:ner_entity_dataset}
\end{table}

\paragraph{Final Dataset}

The resulting corpus is split into train/dev/test sets, each consisting of 21,008, 5,000, and 5,000 sentences (Table~\ref{tab:ner_final_dataset}). The entity-wise statistics is provided in Table~\ref{tab:ner_entity_dataset}. We design the test set to include unseen entities to check the robustness of the models in terms of domain transitions and generalization.

\begin{figure}[h] 
\centering
\includegraphics[width=0.95\textwidth]{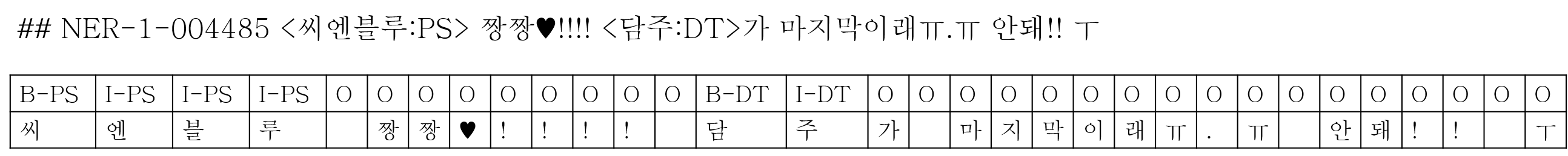}
\caption{An example of BIO scheme for NER tagging. The sentence is translated as: ``<CNBlue:PS> is the best♥!!!! So sad <the next week:DT> is their last weekT.T Nooooo!!'' where 씨엔블루 (\textit{CNBlue}) is a rock band of Korea. 담주 (\textit{the next week}) is tagged as DT here, while it is agglutinated with a functional word  가 (\textit{is}) in this sentence and is separately annotated with the character-level BIO scheme.}
\label{fig:ner_bio}
\end{figure}

The finalized entity types are tagged in the character level BIO tagging scheme (Figure~\ref{fig:ner_bio}). In most English and Korean NER datasets, the entities are tagged with the word-level BIO scheme, following CoNLL 2003 dataset \cite{tjong-kim-sang-de-meulder-2003-introduction}. In Korean, however, it is difficult to adhere to the word level tagging scheme based on whitespace for two reasons. First, whitespace-split units (eojeols) are often not a single word and are a composite of content words and functional words (e.g., ‘담주가 (the next week is)’ = ‘담주 (the next week)’ + ‘가 (is)’) \cite{han2020annotation}. Second, many compound words in Korean contain whitespaces. Therefore, we choose to tag in character level.

\subsubsection{Evaluation Metrics}
\label{sec:ner-metric}

The evaluation metrics for \kluener{} are 1) entity-level macro F1 (Entity F1) and 2) character-level macro F1 (Char F1) scores. Entity F1 score measures how many predicted entities and types are exactly matched with the ground truths in entity-level. Suppose a ground truth is [B-PS, I-PS, O, O, B-OG, I-OG] and a prediction is [B-PS, I-PS, I-PS, O, B-OG, I-OG]. For entity type PS, F1 score is 0 since a model fails to predict the exact span, while in case of OG, a model gets a score. To get a high score, a model should be careful at tokenization. Char F1 score is newly provided to measure a partial overlap between a model prediction and a ground truth. We additionally report this measure to see how well a model decomposes stems and affixes in Korean, which significantly affects the model performance of NER. Char F1 is an average of class-wise F1-scores. In \kluener{}, the classes are B-PS, I-PS, B-LC, I-LC, B-OG, I-OG, B-DT, I-DT, B-TI, I-TI, B-QT, and I-QT. We exclude the majority negative entity class (O), to focus on positive entities.

\subsubsection{Related Work}

CoNLL2003~\cite{tjong-kim-sang-de-meulder-2003-introduction} is the most widely used NER benchmark which covers texts from Reuters newswire articles. It handles English and German and is annotated with four named entity types (persons, locations, organizations, and miscellaneous entities). Another dataset on news articles from the Wall Street Journal, MUC (Message Understanding Conference)~\cite{grishman1996message}, presents an extended tag set, including temporal and numerical entities. The resulting materials, e.g., MUC-6~\cite{grishman1996message} and MUC-7~\cite{chinchor-1998-overview}, which include six and seven classes of entities, respectively, are adopted as a training source for developing the Stanford NER parser~\cite{finkel-etal-2005-incorporating}. 

To handle more informal and less sentence-like documents, WNUT16 \cite{strauss-etal-2016-results} is proposed. It deals with English Twitter texts which are first suggested in TwitterNER \cite{ritter-etal-2011-named}. A total of 15 types of entities are labeled, more subdivided than CoNLL03 and MUC.

For Korean, there are four existing NER datasets which are published by Korea Maritime \& Ocean University (KMOU), Changwon University, National Institute of Korean Language (NIKL) and Electronics and Telecommunications Research Institute (ETRI). All of them follow the tagging schema of Telecommunications Technology Association (TTA).\footnote{KMOU utilizes a modified guideline KMOU-NLP-2018-001 based on the TTA scheme, which is available in \url{https://github.com/kmounlp/NER/blob/master/NER\%20Guideline\%20(ver\%201.0).pdf}} TTA provides a standardized named entity tagging scheme that serves as an integrated guideline for NER research in Korean. It incorporates 15 named entity tags with 146 subcategories, and provides the definition and the examples regarding each tag with the instructions on the tagging procedure. 

No existing Korean NER dataset is both freely accessible and covers diverse text domains. According to \citet{cho-etal-2020-open}, the NER dataset provided by Korea Maritime \& Ocean University (KMOU) and the dataset constructed by Changwon University are publicly accessible. The datasets provided by ETRI and NIKL are not fully public, and the usage is also restricted to domestic researchers. We overcome this issue by making KLUE NER freely available to anyone. None of the aforementioned datasets cover sentences from noisy user generated web texts, which helps model trained on those to be more robust and generalizable. Moreover, except for the KMOU dataset, all the above datasets are tagged in word level, which often conflicts with the morphological characteristics of Korean. In comparison, {\kluener} uses web texts as source corpora and the entities are annotated in character level, thus being more practical and useful.

\subsubsection{Conclusion}
We construct a new Korean NER benchmark that covers broad domains and styles, which is freely accessible to anyone. The entity types are annotated so that a model has to use both morphological and contextual cues. The character-level entity tagging and evaluation method reflects the characteristics of Korean morphology. Since {\kluener}\index{KLUE-NER} dataset covers both formal news articles and informal user-generated web texts, we hope that our benchmark helps develop NER models that can be used in a wide a range of domains, and serve as a resource for developing advanced models for Information Extraction.

\clearpage
\subsection{Relation Extraction (\re)}
Relation extraction (RE)\index{RE (Relation Extraction)} identifies semantic relations between entity pairs in a text. The relation is defined between an entity pair consisting of \textit{subject entity} ($e_{\text{subj}}$) and \textit{object entity} ($e_{\text{obj}}$). For example, in a sentence `Kierkegaard was born to an affluent family in Copenhagen’, the subject entity is `Kierkegaard’ and the object entity is `Copenhagen’. The goal is then to pick an appropriate relationship between these two entities; `\textit{place\_of\_birth}’.

RE is a task suitable for evaluating whether a model correctly understands the relationships between entities. 
In order to ensure {\kluere} captures this aspect of language understanding, we include a large-scale RE benchmark.
Because there is no large-scale RE benchmark publicly available in Korean, we collect and annotate our own dataset.

We formulate RE as a single sentence classification task. A model picks one of predefined relation classes describing the relation between two entities within a given sentence. In other words, the RE model predicts an appropriate relation $r$ of entity pair $(e_{\text{subj}}, e_{\text{obj}})$ in a sentence $s$, where $e_{\text{subj}}$ is the subject entity and $e_{\text{obj}}$ is the object entity. We refer to $(e_{\text{subj}}, r, e_{\text{obj}})$ as a relation triplet. The entities are marked as corresponding spans in each sentence $s$. There are 30 relation classes that consist of 18 person-related relations, 11 organization-related relations, and \textit{no\_relation}. Detailed explanation of these classes are presented in Table~\ref{tab:relation_classes}. We evaluate a model using micro F1 score, computed after excluding \textit{no\_relation}, and area under the precision-recall curve including all 30 classes.

\subsubsection{Data Construction}

Distant supervision \cite{mintz-etal-2009-distant} is a popular way to build a large-scale RE benchmark. It leverages relation triplets $(e_{\text{subj}}, r, e_{\text{obj}})$ in existing large-scale knowledge base (KB) such as Freebase. If a sentence $s$ in a large corpora includes $(e_{\text{subj}}, e_{\text{obj}})$ detected by an NER model simultaneously, it is added to the dataset with relation label $r$ by assuming any sentence which contains the pair will express that relation. This approach does not require expensive human annotation, thus allowing us to build a large-scale RE benchmark in a cost-effective way.

Despite this advantage, distant supervision often ends up with incorrect relation labels when the assumption is not satisfied. In particular, it only considers pairs of entities which are related to each other, which results in an RE model trained on such corpus to over-predict the existence of some relationship between any given pair of entities.
In other words, the predicted relation class distribution from such predictors is not realistic \cite{riedel2010modeling}. \citet{zhang-etal-2017-position} and \citet{nam-etal-2020-effective} thus propose to employ crowdworkers to alleviate erroneous relations extracted by distant supervision. \citet{riedel2010modeling} furthermore intentionally collect irrelevant entity pairs to prevent RE models from overly predicting false positives relations.

\begin{table}[t!]
    \small
    \centering
    \caption{30 relation classes defined in the relation schema of {\kluere}. Relation class $r$ should be one of the followings which consist of 18 person-related relations, 11 organization-related relations, and \textit{no\_relation}.}
    \begin{tabularx}{\textwidth}{lX}
        \toprule
        \textbf{Relation Class} & \textbf{Description} \\
        \midrule
        \textit{no\_relation} & No relation in between $(e_{\text{subj}}, e_{\text{obj}})$\\ \midrule
        \textit{org:dissolved}      & The date when the specified organization was dissolved \\
        \textit{org:founded}      & The date when the specified organization was founded \\
        \textit{org:place\_of\_headquarters}      & The place which the headquarters of the specified organization are located in \\
        \textit{org:alternate\_names}      & Alternative names called instead of the official name to refer to the specified organization \\
        \textit{org:member\_of}      & Organizations to which the specified organization belongs \\
        \textit{org:members}      & Organizations which belong to the specified organization \\
        \textit{org:political/religious\_affiliation}      & Political/religious groups which the specified organization is affiliated in \\
        \textit{org:product}      & Products or merchandise produced by the specified organization \\
        \textit{org:founded\_by}      & The person or organization that founded the specified organization \\
        \textit{org:top\_members/employees}      & The representative(s) or members of the specified organization \\
        \textit{org:number\_of\_employees/members}      & The total number of members that are affiliated in the specified organization \\ \midrule
        \textit{per:date\_of\_birth}      & The date when the specified person was born \\
        \textit{per:date\_of\_death}      & The date when the specified person died \\
        \textit{per:place\_of\_birth}      & The place where the specified person was born \\
        \textit{per:place\_of\_death}      & The place where the specified person died \\
        \textit{per:place\_of\_residence}      & The place where the specified person lives \\
        \textit{per:origin}      & The origins or the nationality of the specified person \\
        \textit{per:employee\_of}      & The organization where the specified person works \\
        \textit{per:schools\_attended}      & A school where the specified person attended \\
        \textit{per:alternate\_names}      & Alternative names called instead of the official name to refer to the specified person \\
        \textit{per:parents}      & The parents of the specified person \\
        \textit{per:children}      & The children of the specified person \\
        \textit{per:siblings}      & The brothers and sisters of the specified person \\
        \textit{per:spouse}      & The spouse(s) of the specified person \\
        \textit{per:other\_family}      & Family members of the specified person other than parents, children, siblings, and spouse(s) \\
        \textit{per:colleagues}      & People who work together with the specified person \\
        \textit{per:product}      & Products or artworks produced by the specified person \\
        \textit{per:religion}      & The religion in which the specified person believes \\
        \textit{per:title}      & Official or unofficial names that represent the occupational position of the specified person \\
        \bottomrule
    \end{tabularx}
    \label{tab:relation_classes}
\end{table}

\paragraph{Overview}

We modify the original strategy of distant supervision above, to address this weakness and to better fit our situation. First, we collect triplets $(e_{\text{subj}}, r, e_{\text{obj}})$ from a small Korean KB\footnote{
\url{https://aihub.or.kr/aidata/84}
}
and build additional ones by parsing the infoboxes in {\wikipedia} and {\namuwiki}\footnote{
\url{https://namu.wiki}
}
to enlarge the pool of the candidate triplets. We then ask crowdworkers to select the correct relation class of each candidate triplet within a sentence, compared to distant supervision which directly uses automatically generated relation labels.
In addition, we randomly sample entity pairs in $s$ to obtain more realistic relation class distribution in our benchmark. Those examples would include unseen entities in existing KB as well as have higher chance to be irrelevant (\textit{no\_relation}).

This procedure can be divided into five steps; (1) candidate sentence collection, (2) relation schema definition, (3) entity detection, (4) entity pair selection and (5) relation annotation. We elaborate each step in the rest of this section.

\paragraph{1. Collect Candidate Sentences}

We sample candidate sentences from {\wikipedia}, {\wikitree} and {\policy} corpora to cover a diverse set of named entities and relational facts. Since our task deals with single sentences, we exploit individual sentences split by Korean Sentence Splitter \footnote{
\url{https://github.com/hyunwoongko/kss}
} at the preprocessing step.
We filter out sentences that contain undesirable social bias and are considered hate speech, using a classifier trained on the Korean hate speech dataset \cite{moon-etal-2020-beep}.

\paragraph{2. Define Relation Schema}  
We design a relation schema based on the schema from Text Analysis Conference Knowledge Base Population (TAC-KBP) \cite{mcnamee2009overview}. Our schema defines entity types and relation classes. Similar to TAC-KBP, we constrain $e_{\text{subj}}$ to be of either PER (Person) or ORG (Organization) type. $e_{\text{obj}}$ can have one of the following types: PER, ORG, LOC (Location), DAT (Date and time), POH (Other proper nouns), and NOH (Other numerals). For the relation classes, we adapt the original classes in TAC-KBP to our corpus, following \citet{yu-etal-2020-dialogue}.

We remove rarely appearing relation classes in our corpus such as \textit{org:website}, \textit{per:shareholders}, \textit{per:cause\_of\_death}, \textit{per:charges}, and \textit{per:age}. For the same reason, we incorporate \textit{org:parents} into \textit{org:member\_of} and \textit{org:subsidiaries} into \textit{org:members}.
Since the taxonomy of TAC-KBP does not precisely reflect the regional hierarchy of Korea, we integrate the prefixes \textit{country\_of}, \textit{city\_of}, and \textit{stateorprovince\_of} into \textit{place\_of}.
We introduce additional classes frequently appearing in our corpus such as \textit{org:product}, \textit{per:product} and \textit{per:colleague}:

\begin{itemize}[leftmargin=*]
    \item \textit{org:product}: A product or merchandise produced by an organization. This includes intangible goods such as an event hosted and a business launched by the organization.
    
    \item \textit{per:product}: A product produced by a person. Artworks (e.g. book, music, movie) or contribution to producing them.
    
    \item \textit{per:colleague}: A person could be a colleague of someone if they work together. Two people in the same group such as political party or alliance are colleagues as well.
\end{itemize}

\paragraph{3. Detect Entities}

We automatically detect named entities in all candidate sentences. We fine-tune a pre-trained ELECTRA for Korean\footnote{
\url{https://github.com/monologg/KoELECTRA}
} 
to build two named entity recognition (NER) models on two existing Korean NER resources respectively. One is provided by National Institute of Korean Language \cite{nikl2020corpora}, and the other is built by Korea Maritime \& Ocean University.\footnote{
\url{https://github.com/kmounlp/NER}
}
We modify the named entity types defined in these resources to be compatible with our own entity types previously defined in the schema. We take the union of both models' predictions to extract as many entities as possible. We use crowdsourcing to correct incorrect boundaries of the detected entities, as described later.

\paragraph{4. Select Entity Pairs}

We select two entities from the entity set $E$ of a given sentence $s$ to make an entity pair $(e_{\text{subj}}, e_{\text{obj}})$. In doing so, we take two distinct approaches; (1) KB-based sampling and (2) uniform sampling. 

For the first approach, we only consider the subset of entities such that each entity pair $(e_{\text{subj}}, e_{\text{obj}})$ appears in the pool of triplets $(e_{\text{subj}}, r, e_{\text{obj}})$. We collect these triplets from two sources. First, we create the initial pool of triplets, using a Korean KB.\footnote{
Released by NIA, a government-funded institution. Available at \url{https://aihub.or.kr/aidata/84}.
} 
Because the number of triplets ($\sim$800k) from the Korean KB is small compared to, for instance, that of Freebase ($\sim$2b), we enlarge this pool of triplets by gathering and then parsing infoboxes in {\wikipedia} and Namuwiki.
In order to avoid over-inclusion of frequent entities, such as the President of Korea,
we set an upper bound to the number of co-occurrence between $(e_{\text{subj}}, e_{\text{obj}})$ during sampling \cite{zhang-etal-2017-position}.

In the second approach, $(e_{\text{subj}}, e_{\text{obj}})$ is uniformly sampled from the entire entity set $E$ of a given sentence $s$, at random. Because there is no cue whether a sampled pair has any relation between them, the pair is highly likely to be irrelevant (\textit{no\_relation}). Irrelevant pairs will account for a large portion of realistic relation distribution between two arbitrary entities. Therefore, this approach helps to set up real-world scenario.
Such a pair is also likely to contain entities that are not selected in the first approach. This leads to capturing entity pairs and their relations independent of KBs.

\paragraph{5. Annotate Relations}

\begin{figure}[t!] 
\centering
\label{fig:re_tool}
\includegraphics[width=0.8\textwidth]{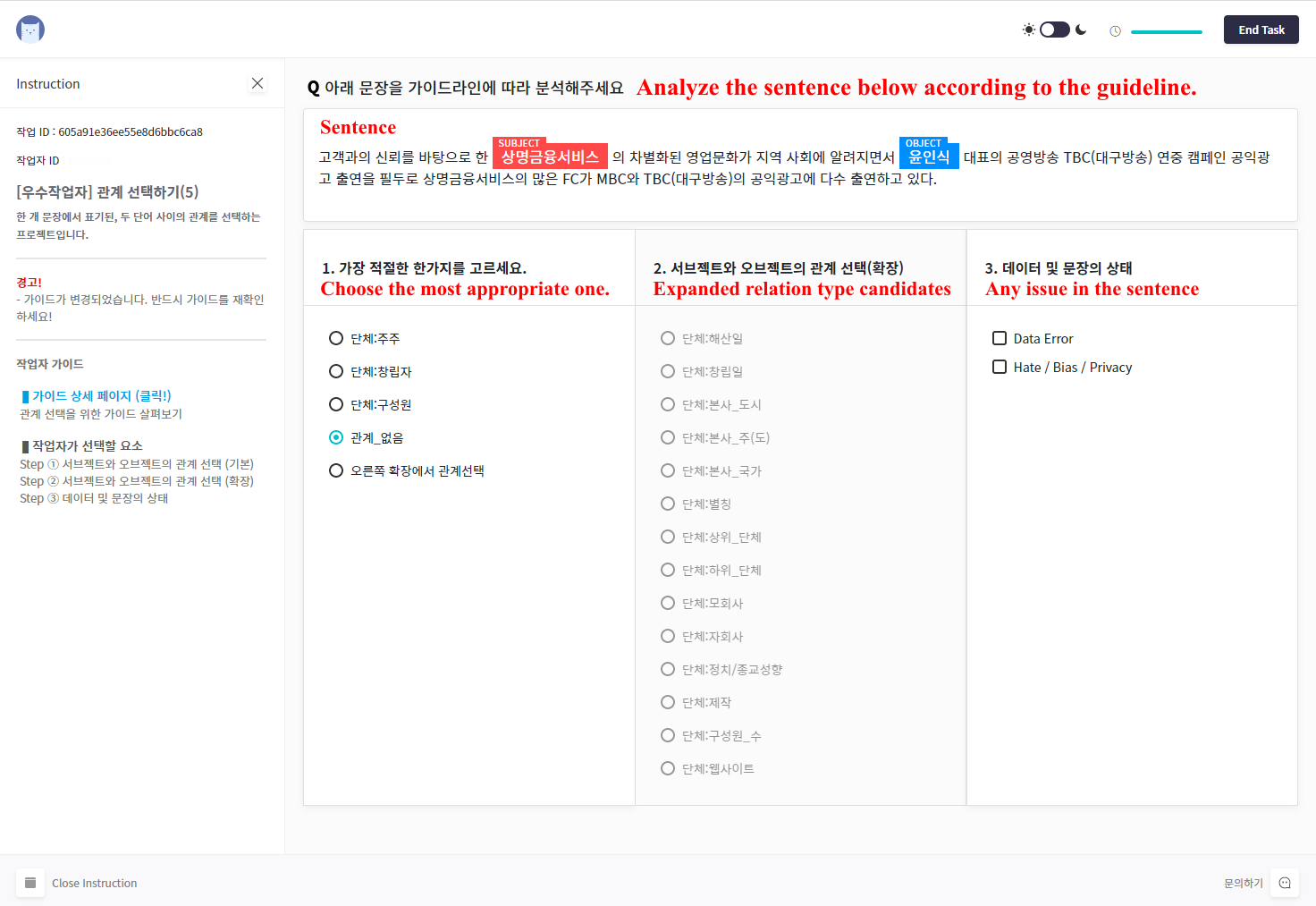}
\caption{Annotation tool for crowdsourcing. Main features are translated in English with red color.}
\end{figure}

We ask workers recruited by DeepNatural,\footnote{\url{https://deepnatural.ai/}} a Korean crowdsourcing platform, to annotate each entity pair $(e_{\text{subj}}, e_{\text{obj}})$ with a relation label $r$. We instruct workers to focus on the current relationship, not ones from the past. For instance, if a person described in a sentence is a former member of a certain organization, workers are asked not to choose the relation \textit{per:employee\_of}. We also ask them to avoid relying on external knowledge, or common sense, to infer the relation from the context solely within a given sentence. Workers report examples that contain hate speech, biased expressions, or personally identifiable information. In addition, they are asked to report sentences with incorrect entity boundaries.

We employ 163 qualified workers, each of which correctly labelled at least 4 out of 5 questions during the pilot annotation phase. After the pilot phase, 3 workers are assigned to each example independently to label the relation. Figure~\ref{fig:re_tool} shows the annotation tool for crowdsourcing. To reduce cognitive burden of annotators, we provide a small number of candidate relations at first. The candidates consist of relations that can be defined between types of entity pair predicted by the NER models. If one cannot find appropriate $r$ in the candidates, they are expanded to all relation classes.

\begin{table}[t!]
    \centering
    \caption{Relation distribution of {\kluere}.}
    \label{tab:relation_distribution}
    \begin{tabular}{@{}l cc cc cc@{}}
        \toprule
        & \multicolumn{2}{c}{\textbf{Train}} & \multicolumn{2}{c}{\textbf{Dev}} & \multicolumn{2}{c}{\textbf{Test}} \\ \cmidrule(lr){2-3} \cmidrule(lr){4-5} \cmidrule(lr){6-7}
        \textbf{Relation Class} & Count & Ratio & Count & Ratio & Count & Ratio \\
        \midrule
        \textit{no\_relation}                           & 9,534 & 29.36\% & 4,631 & 59.64\% & 4,632 & 59.64\% \\ \midrule
        \textit{org:dissolved}                          &   66 &  0.20\% &   11 &  0.14\% &   10 &  0.13\% \\
        \textit{org:founded}                            &  450 &  1.39\% &   20 &  0.26\% &   20 &  0.26\% \\
        \textit{org:place\_of\_headquarters}            & 1,195 &  3.68\% &  194 &  2.50\% &  193 &  2.49\% \\
        \textit{org:alternate\_names}                   & 1,320 &  4.07\% &   78 &  1.00\% &   77 &  0.99\% \\
        \textit{org:member\_of}                         & 1,866 &  5.75\% &  104 &  1.34\% &  105 &  1.35\% \\
        \textit{org:members}                            &  420 &  1.29\% &  122 &  1.57\% &  122 &  1.57\% \\
        \textit{org:political/religious\_affiliation}   &   98 &  0.30\% &   13 &  0.17\% &   13 &  0.17\% \\
        \textit{org:product}                            &  380 &  1.17\% &  235 &  3.03\% &  235 &  3.03\% \\
        \textit{org:founded\_by}                        &  155 &  0.48\% &   11 &  0.14\% &   11 &  0.14\% \\
        \textit{org:top\_members/employees}             & 4,284 & 13.19\% &  513 &  6.61\% &  514 &  6.62\% \\
        \textit{org:number\_of\_employees/members}      &   48 &  0.15\% &   17 &  0.22\% &   18 &  0.23\% \\ \midrule
        \textit{per:date\_of\_birth}                    & 1,130 &  3.48\% &   12 &  0.15\% &   12 &  0.15\% \\
        \textit{per:date\_of\_death}                    &  418 &  1.29\% &   13 &  0.17\% &   13 &  0.17\% \\
        \textit{per:place\_of\_birth}                   &  166 &  0.51\% &   11 &  0.14\% &   10 &  0.13\% \\
        \textit{per:place\_of\_death}                   &   40 &  0.12\% &   10 &  0.13\% &   11 &  0.14\% \\
        \textit{per:place\_of\_residence}               &  193 &  0.59\% &  124 &  1.60\% &  125 &  1.61\% \\
        \textit{per:origin}                             & 1,234 &  3.80\% &  118 &  1.52\% &  118 &  1.52\% \\
        \textit{per:employee\_of}                       & 3,573 & 11.00\% &  242 &  3.12\% &  241 &  3.10\% \\
        \textit{per:schools\_attended}                  &   82 &  0.25\% &   11 &  0.14\% &   11 &  0.14\% \\
        \textit{per:alternate\_names}                   & 1,001 &  3.08\% &  104 &  1.34\% &  103 &  1.33\% \\
        \textit{per:parents}                            &  520 &  1.60\% &   27 &  0.35\% &   27 &  0.35\% \\
        \textit{per:children}                           &  304 &  0.94\% &   27 &  0.35\% &   27 &  0.35\% \\
        \textit{per:siblings}                           &  136 &  0.42\% &   24 &  0.31\% &   24 &  0.31\% \\
        \textit{per:spouse}                             &  795 &  2.45\% &   41 &  0.53\% &   40 &  0.52\% \\
        \textit{per:other\_family}                      &  190 &  0.59\% &   34 &  0.44\% &   35 &  0.45\% \\
        \textit{per:colleagues}                         &  534 &  1.64\% &  220 &  2.83\% &  220 &  2.83\% \\
        \textit{per:product}                            &  139 &  0.43\% &   67 &  0.86\% &   69 &  0.89\% \\
        \textit{per:religion}                           &   96 &  0.30\% &   13 &  0.17\% &   12 &  0.15\% \\
        \textit{per:title}                              & 2,103 &  6.48\% &  718 &  9.25\% &  718 &  9.25\% \\
        \midrule
        \textbf{Total} & \textbf{32,470} & 100.00\% & \textbf{7,765} & 100.00\% & \textbf{7,766} & 100.00\% \\
        \bottomrule
    \end{tabular}
\end{table}

We take majority-voted labels as gold labels. For each example without a majority label, the top 30 annotators select the final label from the annotated labels. We do not include examples reported as hate speech, biased, or to have privacy issues. The inter-annotator agreement (Krippendorff’s $\alpha$) on the annotated dataset is 0.701 \cite{Krippendorff2011ComputingKA}.

\paragraph{Final Dataset}

{\kluere} consists of 32,470 training, 7,765 development and 7,766 test examples. 
For real-world scenario, we only use examples created from uniform sampling when building the development and test sets. In the test set, we only include sentences with entities that do not appear in the training set.

The average length of a sentence in {\kluere} is 95.9 characters including whitespaces. The proportions of the entity types are: PER (38.1\%), ORG (36.3\%), LOC (6.2\%), DAT (6.2\%), POH (11.9\%), and NOH (1.3\%). The distribution of the relation classes is shown in Table~\ref{tab:relation_distribution}.

\subsubsection{Evaluation Metrics}
\label{sec:re-metric}

The evaluation metrics for \kluere{} are 1) micro F1 score on relation existing cases, and 2) area under the precision-recall curve (AUPRC) on all classes. Micro F1 score is a harmonic mean of micro-precision and micro-recall. It measures the F1 score of the aggregated contributions of all classes. It gives each sample the same importance, thus naturally weighting more on the majority class. We remove the dominant class ($no\_relation$) for this metric to not incentivize the model that focus more on predicting negative class. AUPRC is an averaged area under the precision-recall curves whose x-axis is recall and y-axis is the precision of all relation classes. It is a useful metric for this imbalanced data setting where important positive examples are rarely occurred.

\subsubsection{Related Work}

Many researchers attempt to build KBs from unstructured text through automatically identifying relational facts between entity pairs in plain text by applying machine learning techniques. \citet{doddington-etal-2004-automatic} and \citet{hendrickx-etal-2010-semeval} construct English datasets to train such models, including a relatively small number of relation classes for general domain text. \citet{mintz-etal-2009-distant} further propose distant supervision to automatically annotate plain text by aligning it to the schema of KBs. This allows researchers to scale up the size of RE datasets \cite{riedel2010modeling, zhang-etal-2017-position, han-etal-2018-fewrel, yao-etal-2019-docred, yu-etal-2020-dialogue}. Among these recent studies, TACRED \cite{zhang-etal-2017-position} is the most widely used dataset, built based on the popular relation schema TAC-KBP \cite{mcnamee2009overview} which mainly focuses on person and organization entities. Specifically, TACRED contains 106,264 examples annotated with the 42 relation classes. \citet{yu-etal-2020-dialogue} also proposes a dialogue-based RE task by refining TAC-KBP to obtain 36 relation classes adapted for the dialogue domain. We also follow TAC-KBP to build the relation schema and modify them suitable for our situation.

In the cases of languages other than English, there are only a few existing benchmarks, including one in Chinese \cite{xu2017discourse}, one in German \cite{schiersch-etal-2018-german}, and one in French \cite{jabbari-etal-2020-french}.  \citet{nam-etal-2020-effective} propose an RE dataset in Korean using distant supervision to automatically generate and annotate examples. It however has a relatively small test set ($\sim$3k), making it difficult to evaluate performance for the total 49 relation classes properly. Moreover, since there is no negative class (\textit{no\_relation} in ours), it is likely to encourage models to overly predict false positives \cite{zhang-etal-2017-position}. We thus consider {\kluere} as a standard large-scale RE benchmark to properly evaluate Korean language models.

\subsubsection{Conclusion}

We propose {\kluere}\index{KLUE-RE}, a large-scale human-annotated RE benchmark for Korean. To overcome the lack of large-scale and up-to-date Korean KBs, we design an efficient candidate collection method, coupled with an effective annotation scheme. {\kluere} can not only be used for online information extraction but also contribute to building a large-scale knowledge graph from unstructured texts. 
We therefore expect {\kluere} to be a starting point for building a large-scale, ever-growing public KB in Korean, as well as a valuable Korean NLU benchmark.

\clearpage
\subsection{Dependency Parsing (\posdp)}

Dependency parsing (DP)\index{DP (Dependency Parsing)} is an NLP task that aims at finding relational information among words. It has been an important component in many NLP systems, because of its ability to capture the syntactic feature of a sentence. We include DP in KLUE to evaluate the representational power of language models in terms of syntactic features.

Formally, a dependency parser predicts a graph structure of an input sentence based on the dependency grammar \cite{de2006generating,de2008stanford}. In general, a parsed tree consists of dependency arcs, connecting dependents to their heads, and the dependency labels attached to the arcs that represent the relations between dependents and their heads. 

\begin{figure}[htbp!]
\centering
\includegraphics[scale=0.13]{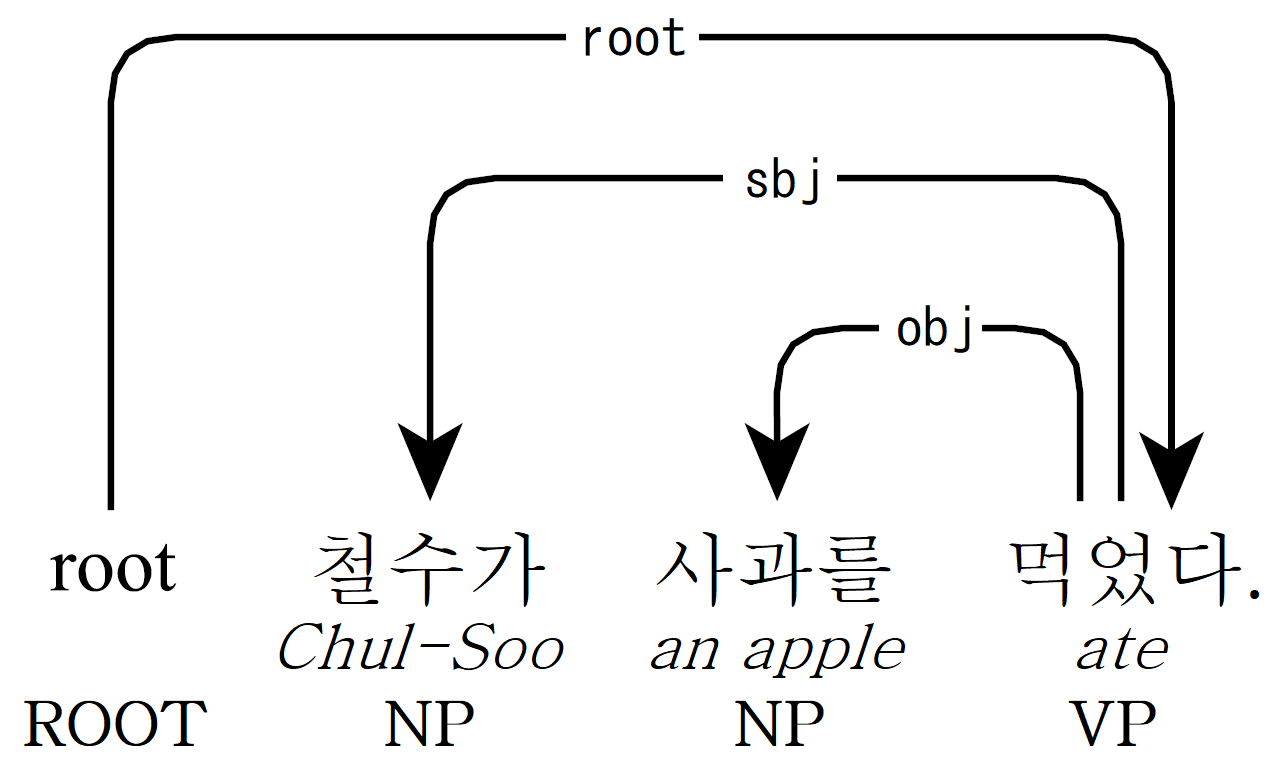}
\caption{An example of dependency parsing, that translates to "Chul-Soo ate an apple."}
\label{fig:dp1}
\end{figure}

For example, Figure~\ref{fig:dp1} shows a parsed result of the example sentence: ``철수가 사과를 먹었다 (Chul-Soo ate an apple.)''. In the tree, arrows depart from \textit{head} and point to their \textit{dependents}. Thus `철수가 (Chul-Soo)' and `사과를 (an apple)' are \textit{dependents} of `먹었다(ate)' and `먹었다(ate)' is the \textit{head} of `철수가 (Chul-Soo)' and `사과를 (an apple)'. Also, `철수가 (Chul-Soo)' is dependent on `먹었다 (ate)' with a ``Subject'' relation. This dependency relation label is called DEPREL. For DEPREL, we follow the TTA Dependency annotation
scheme\footnote{ 
\url{https://aiopen.etri.re.kr/data/003.\%EC\%9D\%98\%EC\%A1\%B4\%EA\%B5\%AC\%EB\%AC\%B8\%EB\%B6\%84\%EC\%84\%9D_\%EA\%B0\%80\%EC\%9D\%B4\%EB\%93\%9C\%EB\%9D\%BC\%EC\%9D\%B8.pdf}}  consisting of a combination of 9 syntax tags and 6 function tags.

Since each word in a sentence has a pair of dependency information (HEAD, DEPREL), DP is conventionally formulated as a word-level sequence tagging task. We evaluate a model’s performance using unlabeled attachment score (UAS) and labeled attachment score (LAS). During the evaluation, labels with a cumulative frequency of 1\% from the bottom are grouped into the OTHERS label to compensate for the negative impact of lower frequency labels on LAS.

\begin{figure}[htbp!]
\centering
\includegraphics[scale=0.16]{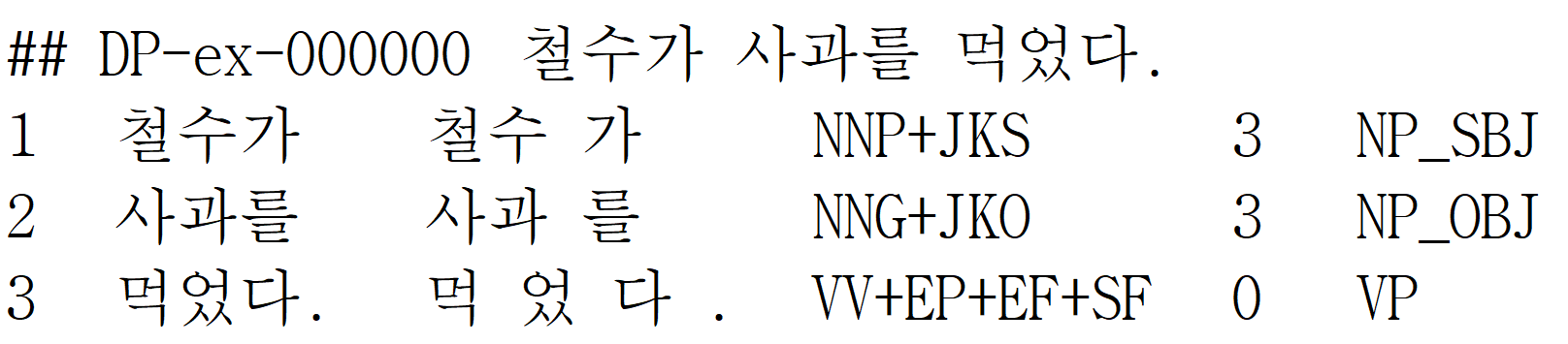}
\caption{A demonstration of the KLUE-DP output format using a sentence that translates to ``Chul-Soo ate an apple.''}
\label{fig:dp2}
\end{figure}

We represent the output in a CoNLL-like format, as shown in Figure~\ref{fig:dp2}. This format consists of 6 columns, each column contains a word index (ID), word form (FORM), lemma of word form (LEMMA), part-of-speech tag (POS), head of the current word (HEAD), and dependency relation (DEPREL).

\subsubsection{Dataset Construction}

\paragraph{Source Corpora}
To build our corpus as a suitable dataset for a general-usage DP model, we take into account both formal and informal texts. We use {\wikitree} and {\airbnb} as source corpora. {\wikitree} consists of news articles and represents grammatically sound formal texts. On the other hand, {\airbnb} mostly consists of user-generated reviews containing web texts, thus showing frequent omission of components and free word order. We collect the same rate of data from both {\wikitree} and {\airbnb} so that our dataset represents both refined written sentences and noisy colloquial texts. 

\paragraph{Annotation Protocol}
Since the part-of-speech indicates how the word functions grammatically in a sentence, it is highly related to the dependency relations. To utilize POS information as an additional syntactic feature, we first annotate POS on the corpus prior to dependency relation annotation. To this end, we follow TTA POS tagging guideline.\footnote{\url{https://aiopen.etri.re.kr/data/001.\%ED\%98\%95\%ED\%83\%9C\%EC\%86\%8C\%EB\%B6\%84\%EC\%84\%9D_\%EA\%B0\%80\%EC\%9D\%B4\%EB\%93\%9C\%EB\%9D\%BC\%EC\%9D\%B8.pdf}} We use annotated POS information when constructing the DP corpus. 

Next, we modify the original TTA DP guideline for dependency relation annotation. We add guides for spoken and web data, since the original guideline only contains instruction for annotating written data. 

\begin{table}[t!]
\centering
\caption{Syntax and function tagset (label types) of TTA DP guideline.}
\begin{tabular}{@{}ll@{}}
\toprule
\textbf{Label Type} & \textbf{Description} \\ \midrule
\multicolumn{2}{l}{Syntax} \\ \midrule
NP & Noun Phrase \\
VP & Verb Phrase \\
AP & Adverb Phrase \\
VNP & Copula Phrase \\
DP & Adnoun Phrase \\
IP & Interjection Phrase \\
X & Pseudo Phrase \\
L & Left Parenthesis and Quotation Mark \\
R & Right Parenthesis and Quotation Mark \\ \midrule
\multicolumn{2}{l}{Function} \\ \midrule
SBJ & Subject \\
OBJ & Object  \\
MOD & Noun Modifier \\
AJT & Predicate Modifier \\
CMP & Complement \\
CNJ & Conjunction \\ \bottomrule
\end{tabular}
\label{table:tta_dp_syntax_function}
\end{table}

We follow TTA DP tagset, which is a standard in Korean. As shown in Table \ref{table:tta_dp_syntax_function}, it is made up of a combination of 9 syntax tags and 6 function tags. The syntax tag indicates the POS for token, and there are NP (noun phrase), VP (verb phrase), AP (adverb phrase), VNP (copula phrase), DP (adnoun phrase), IP (interjection phrase), X (pseudo phrase), L (left parenthesis and quotation mark) and R (right parenthesis and quotation mark). The function tag indicates what function the token performs in relation to the head, and there are SBJ (subject), OBJ (object), MOD (noun modifier), AJT (predicate modifier), CMP (complement), and CNJ (conjunction). The TTA DP combines a syntax tag and a function tag into a single tag, such as NP\_SBJ (noun phrase subject) and VP\_AJT (verb phrase adverb).

\paragraph{Annotation Process}
To build highly reliable corpora for both POS and DP annotation, we use an annotation tool provided by DeepNatural,\footnote{\url{https://deepnatural.ai/}} a Korean crowdsourcing platform, which asks the annotators to annotate cross-reference. After POS annotation is completed, dependency relation annotation proceeds on the same sentences by using POS information. Both POS and DP are annotated by ten PhD students in Korean linguistics, who previously contributed to constructing \modu{}\footnote{\url{https://corpus.korean.go.kr/main.do}} \cite{nikl2020corpora} opened by the National Institute of the Korean Language. Prior to annotation, we instruct them with our guideline. During annotation, we respond to questions from annotators in real-time, and accordingly, the guidelines are iteratively updated. Annotators report sentences containing hate speech or bias during the annotation. In addition, sentences containing personal information (e.g., name, address, phone number, etc.) are reported. The reported sentences are removed from the dataset after our inspection.

Annotations are validated in two steps. First, each annotator reviews and corrects POS and DP labels annotated by other annotators. Then we finally review all data and revise remaining mislabeled sentences.

\paragraph{Final Dataset}
{\klueposdp} consists of a total of 14,500 sentences, including 7,250 sentences for {\wikinews} and 7,250 sentences for {\airbnb}. POS annotation is given in the 4th column of CoNLL-like format, along with HEAD and DEPREL information in next columns. We set the train/dev/test split as 10,000, 2,000, 2,500 sentences. Tables~\ref{table:dp_final_dataset} and \ref{table:dp_label_dataset} show detailed statistics of {\klueposdp}.

\begin{table}[t!]
\centering
\caption{Statistics for {\klueposdp}.}
\begin{tabular}{@{}lcccc@{}}
\toprule
\textbf{Source}           & \textbf{|Train|} & \textbf{|Dev|} & \textbf{|Test|} & \textbf{Total} \\ \midrule
\wikitree  & 5,000    & 1,000   & 1,250    & 7,250  \\
\airbnb  & 5,000    & 1,000   & 1,250    & 7,250  \\ \midrule
\textbf{Total}                   & \textbf{10,000}   & \textbf{2,000}  & \textbf{2,500}   & \textbf{14,500} \\ \bottomrule
\end{tabular}
\label{table:dp_final_dataset}
\end{table}

\begin{table}[t!]
\centering
\caption{Label statistics for {\klueposdp}. Predictions on the labels from VP\_CMP to NP\_SVJ is be merged into OTHERS when calculating LAS.}
\begin{tabular}{@{}lcccc@{}}
\toprule
\textbf{Label}           & \textbf{|Train|} & \textbf{|Dev|} & \textbf{|Test|} & \textbf{Total} \\ \midrule
NP & 23,902 (20.88\%) & 4,559 (20.27\%) & 4,884 (19.51\%) & 33,345 (20.58\%) \\
NP\_AJT & 17,552 (15.33\%) & 3,415 (15.18\%) & 3,526 (14.09\%) & 24,493 (15.12\%) \\
VP & 16,786 (14.66\%) & 3,322 (14.77\%) & 3,917 (15.65\%) & 24,025 (14.83\%) \\
NP\_SBJ & 13,412 (11.71\%) & 2,737 (12.17\%) & 3,112 (12.43\%) & 19,261 (11.89\%) \\
VP\_MOD & 12,210 (10.66\%) & 2,457 (10.92\%) & 2,682 (10.71\%) & 17,349 (10.71\%) \\
NP\_OBJ & 8,531 (7.45\%) & 1,691 (7.52\%) & 1,951 (7.79\%) & 12,173 (7.51\%) \\
AP & 6,638 (5.80\%) & 1,274 (5.66\%) & 1,716 (6.85\%) & 9,628 (5.94\%) \\
NP\_CNJ & 3,945 (3.45\%) & 795 (3.53\%) & 764 (3.05\%) & 5,504 (3.40\%) \\
NP\_MOD & 3,450 (3.01\%) & 659 (2.93\%) & 727 (2.90\%) & 4,836 (2.98\%) \\
VNP & 2,550 (2.23\%) & 495 (2.20\%) & 558 (2.23\%) & 3,603 (2.22\%) \\
DP & 1,994 (1.74\%) & 376 (1.67\%) & 419 (1.67\%) & 2,789 (1.72\%) \\
VP\_AJT & 882 (0.77\%) & 196 (0.87\%) & 182 (0.73\%) & 1,260 (0.78\%) \\
VNP\_MOD & 854 (0.75\%) & 180 (0.80\%) & 164 (0.66\%) & 1,198 (0.74\%) \\
NP\_CMP & 408 (0.36\%) & 83 (0.37\%) & 97 (0.39\%) & 588 (0.36\%) \\
VP\_SBJ & 338 (0.30\%) & 59 (0.26\%) & 94 (0.38\%) & 491 (0.30\%) \\ 
\midrule
VP\_CMP & 330 (0.29\%) & 59 (0.26\%) & 91 (0.36\%) & 483 (0.30\%) \\
VP\_OBJ & 279 (0.24\%) & 39 (0.17\%) & 68 (0.27\%) & 386 (0.24\%) \\
VNP\_CMP & 131 (0.11\%) & 27 (0.12\%) & 26 (0.10\%) & 184 (0.11\%) \\
AP\_MOD & 59 (0.05\%) & 15 (0.07\%) & 11 (0.04\%) & 84 (0.05\%) \\
X\_AJT & 41 (0.04\%) & 14 (0.06\%) & 10 (0.04\%) & 63 (0.04\%) \\
VNP\_AJT & 37 (0.03\%) & 13 (0.06\%) & 9 (0.04\%) & 59 (0.04\%) \\
VP\_CNJ & 37 (0.03\%) & 7 (0.03\%) & 7 (0.03\%) & 46 (0.03\%) \\
IP & 27 (0.02\%) & 4 (0.02\%) & 6 (0.02\%) & 41 (0.03\%) \\
X & 26 (0.02\%) & 4 (0.02\%) & 2 (0.01\%) & 31 (0.02\%) \\
VNP\_OBJ & 18 (0.02\%) & 3 (0.01\%) & 2 (0.01\%) & 26 (0.02\%) \\
X\_SBJ & 17 (0.01\%) & 3 (0.01\%) & 2 (0.01\%) & 21 (0.01\%) \\
X\_OBJ & 12 (0.01\%) & 2 (0.01\%) & 2 (0.01\%) & 17 (0.01\%) \\
VNP\_SBJ & 11 (0.01\%) & 1 (0.00\%) & 2 (0.01\%) & 14 (0.01\%) \\
L & 3 (0.00\%) & 1 (0.00\%) & 1 (0.00\%) & 4 (0.00\%) \\
AP\_AJT & 2 (0.00\%) & 1 (0.00\%) & 1 (0.00\%) & 4 (0.00\%) \\
X\_CMP & 2 (0.00\%) & 1 (0.00\%) & 0 (0.00\%) & 4 (0.00\%) \\
X\_CNJ & 2 (0.00\%) & 1 (0.00\%) & 0 (0.00\%) & 3 (0.00\%) \\
X\_MOD & 2 (0.00\%) & 0 (0.00\%) & 0 (0.00\%) & 2 (0.00\%) \\
AP\_CMP & 1 (0.00\%) & 0 (0.00\%) & 0 (0.00\%) & 2 (0.00\%) \\
R & 1 (0.00\%) & 0 (0.00\%) & 0 (0.00\%) & 1 (0.00\%) \\
VNP\_CNJ & 1 (0.00\%) & 0 (0.00\%) & 0 (0.00\%) & 1 (0.00\%) \\
AP\_SBJ & 0 (0.00\%) & 0 (0.00\%) & 0 (0.00\%) & 1 (0.00\%) \\
NP\_SVJ & 0 (0.00\%) & 0 (0.00\%) & 0 (0.00\%) & 0 (0.00\%) \\ \midrule
\textbf{Total}  & \textbf{114,491} & \textbf{22,496} & \textbf{25,033} & \textbf{162,020} \\ \bottomrule
\end{tabular}
\label{table:dp_label_dataset}
\end{table}

\subsubsection{Evaluation Metrics}
\label{sec:dp-metric}

The evaluation metrics for \klueposdp{} are 1) unlabeled attachment score (UAS) and 2) labeled attachment score (LAS), which are popular metrics for DP. Given the goal of DP is to predict head indices (HEAD) and dependency relation classes (DEPREL), UAS only counts HEAD prediction while LAS counts both HEAD and DEPREL. Specifically, UAS calculates macro F1 score on HEAD prediction, while LAS calculates macro F1 score on DEPREL whose HEAD prediction is correct. Both scores give the same importance to all classes. For LAS, since DEPREL distribution is highly skewed, we combine the predictions on the labels with a cumulative frequency of 1\% from the bottom into a single label (OTHERS) and then calculate F1 score. The less-appeared labels are referred in Table~\ref{table:dp_label_dataset}.

\subsubsection{Related Work}
The Penn Treebank \cite{marcus1993building} is a constituency parsed dataset created from 1989 to 1996. It has a size of about 3 million words, including IBM computer manuals, nursing notes, Wall Street Journal articles, phone conversations. A total of 48 POS tags and 18 syntax tags are used by combining meta tags such as symbols. The Penn Treebank was the best-known parsing data set before dependency parsing became prevalent. Later studies convert the Penn Treebank to dependency parsing \cite{mcdonald2013universal, chun2018building}.

A representative DP corpus is the Universal Dependencies(UD) dataset.\footnote{\url{https://universaldependencies.org}} UD is de facto standard of DP data, aiming for a unified treebank annotation in various languages. Google Universal POS~\cite{petrov2011universal} composed of 12 tags, was developed and a corpus was built that applied it to 25 different languages. Also \citet{de2008stanford} studied guidelines for dependency parsing markers used in Stanford parsers \cite{klein2003accurate}. The Universal Dependency Treebank Project in 2013 attempted to combine the two studies above to have a consistent annotation system for multiple languages. UD started by modifying and supplementing this. UD first started in 2015 and \citet{nivre2016universal} and 10 corpora in a total of 10 languages were released through the website. As of 2021, UD (UD 2.7v) offers 104 languages and 183 corpora.

The corpus constructed with UD scheme is made according to a certain structure called CoNLL-U format. Since UD aims to study general linguistic universality by using the same tag and annotation system in different languages, a unified format for integrating and managing each corpus is needed. The CoNLL-U format is a modified version of this CoNLL format so that it can well represent Universal dependency parsing. CoNLL-U format consists of 10 columns, each column display a word index (ID), word form (FORM), lemma of word form (LEMMA), universal part-of-speech tag (UPOS), language-specific part- It represents of-speech tag (XPOS), list of morphological features (FEATS), head of the current word (HEAD), dependency relation (DEPREL), enhanced dependency graph (DEPS), any other annotation (MISC). In this format, each word stands on a line along with different associated features (word form, lemma, POS tag, etc.) and we adopt this format in our final dataset.

In Korean, DP corpus is divided into those that follow the UD scheme and those that do not. Among the former are The Google Korean Universal Dependency Treebank (GKT), The KAIST Korean Universal Dependency Treebank (KTB), and The Penn Korean Universal Dependency Treebank (PKT). These three datasets are converted from The Google Korean Treebank \cite{mcdonald2013universal}, The Kaist Treebank \cite{choi1994kaist}, and The Penn Korean treebank \cite{han2006korean} according to the UD scheme, respectively~\cite{chun2018building}. These were first automatically converted according to the head-finding rule and then heuristically modified. These are composed of 6k, 27k, and 5k sentences, respectively, and include the genres of blog, newswire, literature, academic, and manuscript. Among them, PKT was revised by changing the analysis unit and several rules to further reveal the characteristics of Korean \cite{oh2020analysis}.

Corpora that do not follow the UD scheme include the TTA DP Corpus built by Electronics and Telecommunications Research Institute (ETRI) and the Modu Corpus \cite{nikl2020corpora} built by the National Institute of the Korean Language (NIKL). Both corpora follow the CoNLL format, and use their own tagset, which was developed from the 21st century Sejong Plan corpus. The syntactic analysis corpus of the 21st century Sejong Plan is constructed according to the constituency grammar, following a scheme similar to the Penn Treebank. Unlike the dependency grammar, which grasps only the dominant relationship between two words, the constituency grammar identifies the relationship between words hierarchically. However, since Korean has relatively free word order, dependency parsing is more suitable than phrase-structure parsing. Studies on converting the 21st century Sejong Plan corpus into DP format have been conducted, and since then, corpora that use the 21st century Sejong Plan's tagset but follow dependency parsing have been constructed. The size of TTA DP corpus is about 27k, and the Modu corpus about 2000k. Unlike UD, which emphasizes general linguistic characteristics, better represents the characteristics of Korean as an individual language, and serves as a national standard for Korean DP tagging. Also, there are already corpus annotated according to the TTA scheme, so we consider compatibility between them and our benchmark. For this reason, we constructed the {\klueposdp} using the TTA tagset.

\subsubsection{Conclusion}
We build a Korean DP benchmark {\klueposdp}\index{KLUE-DP} consisting of formal news and informal user-generated web data. {\klueposdp} is helpful for developing a DP model that can be used in multiple domains. POS tagging is performed together to improve DP performance, and the tagset and guideline for DP and POS tagging are applied by revising the existing TTA dataset. This guideline is customized to reflect the characteristics of Korean (agglutinative, free word order, etc.), and it also tackles omission of predicates in web data or errors in spacing. We hope that our benchmarks will help in the development of Korean DP models and other natural language processing.

\clearpage
\subsection{Machine Reading Comprehension (\mrc)}
Machine reading comprehension (MRC)\index{MRC (Machine Reading Comprehension)} is a task designed to evaluate models' abilities to read a given text passage and then answer a question about the passage, that is, its ability of comprehension.

Most of existing, widely-used MRC benchmarks are largely in English \cite{clark2019boolq,JoshiTriviaQA2017,MultiRC2018,rajpurkar2016squad,rajpurkar2018know,yang2018hotpotqa, zhang2018record}. Those resources are widely used in evaluating pre-trained language models since it is one of the most intuitive methods for measuring text comprehension. SQuAD 1.1 \cite{rajpurkar2016squad} and SQuAD 2.0 \cite{rajpurkar2018know} are popular evaluation tasks along with GLUE \cite{devlin2019bert, liu2019roberta, clark2020electra, lan2020ALBERT}. BoolQ \cite{clark2019boolq}, ReCoRD \cite{zhang2018record}, and MultiRC \cite{MultiRC2018} are selected as a member of SuperGLUE for rigorous evaluation of language models. Recently, open-domain QA task which can be viewed as an MRC task without a given text passage \cite{tom2019naturalqa, JoshiTriviaQA2017, yang2018hotpotqa, fan2019eli5}, is included in a knowledge-intensive NLP task benchmark \cite{petroni2020kilt}. 

Motivated by those datasets, MRC has become an essential task in NLU benchmarks for various languages such as Indonesian \cite{wilie2020indonlu}, Chinese \cite{xu2020clue}, and Russian \cite{shavrina2020russiansuperglue}. In Korean, however, an appropriate MRC benchmark is not available because existing Korean MRC datasets are either less challenging, limited in access, or simply machine-translated from an English dataset \cite{lim2019korquad1, nia2018aihub, lee2018kquad}. We therefore include MRC in KLUE and create a new challenging Korean MRC benchmark (\kluemrc{}) with the following contributions:

\begin{itemize}[leftmargin=*]
\item \textbf{Providing multiple question types}: In order to evaluate different aspects of MRC capability of models, we provide three question types: paraphrase, multi-sentence reasoning, and unanswerable. We collect questions by following strict guidelines with specific sets of rules for each type.
\item \textbf{Preventing reasoning shortcuts}: We prevent MRC models from exploiting reasoning shortcuts with simple word-matching by enforcing lexical and syntactic variations when workers generate questions. Also, we aim to generate questions which can be answered by considering the full query sentence.
\item \textbf{Multiple passage domains accessible to everyone} : We include news domain passages as well as Wikipedia. To guarantee CC BY-SA license of \kluemrc{}, we made signed contracts with corresponding news providers. 
\end{itemize}

We formulate MRC as a task of predicting the answer span of the question from the given text passage. The input is a concatenated sequence of the question and the passage separated with a delimiter. The output is the start and end positions of the predicted answer span within the passage.

We evaluate models with two metrics: 1) exact match (EM) and 2) character-level ROUGE-W. Note that character-level ROUGE-W is different from the character-level F1 score used in the previous Korean MRC datasets. If the question is unanswerable within the given passage, the model should predict the empty answer string. The motivation of our metrics are described in Section~\ref{sec:mrc-metric}.

\subsubsection{Dataset Construction}


\paragraph{Source Corpora}
First, we collect passages from Korean {\wikipedia} and news articles provided by {\hankyung} and {\acrofan}. {\wikipedia} articles are one of the most commonly used resources for creating MRC datasets. 
We additionally include news articles reporting contemporary social issues to enhance diversity of passages. They are provided by {\hankyung} and {\acrofan}. As news articles are generally copyrighted work, we sign a contract with the news providers to use and redistribute the articles under CC BY-SA license only for building a dataset for machine learning purposes. We believe multi-domain corpus can help MRC models enhance their generalizability.

We preprocess the corpus to collect passages. For \wikipedia{} articles, we remove duplicates in other existing Korean MRC benchmarks (e.g., KorQuAD) for precise evaluation of models. Then, we split each article by its sections to obtain passages. For the news articles, we filter out political articles and articles belonging to categories which have less than 100 articles. We finally gather all preprocessed passages whose length is longer than 512 and shorter than 2048 in characters.

\paragraph{Annotation Protocol}

We annotate questions and answers by giving passages to crowdworkers. We provide a detailed tutorial session to introduce our guidelines. 60 out of 80 workers are selected after a pilot test of creating 15 question-answer pairs with a given passage. The selected workers generate questions and label corresponding answers spans (for Types 1 and 2) or fake answers spans (for Type 3). We use Tagtog annotation toolkit \footnote{https://www.tagtog.net/} for the annotation. 
We assign three inspectors for each question type to validate the generated questions and answers following common and type-specific guidelines.
If the generated question-answer pair fails to pass the inspection, the worker refines it based on the feedback given by the inspector.

\paragraph{1. Common Guidelines} 
We first build common annotation guidelines for crowdworkers. The workers are required to follow the instructions during corresponding question generation and answer span annotation for all three question types as below.

\textbf{A question should:}

\begin{itemize}[leftmargin=*]
\item \textbf{Be natural as a web search query}: Trying to generate challenging questions can lead to the use of unnatural expressions. We guide workers to make natural questions as if they are for web search queries. We care about the generalizability of questions to extend the task to open-domain QA tasks in future work.
\item \textbf{Avoid omission}: Pronoun-dropping is prevalent in Korean, hence it tends to omit the subject or object. The omission can result in ambiguity for finding answers. We explicitly guide workers to keep all grammatical components in generated questions.
\item \textbf{Not copy a phrase in the passage}: Questions might have similar meaning to some phrases in the passage but should not contain exactly the same phrase. This mitigates high word-overlap between questions and the passage as reported in previous work \cite{sen2020wdmlqa}.
\item \textbf{Not refer to external knowledge}: Questions need to be fully understood without any external knowledge. We do not allow workers using their background knowledge or any world knowledge to generate a question. Questions should be derived solely based on the given passage.
\item \textbf{Be meaningful in every part for finding the answer}: Answers should not be found with only a small fraction of the question. We encourage workers to generate questions that require understanding of the whole question text to find an answer. We do not allow the use of expressions that appear only once in the passage because models can easily infer answers without understanding the whole question.
\end{itemize}

\textbf{An answer should:}

\begin{itemize}[leftmargin=*]

\item \textbf{Be a unique entity within a passage}: To clarify what to ask, only a single answer should be inferred from the question. When answers can be represented in various lexical forms, workers should mark all answer spans (e.g. Television, TV).
\item \textbf{Not be the main topic or title}: We aim to prevent a known artifact which the most frequently appeared words within a given passage are likely to be the answer \cite{ko-etal-2020-look}.

\end{itemize}

\paragraph{2. Type-Specific Guidelines} 
Here we elaborate question type-specific guidelines. These guidelines are additionally presented to workers along with the common guideline above. 

\textbf{2.1. Question Paraphrasing (Type 1)}
Type 1 examples focus on paraphrasing the passage sentences when generating questions to reduce word overlap between them. The paraphrasing enables us to validate whether the model can correctly understand the semantics of the paraphrased question and infer the answer \cite{sen2020wdmlqa}. 

\begin{table}[h]
    \centering
    \caption{Paraphrase question type example for Korean and its English translation}
    \begin{adjustbox}{width=0.9\textwidth}
    \begin{tabular}{p{0.1\textwidth}p{0.4\textwidth}p{0.4\textwidth}}
    \toprule
    \textbf{Type} & \textbf{Korean} & \textbf{English (Translated)} \\
    \midrule
        Passage & \small{브르타뉴 공국은 939년 트랑라포레 전투에서 기원을 했으며, 브르타뉴와 노르망디 간에 경계인 쿠에농강에 세워졌다.} & Duchy of Brittany originates in the Battle of Trans-la-Forêt of year 939, and was established on and around the Couesnon River, the boundary of Britanny and Normandy. \\
        \midrule
        \begin{tabular}[c]{@{}l@{}}Good\\ Question\end{tabular} & \small{브르타뉴와 노르망디를 구분짓는 것은?} & What distinguishes Britanny and Normandy? \\
        \midrule
        \begin{tabular}[c]{@{}l@{}}Bad\\ Question\end{tabular} & \small{브르타뉴와 노르망디 간의 경계는?} & What is the boundary of Britanny and Normandy? \\
        \midrule
        Answer & \small{쿠에농 강} & Cuesnon River \\
        \bottomrule
    \end{tabular}
    \end{adjustbox}
    \label{table:mrc_question_type1_example}
\end{table}

In our annotation guide for paraphrased questions, we further prevent workers from generating them by simply shifting the order of subsequent phrases or changing the functional particles. Specifically, workers create questions by following principles:

\begin{itemize}[leftmargin=*]
    \item Either syntactic or lexical variation should be applied to text snippets in the passage. 
        \begin{itemize}
            \item For syntactic variation, reconstructing the structure of the original sentence is preferred. Minimal changes such as swapping order between the nearby phrases is not allowed.
            \item For lexical variation, changing verbs or modifiers is required, and variation of noun phrases is recommended.
        \end{itemize}
    \item More than half of the words in the question should not overlap with the corresponding part (sentence) of the passage.
\end{itemize}

Then we comprehensively check whether the question has low word overlap with the passage. As shown in Table \ref{table:mrc_question_type1_example}, where Cuesnon River is the answer, the question paraphrases the given evidence sentence (“the boundary of Britanny and Normandy”) by providing a new sentence structure. The structure is changed from a wh-question with copula to a question with an intransitive verb. Also, a new term ``distinguishes'' is introduced which replaces the phrase ``is the boundary of''.

\paragraph{2.2. Multiple-Sentence Reasoning (Type 2)}
Type 2 examples focus on making questions requiring multiple-sentence reasoning. Multiple-sentence reasoning requires models to derive answers from the questions by reasoning over at least two sentences in the passage. We focus on evaluating whether an MRC model can infer the answer span by comprehensively aggregating the information spread across the passage. We carefully design annotation guidelines for multiple-sentence reasoning examples. Since \citet{min-etal-2019-compositional} report multiple-sentence reasoning questions can easily fall back to single-sentence reasoning, we aim to avoid such cases by guiding workers to follow the steps below:
\begin{itemize}[leftmargin=*]
    \item \textbf{(Step 1.)} Find at least two statements in the given passage that share some properties. 
    \item \textbf{(Step 2.)} Select one as the answer among the entities regarding the shared properties.
    \item \textbf{(Step 3.)} Generate a question with the selected entity.
\end{itemize}

\begin{table}[h!]
    \centering
    \caption{Multiple sentence reasoning question type example for Korean and its English translation}
    \begin{adjustbox}{width=0.9\textwidth}
    \begin{tabular}{p{0.11\textwidth}p{0.4\textwidth}p{0.4\textwidth}}
    \toprule
    \textbf{Type} & \textbf{Korean} & \textbf{English (Translated)} \\
    \midrule
        Passage & \small{트라야누스 황제 시대(98년-117년)에 로마 제국은 북으로는 스코틀랜드에서 남으로는 아프리카 수단까지, 서로는 포르투갈의 대서양 연안에서 동으로는 카프카스 지방까지 최대 판도를 이룩했다. 오늘날 면적으로 환산하면 현재 미국 면적의 2/3에 달하고 인구도 현 미국의 절반에 약간 안되는 정도로 추산된다. 서기 5세기 경 서로마 제국은 멸망 후 게르만족의 여러 독립 국가로 갈라져 프랑크 왕국, 신성 로마 제국 등 로마의 후계자를 자처하는 여타 서유럽의 정치 세력들이 나타난다.} & Under Trajan(AD 98-AD 117), Roman Empire reached its territorial peak. In terms of the area today, it is estimated to represent two-thirds of the current U.S. area and its population is only slightly less than half that of the current U.S. Plagued by internal instability and attacked by various migrating peoples, the western part of the empire broke up into independent barbarian kingdoms in the 5th century. \\
        \midrule
        Statement A & \small{오늘날 면적으로 환산하면 현재 미국 면적의 2/3에 달하고 인구도 현 미국의 절반에 약간 안되는 정도로 추산된다.} & In terms of the area today, it is estimated to represent two-thirds of the current U.S. area \\ 
        \midrule
        Statement B & \small{트라야누스 황제 시대(98년-117년)에 로마 제국은 북으로는 스코틀랜드에서 남으로는 아프리카 수단까지, 서로는 포르투갈의 대서양 연안에서 동으로는 카프카스 지방까지 최대 판도를 이룩했다.} & Under Trajan(AD 98-AD 117), Roman Empire reached its territorial peak. \\
        \midrule
        \begin{tabular}[c]{@{}l@{}}Shared\\Properties\end{tabular} & \small{로마제국, 트라야누스 황제시대(98년-117년)} & Roman Empire, Trajan(AD 98-AD 117) \\
        \midrule
        Question & \small{로마 제국의 면적이 현재 미국 면적의 2/3에 다다르던 시기는 언제인가?} & When was the area of the Roman Empire two-thirds of that of the current U.S.? \\
        \midrule
        Answer & \small{트라야누스 황제 시대(98년-117년) / 트라야누스 황제 시대 / 98년-117년} & Trajan(AD 98-AD 117) / Trajan / AD 98-AD 117 \\
        \bottomrule
    \end{tabular}
    \end{adjustbox}
    \label{table:mrc_question_type2_example}
\end{table}

The example in Table \ref{table:mrc_question_type2_example} follows the steps described above. Shared properties between the two statements (A: ``In terms of the area today, it is estimated to represent two-thirds of the current U.S. area'', B: ``Under Trajan(AD 98-AD 117), Roman Empire reached its territorial peak.'') are ``Roman Empire'' and ``(Under) Trajan(AD 98 -AD 117)''. Assume we pick ``Trajan(AD 98-AD 117)'' as the answer and generate a question ``When was the area of the Roman Empire two-thirds of that of the current U.S.?''. Neither of the two statements is enough to single out the answer alone. When given the question and statement A alone, it is not possible to resolve an answer between ``Trajan(AD 98-AD 117)'' and ``the 5th century'' which both represent time period in the passage. Since statement B does not include specific information on the territory size, it is also not sufficient to narrow down the answer by itself. Therefore, statement A and B both needs to be aggregated to correctly answer the question.


\paragraph{2.3. Unanswerable Questions (Type 3)}
Type 3 examples are questions unable to be answered within the given passage. We name these as `unanswerable’ questions. In the real world, a question is often unanswerable if a passage is not available. If a model is built upon the premise that an answer always exists within the passage, it would not effectively handle such cases as SQuAD 2.0 \cite{rajpurkar2018know} pointed out. We therefore add unanswerable questions in our benchmark to incentivize models to identify whether a question is answerable or not. 

In the annotation guideline, we present the following principles to generate desirable unanswerable questions. The unanswerable questions should:

\begin{itemize}[leftmargin=*]
    \item \textbf{Be relevant to the passage:} An entity appearing in the given passage should be included in the question. The entity makes the question relevant to the passage. Otherwise the question would be too easy to be determined as unanswerable.
    \item \textbf{Have fake answers within the passage:} A fake answer is plausible but incorrect regarding the corresponding question. The fake answer should exist in the passage as a distractor.
    \item \textbf{Not have correct answers within the passage:} Despite the existence of fake answers, the correct answer for the generated question must not exist in the passage.
\end{itemize}

In our setting, a model is likely to fail to identify the question's unanswerability based on the word overlap between the question and the passage. 
Our question includes entities from the passage which increases the overlap. In that case, an MRC model would choose plausible fake answers that exist in the passage. 

Table \ref{table:mrc_question_type3_example} shows an example of the Type 3 problem. The question includes ``bandwidth'' which also exists in the given passage. 
The question is asking for the bandwidth, so ``1 to 5 Gigabit Ethernet'' would be a plausible but incorrect answer. 
There is no cue about the correct answer to ``bandwidth-hungry server'' from the question in the passage. The model should predict the empty string using the out of the context span range, such as using the [CLS] token from BERT.

\begin{table}[h]
    \centering
    \caption{Unanswerable question type example for Korean and its English translation}
    \begin{adjustbox}{width=0.9\textwidth}
    \begin{tabular}{p{0.1\textwidth}p{0.4\textwidth}p{0.4\textwidth}}
    \toprule
    \textbf{Type} & \textbf{Korean} & \textbf{English (Translated)} \\
    \midrule
        Passage & \small{아직도 대역폭이 많이 필요하지 않은 서버는 1-5기가비트 이더넷을 사용한다.} & Servers that still do not require much bandwidth use 1 to 5 Gigabit Ethernet. \\
        \midrule
        Question & \small{대역폭이 많이 필요한 서버는 어느 대역을 사용하는가?} & Which bandwidth does the bandwidth-hungry server use? \\
        \midrule
        \begin{tabular}[c]{@{}l@{}}Fake\\Answers\end{tabular} & \small{1-5기가비트 이더넷} & 1 to 5 Gigabit Ethernet \\
        \bottomrule
    \end{tabular}
    \end{adjustbox}
    \label{table:mrc_question_type3_example}
\end{table}

\paragraph{Annotation Process}
We employ workers recruited by SelectStar,\footnote{\url{https://selectstar.ai/}} a Korean crowdsourcing platform, to make each type of question and answers for a given passage. For type 1, 28 workers annotate examples and other 3 workers inspect all of them. For types 2 and 3, 19 and 13 workers annotate examples and other 3 and 2 workers validate them, respectively. If the generated question is rejected by the inspectors, it is regenerated based on the feedback.

To meet the ethical standards, examples which contain sexual, violent, hate, bias or any other ethically inappropriate contents are eliminated. We manually re-check all examples at the end of the annotation process. Through the filtering process, we remove 173 examples in total.

\paragraph{Final Dataset}

\kluemrc{} consists of 12,207 paraphrasing-based questions, 7,895 multi-sentence reasoning questions, and 9,211 unanswerable questions. In total, 29,313 examples are made with 22,343 documents and 23,717 passages. As we prioritize creating a challenging dataset for evaluating MRC models, we give more examples to dev and test splits. We set the train/dev/test split ratios as 6:2:2 for robust evaluation, resulting in 17,554 training, 5,841 development and 5,918 test examples. For each split, we randomly sample the examples from each question type independently to balance the number of question types. Tables \ref{table:mrc_dataset_stats} and \ref{table:mrc_dataset_stats2} show the detailed statistics of \kluemrc{}.

\begin{table}[h]
    \centering
    \caption{Number of examples per each dataset split and question types.}
    \begin{tabular}{lcccc}
    \toprule
         & \textbf{Paraphrase} & \textbf{Multi-sentence} & \textbf{Unanswerable} & \textbf{Total}  \\
         & (41.65\%) & (26.93\%) & (31.42\%) & (100.0\%) \\
    \midrule
        Train (60\%) & 7,308 & 4,729 & 5,517 & 17,554 \\
        Dev (20\%) & 2,437 & 1,571 & 1,833 & 5,841 \\
        Test (20\%) & 2,462 & 1,595 & 1,861 & 5,918 \\
    \midrule
        Total (100\%) &12,207 &7,895 &9,211 &29,313 \\
    \bottomrule
    \end{tabular}
    \label{table:mrc_dataset_stats}
\end{table}

\begin{table}[h]
    \centering
    \caption{Statistics of \kluemrc{}.}
    \begin{tabular}{lcccc}
        \toprule
         & \textbf{|Train|} &\textbf{|Dev|} &\textbf{|Test|} &\textbf{Total} \\
        \midrule
        \# Documents &12,174 &5,075 &5,094 &22,343 \\
        \# Passages &13,072 &5,310 &5,335 &23,717 \\
        \# Questions &17,554 &5,841 &5,918 &29,313 \\
        \midrule
        Avg Length of Passage &1,004.62 &1,014.64 &1,010.13 &1,008.10 \\
        Avg Length of Question &29.00 &29.05 &29.01 &29.01 \\
        Avg Length of Answer &6.03 &6.03 &5.82 &5.99 \\
        \bottomrule
    \end{tabular}
    \label{table:mrc_dataset_stats2}
\end{table}

\subsubsection{Evaluation Metrics}
\label{sec:mrc-metric}
The evaluation metrics for \kluemrc{} are 1) exact match (EM) and 2) character-level ROUGE-W (ROUGE) \cite{lin2004rouge}, which can be viewed as longest common consecutive subsequence (LCCS)-based F1 score. EM is the most commonly used metric for MRC tasks, which measures the equality of ground truth and predicted answer string. If there are multiple gold labels, a model can earn score when at least one prediction is matched. In contrast, ROUGE gives a partial score although a model fails to predict exactly matched answer. Due to the characteristics of Korean, an answer span can be located inside a single word, hence subword-level span should be considered. ROUGE calculates F1 score of the length ratio of LCCS to a prediction and the length ratio of LCCS to the ground truth string. In the case when multiple ground truth answer spans have the same meaning but different lexical variations (e.g. TV, Television), we use the maximum ROUGE score among the combinations of answers and the prediction. We do not adopt character-level F1 score (char F1), which is used in all the previous Korean MRC datasets, since it measures character overlap regardless of the order. When a model predicts ``한국의 위인들 (great people in Korea)'' and an answer is ``국한된 범위 (limited scope)'', a metric should give a low score. ROUGE scores 15.38, whereas char F1 gives 54.55 due to the overlap of ``한'', ``국'', and ``위''.

\subsubsection{Analysis}

We investigate \kluemrc{} by comparing with other Korean MRC datasets. KorQuAD 1.0 \cite{lim2019korquad1} and 2.0 \cite{youngmin2020korquad2} are commonly used in Korean MRC research, released with training and evaluation sets separately. However, KorQuAD 2.0 shows a quite different composition in contents compared to ours and KorQUAD 1.0, in specific, HTML tags and tables. Therefore, we conduct comparison only with KorQuAD 1.0 dataset. KorQuAD 1.0 dev set was leveraged because its test set is not available.

\paragraph{Question Difficulty of \kluemrc{} Evaluation Set} 

As we aim to make \kluemrc{} challenging, it is necessary to check the difficulty of test sets. We compare the performance on the evaluation set of ours and KorQuAD 1.0 with the model trained with both datasets in Table \ref{table:mrc_analysis_difficulty}. We fine-tune the model\footnote{We fine-tune pretrained RoBERTa-base model with following hyperparameters: epochs 5, batch size 16, learning rate 3e-5, lr warmup ratio 0.0.} with a collection of the train set from both datasets and test the model on each evaluation set. 

KorQuAD 1.0 train examples (60,407) are almost four times larger than \kluemrc{} (17,554), and it may have resulted in the higher performance for KorQuAD 1.0 dev set. We additionally conduct fine-tuning with a collection of the same amount of both train datasets for fair comparison. We adjust KorQuAD 1.0 train set to fit the size of \kluemrc{} train set by random sampling. Table \ref{table:mrc_analysis_difficulty} shows consistently lower scores of \kluemrc{} compared to KorQuAD 1.0. Thus, our dataset is more challenging regardless of the size of train set.

\begin{table}[h]
    \centering
    \caption{Difficulty comparison between \kluemrc{} test and KorQuAD 1.0 dev set.}
    \begin{tabular}{l cc cc}
    \toprule
        &\multicolumn{2}{c}{\textbf{KLUE + KorQuAD (Full:Full)}} &\multicolumn{2}{c}{\textbf{KLUE + KorQuAD (1:1)}} \\ \cmidrule(lr){2-3} \cmidrule(lr){4-5} 
        \textbf{Evaluation Dataset} &EM &ROUGE &EM &ROUGE \\  \midrule
        KorQuAD 1.0 (Dev) & 86.59 & 94.19 & 85.00 & 93.07 \\
        KLUE-MRC (Test) & \textbf{70.42} & \textbf{75.42} & \textbf{69.75} & \textbf{75.20} \\
    \bottomrule
    \end{tabular}
    \label{table:mrc_analysis_difficulty}
\end{table}

\paragraph{Lexical Overlap}

As high lexical overlap between question and passage can cause shortcut reasoning, reducing lexical overlap is important to build a challenging dataset. To investigate the effects of the proposed strict guidelines and challenging question types, we calculate the lexical overlap of ours and KorQuAD 1.0. The lexical overlap ratio is calculated by dividing the number of common components between question and passage into the number of components in the question. We exclude functional particles such as postposition(조사, josa) and ending components (어미, eomi) when computing the overlap ratio via an open-sourced Korean POS tagger.\footnote{Twitter tagger of KoNLPy \cite{park2014konlpy}.} We observe our lexical overlap ratio is almost 10\%p lower than that of KorQuAD dataset (70\%). For each questions types, Types 1 and 3 show similar ratio in range from 55\% to 59\%. Type 2 exhibits 68\% overlap ratio.

\paragraph{Human Evaluation} 
We evaluate human performance on our {\kluemrc} to measure its difficulty concerning human reading comprehension capabilities. We randomly sample 1,000 examples from our test set and hire three workers to solve them. We select the score of the top scoring worker as human performance. Table \ref{table:mrc_analysis3} reports the comparison between human performance and base model.

\begin{table}[t]
    \centering
    \caption{Comparison of evaluation scores between model prediction and human answer}
    \begin{tabular}{l cc cc cc cc}
        \toprule
         & \multicolumn{2}{c}{\textbf{Paraphrase}} & \multicolumn{2}{c}{\textbf{Multi-sentence}} & \multicolumn{2}{c}{\textbf{Unanswerable}} & \multicolumn{2}{c}{\textbf{Total}} \\ \cmidrule(lr){2-3} \cmidrule(lr){4-5} \cmidrule(lr){6-7} \cmidrule(lr){8-9}
         & EM & ROUGE & EM & ROUGE & EM & ROUGE & EM & ROUGE \\
        \midrule
        \robertabase{} & 67.74 & 75.73 & 65.07 & 73.13 & 72.48 & 72.48 &	68.51 &	74.01 \\
        Human & \textbf{84.18} & \textbf{88.33} & \textbf{87.72} & \textbf{90.91} & \textbf{86.53} & \textbf{86.53} & \textbf{85.90} & \textbf{88.48} \\
        \bottomrule
    \end{tabular}
    \label{table:mrc_analysis3}
\end{table}

\subsubsection{Related Work}
In recent years, significant progress has been achieved in English MRC research with challenging datasets of various question types, including but not limited to: paraphrase, multi-sentence, and unanswerable.

Paraphrased questions have low word overlap between the question and reading passage, which prevents MRC models from exploiting simple word-matching. \citet{trischler2017newsqa} create a NewsQA dataset by generating questions from news headlines and summarizing them via crowdsourcing. They reduce the word overlap by annotating answers on the main articles which are not given during question generation. \citet{saha2018duorc} leverage pairs of plot summaries for the same movies from Wikipedia and IMDb.\footnote{\url{https://www.imdb.com/}} They generate questions from the shorter plots and annotate answers on the longer ones to obtain naturally paraphrased questions. \citet{sen2020wdmlqa} report that datasets with low question-passage overlap will enhance the generalizability of MRC models.

Multi-sentence questions require reasoning over multiple sentences. As a result, they are more difficult compared to single-sentence questions. \citet{JoshiTriviaQA2017} introduce TriviaQA, a dataset of questions from trivia websites. Since they gather evidence passages from various sources (e.g., Wikipedia and the Web), multiple sentences are naturally required for answering the given question. \citet{MultiRC2018} explicitly generate multi-sentence questions on various texts through crowdsourcing and release the MultiRC dataset. The SuperGLUE \cite{wang2019superglue} benchmark adopts the MultiRC dataset as one of its tasks.

Several MRC datasets have incorporated unanswerable questions \cite{yang-etal-2015-wikiqa,trischler2017newsqa,nguyen2016msmarco,rajpurkar2018know,tom2019naturalqa}. \citet{rajpurkar2018know} report performance drop in MRC models when unanswerable questions are included in the dataset.

Compared to MRC research on English, Korean MRC research stands on a small number of existing datasets. The primary benchmark for Korean MRC has been KorQuAD \cite{lim2019korquad1, youngmin2020korquad2}, which adopts the same data collection process as SQuAD 1.0 \cite{rajpurkar2016squad}. However, the model performance on KorQuAD has already exceeded human performance in a short period, leaving little headroom for further research. Moreover, unlike SQuAD, KorQuAD is under CC BY-ND license and does not allow derivative works (e.g., adding unanswerable questions).
AI Hub MRC dataset \cite{nia2018aihub} is based on newspapers and includes unanswerable questions. However, its access is strictly limited to native Korean researchers, prohibiting collaboration even with international researchers residing in Korea.
K-QuAD \cite{lee2018kquad} leverage Google Translate\footnote{\url{https://translate.google.co.kr/}} to translate SQuAD 1.0 \cite{rajpurkar2016squad} into Korean. Since the K-QuAD dataset does not get updated over time, its quality depends on the machine translator's performance at the time of release.
Our \kluemrc{} is different from the existing Korean MRC benchmarks in terms of accessibility-enhance license and more challenging difficulty. 

\subsubsection{Conclusion}
We create a new challenging Korean MRC benchmark named (\kluemrc{})\index{KLUE-MRC}. In order to evaluate different aspects of MRC capabilities, \kluemrc{} includes multi-domain passages and three types of questions: paraphrase, multi-sentence reasoning, and unanswerable. \kluemrc{} shows improvements in question type diversity, difficulty, and lexical overlap compared to existing Korean MRC datasets.

\clearpage
\subsection{Dialogue State Tracking (\dst)}
Building a human-computer conversation system has been increasingly attracting attention, and a task-oriented dialogue system is one type of the dialogue systems~\cite{chen2017survey}. A core module of task-oriented dialogue systems, namely, Dialogue State Tracking (DST)\index{DST (Dialogue State Tracking)} is about predicting the \textit{dialogue states} from a given dialogue context. As illustrated in Table~\ref{table:dst_dialogue_state_example}, dialogue states are sets of slot and value pairs that are relevant categories (e.g. hotel type) and their possible values (e.g. guest house, hotel, motel), respectively. 

Several recent works have considered task-oriented dialogue (TOD) as an important problem of natural language understanding. For instance, DecaNLP~\cite{mccann2018natural} includes a DST, which is a key component of TOD, into one of their benchmark tasks, while DialoGLUE~\cite{mehri2020dialoglue} releases the first task-oriented dialogue benchmark containing various sub-tasks including DST. In light of such, we include DST as a part of the KLUE benchmark. 

The task of dialogue state tracking is to predict slot and value pairs after each user utterance, and the potential pairs are predefined by a task schema and knowledge base (KB), tied to the choice of a scenario. For evaluation, we use joint goal accuracy (JGA) and slot micro F1 score. The JGA checks if all of the predicted slot-value pairs are exactly matched with the ground-truths for every turn, while the slot micro F1 computes f1 score for each slot-value pair independently.\footnote{We adopt the evaluation script of \url{https://github.com/jasonwu0731/trade-dst}}

\begin{table}[h]
\centering
\caption{An example of dialogue state tracking in our {\wizard}. Note that all dialogue states are cumulative in the actual dataset and that we only track states in the user turns.}
    \resizebox{0.85\textwidth}{!}{
    \begin{tabular}{@{}ll@{}}
        \toprule
        \multicolumn{1}{c}{\textbf{Utterances (English Translations)}} &
          \multicolumn{1}{c}{\textbf{Dialogue States}} \\ \midrule
        \textbf{User}: \small{안녕하세요.} (Hello.) & - \\ \midrule
        \begin{tabular}[c]{@{}l@{}}
            \textbf{Sys}: \small{네. 안녕하세요. 무엇을 도와드릴까요?} (Hello. How can I help?)\\ \\ 
            \textbf{User}: \small{\textbf{서울 중앙}에 위치한 \textbf{호텔}을 찾고 있습니다.  외국인 친구도 함께}\\ \small{갈 예정이라서 원활하게 \textbf{인터넷을 사용할 수 있는 곳}이었으면 좋겠어요.}\\ (I'm looking for a \textbf{hotel} at the city \textbf{center}. I'm going with a foreign friend, \\ so easy access to the \textbf{Internet should be available}.)
        \end{tabular} &
        \begin{tabular}[c]{@{}l@{}}
            \textcolor{ProcessBlue}{\textbf{Hotel-area}: center}\\
            \textcolor{ProcessBlue}{\textbf{Hotel-type}: hotel}\\
            \textcolor{ProcessBlue}{\textbf{Hotel-internet}: yes}
        \end{tabular} \\ \midrule
        \begin{tabular}[c]{@{}l@{}}
            \textbf{Sys}: \small{네 확인해보겠습니다. 혹시 추가로 필요하신 사항이 있으실까요?}\\ (Sure, let me check. Do you need anything else?)\\ \\
            \textbf{User}: \small{음... 예약 인원은 총 \textbf{8명}이고요. 아, \textbf{가격대는 크게 상관 없습니다}.}\\ (Hmm.. I want to reserve for \textbf{8 people}. Ah, the \textbf{price range doesn't matter}.)
        \end{tabular} &
        \begin{tabular}[c]{@{}l@{}}
            \textbf{Hotel-area}: center\\
            \textbf{Hotel-type}: hotel\\
            \textbf{Hotel-internet}: yes\\
            \textcolor{ProcessBlue}{\textbf{Hotel-book people}: 8}\\ 
            \textcolor{ProcessBlue}{\textbf{Hotel-price range}: dontcare}
        \end{tabular} \\ \midrule
        \begin{tabular}[c]{@{}l@{}}
            \textbf{Sys}: \small{네, 확인 감사합니다. 숙박을 원하시는 요일과 기간 같이 확인 부탁드립니다.}\\ (Great, thanks for confirming. Please let us know when and how long \\you want to stay.)\\ \\
            \textbf{User}: \small{아, 중요한 걸 깜빡했네요. \textbf{일요일}에 \textbf{2일}간 예약하고 싶습니다.}\\ (Right, I forgot an important thing. I would like to book for \textbf{two days} \\ from \textbf{Sunday}.)
        \end{tabular} &
        \begin{tabular}[c]{@{}l@{}}
            \textbf{Hotel-area}: center\\
            \textbf{Hotel-type}: hotel\\
            \textbf{Hotel-internet}: yes\\
            \textbf{Hotel-book people}: 8\\
            \textbf{Hotel-price range}: dontcare\\
            \textcolor{ProcessBlue}{\textbf{Hotel-book day}: Sunday}\\
            \textcolor{ProcessBlue}{\textbf{Hotel-book stay}: 2}
        \end{tabular} \\
        \bottomrule
    \end{tabular}
    }
    \label{table:dst_dialogue_state_example}
\end{table}

\subsubsection{Dataset Construction}

Our dataset construction protocol is a modified version of the Wizard-of-Oz framework (WOZ) \cite{kelley1984iterative}, which is a widely used paradigm for building dialogue datasets. The WOZ setting is a particular type of human-to-human dialogue collection, which employs two people that either takes a user and a system role. However, it is arguably time-consuming, complex, and expensive~\cite{byrne2019taskmaster1}. To overcome the limitations, we adopt `Self-dialog' scheme which requests a single worker to play both user and system roles~\cite{byrne2019taskmaster1}. In addition, we introduce a new design choice to obtain a more precise dialogue dataset.

\paragraph{Overview} We construct {\wizard} by following five steps: 1) define a task schema, 2) create knowledge base (KB), 3) design an annotation system, 4) collect and annotate a dataset, and 5) finalize the dataset. Different from the other datasets in our benchmarks, {\wizard} is not following an ordinary protocol: collecting raw corpus followed by annotating labels. Rather, we collect dialogues (raw corpus) and their corresponding dialogue states (labels) at the same time.

\begin{table}[t]
\label{task_schema}
    \centering
    \caption{Task schema for all five domains in Wizard of Seoul (\wizard) which shows names of the domain and their slots. Star$^\star$, asterisk$^*$, cross$^\dagger$, doubly-crosses$^\ddagger$ denote required, boolean type, booking-related, and requestable after booking slot, respectively.}
    \label{table:wos_schema}
    \begin{tabular}{lll}
        \toprule
        \textbf{Domains} & \textbf{Informable Slots} & \textbf{Requestable Slots}  \\
        \midrule
        Hotel & \makecell[l]{name, type$^\star$, area$^\star$, price range$^\star$,\\ book day$^\dagger$, book stay$^\dagger$, book people$^\dagger$, \\ walkability$^*$, parking$^*$, internet$^*$,\\ breakfast$^*$, smoking$^*$, fitness$^*$,\\ swimming pool$^*$, spa$^*$} & \makecell[l]{rating, nearby station,\\ minutes walk from station,\\ address, phone number,\\ business hour, reference number$^\ddagger$} \\ \midrule
        Restaurant & \makecell[l]{name, type$^\star$, area$^\star$, price range$^\star$,\\ book day$^\dagger$, book time$^\dagger$, book people$^\dagger$, \\ alcohol$^*$, walkability$^*$, parking$^*$,\\ internet$^*$, smoking$^*$, outdoor table$^*$} & \makecell[l]{rating, nearby station,\\ minutes walk from station,\\ address, phone number,\\ business hour, last order time,\\ representative menu, reference number$^\ddagger$} \\ \midrule
        Attraction & \makecell[l]{name, type$^\star$, area$^\star$,\\ walkability$^*$, parking$^*$, heritage$^*$,\\ educational$^*$, scenic$^*$, cultural$^*$} & \makecell[l]{rating, nearby station,\\ minutes walk from station,\\ address, phone number,\\ business hour, entrance fee} \\ \midrule
        Taxi & \makecell[l]{leave at$^\star$, departure$^\star$,\\ arrive by, destination$^\star$, type} & \makecell[l]{phone number, cost, duration} \\ \midrule
        Metro & \makecell[l]{leave at, departure$^\star$, destination$^\star$} & \makecell[l]{departure line, destination line,\\ arrive by, cost, duration,\\ transfer, optimal path} \\
        \bottomrule
    \end{tabular}
\end{table}

\paragraph{1. Defining Task Schema}

We first define a task schema which expresses the scenario of task-oriented dialogue. Our task schema consists of slots across five domains (hotel, restaurant, attraction, taxi, and metro) as shown in Table~\ref{table:wos_schema}. Typically, slots are categorized into informable slots and requestable slots. The informable slots cover properties which can constrain a user goal\footnote{A user goal is what the worker playing user should follow as shown in Table~\ref{table:dst_goal_instrction_example}.} such as ``price range'', ``area'', and ``booking day''. The requestable slots provide additional information that a user may ask, but not necessarily need to be specified as a user goal constraint. A typical example of a requestable slot is ``phone number'', which a user may ask for, but would not work to narrow down the probable candidates of the goal \cite{henderson-etal-2014-second, henderson2014third}.

Based on this schema, we include additional attributes to the informable and requestable slots to provide an easy-to-operate annotation system and easy-to-follow guidelines. A slot could have one or more attributes among whether it is 1) boolean type, 2) required or not (\textit{Required}), 3) related to booking (\textit{Booking-related}), and 4) only available after booking is confirmed (\textit{Requestable after booking}. e.g., reference number). The boolean type slots can have either \textit{yes} or \textit{no} as their values, such as ``Parking (availability)'' and ``(has) swimming pool''. Such boolean type values do not appear in the dialogue context explicitly. In other words, they have abstractive properties. A model which understands abstractive properties is desirable, so we have much more boolean type slots than MultiWOZ~\cite{budzianowski2018multiwoz}; {\wizard} has 20 boolean slots across the domains while MultiWOZ includes only 2. Meanwhile, the required slots have to be specified with values in order to fill out a user intent. This helps us to simulate an actual service scenario in which an agent is not allowed to take the next steps without specifying required values \cite{rastogi2020towards}.

\begin{table}[t]
    \centering
    \caption{An example of goal instruction. Unlike MultiWOZ, we present all instructions from the beginning to prevent ordering bias as in CoCo~\cite{li2020coco}. The booking-related slots, ``restaurant-book time (22:41)'' and ``restaurant-book day (Wednesday)'' appear before a confirmation of KB entity.}
    \begin{tabular}{p{0.4\textwidth}p{0.4\textwidth}}
    \toprule
    \textbf{Korean} & \textbf{English (Translated)} \\
    \midrule
        \small{당신은 오늘 \textbf{22:41}에 \textbf{서울 중앙}에서 식사할 계획을 가지고 있습니다.
        아참 오늘은 \textbf{수요일} 입니다.
        그런 곳을 찾았다면 먼저 \textbf{대표 메뉴}를 확인하세요.
        그리고 나선 \textbf{1}명으로 예약 거세요.
        예약 이후엔 \textbf{영업 시간}을 문의하시구요.
        그리고 나선 \textbf{식당 근처}에서 잘 곳을 찾아야 합니다.
        그 곳은 반드시 \textbf{흡연이 불가}해야 합니다.
        찾았다면 \textbf{같은 요일}에 예약하세요.
        \textbf{같은 인원}으로 \textbf{4}일간 머물러야 합니다.
        예약에 성공했다면 \textbf{예약 번호}를 묻고, \textbf{흡연 가능 유무}를 더블 체크하세요.
        그런 다음 마지막으로 택시를 하나 부르세요.
        \textbf{식당}에서 \textbf{숙소}로 향해야 합니다.
        찾았다면 \textbf{소요 시간}을 문의하세요.}
        & \small{You have a plan to eat in the \textbf{center of Seoul} at \textbf{22:41} today.
        Oh, today is \textbf{Wednesday}.
        If you find such a place, first check the \textbf{representative menu}.
        Then, make a booking for \textbf{1} person.
        After booking, inquire about the \textbf{business hour}.
        Then, you have to find a hotel to sleep in \textbf{near the restaurant}.
        The restaurant must be \textbf{non-smoking}.
        If you find it, book on the \textbf{same day}.
        You must stay for \textbf{4} days as the \textbf{same number of people}.
        If the booking is done, ask for the \textbf{reference number} and double-check the \textbf{smoking allowed}.
        Then, finally, call a taxi.
        You have to go to the \textbf{hotel} from the \textbf{restaurant}.
        If you call the taxi, inquire about the \textbf{duration}.} \\
        \bottomrule
    \end{tabular}
    \label{table:dst_goal_instrction_example}
\end{table}

\paragraph{2. Creating Knowledge Base}

We construct a knowledge base (KB) based on the task schema of each domain to obtain a set of predefined realization candidates of a user's goal. For the hotel and restaurant domains, we manually create virtual instances, whereas for the attraction and metro domains, we leverage real names (e.g. Gangnam Station or Namsan Tower) collected from the web. On the other hand, for the taxi domain, we do not define instances in advance, but dynamically generate the instances during the dialogue collection as in MultiWOZ \cite{budzianowski2018multiwoz}. With ethical considerations in mind, any personally identifiable information (PII, e.g., phone number, address) is replaced with randomly generated instances using \texttt{faker}.\footnote{\url{https://faker.readthedocs.io}} Table~\ref{table:wos_kb} shows the KB statistics for each domain.

\begin{table}[h!]
    \centering
    \caption{Statistics of Knowledge Base in {\wizard}.}
    \begin{tabular}{lcc}
        \toprule
        \textbf{Domain} & \textbf{\# Instances} & \textbf{\# Slots}  \\
        \midrule
        Hotel & 101 & 19  \\
        Restaurant & 56 & 20 \\
        Attraction & 100 & 17 \\
        Taxi & - & 8  \\
        Metro & 3,306 & 10 \\
        \bottomrule
    \end{tabular}
    \label{table:wos_kb}
\end{table}

\paragraph{3. Designing the Annotation System}
In this section, we describe the annotation platform we used for collecting data from both the User-side and System-side.


\paragraph{3.1. User Side}
We provide a goal instruction for a user side role. The instruction includes descriptions of a user’s specific goal with corresponding slot values in natural language. It also contains the user's context including persona for variety of dialogues. A user is asked to generate utterances following the instruction. An example is shown Table~\ref{table:dst_goal_instrction_example}.

To devise a multi-domain dialogue scenario where slots are shared across multiple domains, we include \textit{domain transition} in the instruction \cite{budzianowski2018multiwoz, zhu2020crosswoz}. For example, in case of a user aiming to book a hotel, the user might seek information about transportation (taxi, metro, etc.) to get there. In this dialogue, initial domain is changed to another (hotel to taxi/metro). It is more challenging compared with single domain in terms of dialogue state tracking. Because user could express their goal implicitly where values should be inferred by co-referencing other values of preceding domains.

The goal instructions are realized by the templates containing placeholders for goal constraining slots and their values. We design diverse templates for each domain, in order to cover various scenarios of the dialogues. Like MultiWOZ, the goal templates include a series of subgoals with corresponding slots. We carefully design the sentences to promote lexical entailment or co-referencing during conversation, which can be naturally observed in the user context or during the domain transition. When filling a template to complete instructions, we randomly assign the instances from KB built upon the domain-specific task schema to the given placeholders. The values of the instruction should be specifically mentioned by a user during a conversation. Each trackable slot either has valid values, \textit{None} or \textit{Dontcare}.\footnote{\textit{Dontcare} means a user has no preference and \textit{None} means a user is yet to specify a valid value for given slot \cite{henderson-etal-2014-second, henderson2014third}.}

To properly evaluate a model's generalization ability, we further add counterfactual goals and introduce unseen KB instances during the process \cite{li2020coco}. To add new goal instruction based on the current slot distribution, we keep monitoring the slot value frequency and co-occurrence over slots during the construction. Specifically, we add new goal instructions which cover infrequent slot values or rarely co-occurring combination among slots. For example, when ``(hotel-parking, no)'' is infrequent pair in the as-is distribution, we promote it to appear in dialogues by designing goal instruction including it as constraint. Moreover, we differentiate KB instances between particular subset of dataset (train and dev/test set) to simulate realistic scenario regarding unseen slot values in the test time.

\begin{figure}[t] 
\centering
\label{fig:wos_gui}
\includegraphics[width=0.8\textwidth]{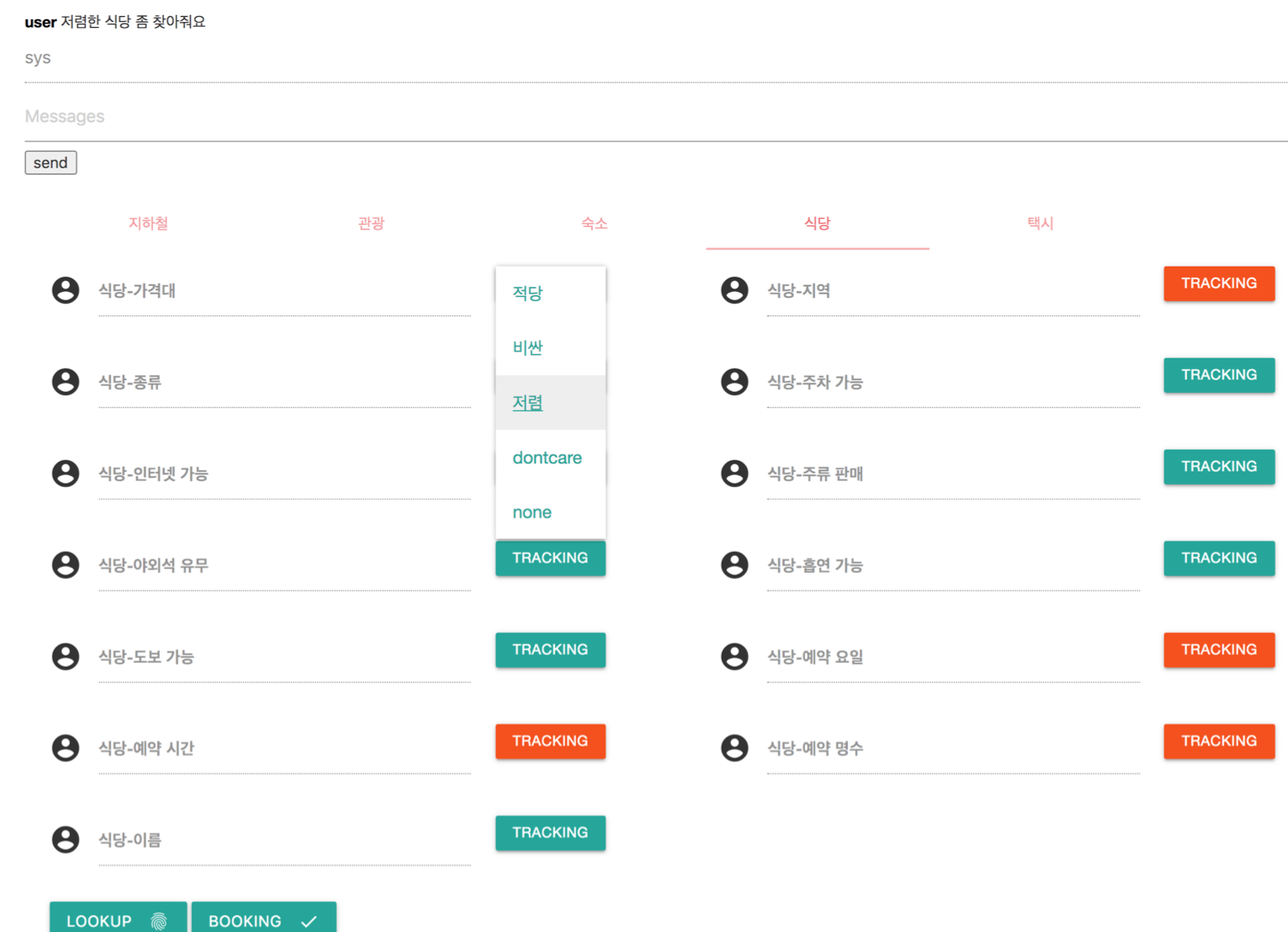}
\caption{Graphical web interface for system side worker.}
\end{figure}

\paragraph{3.2. System Side}

The role of a system side worker (wizard) is to 1) annotate dialogue states of user utterances while 2) generate responses by accessing to the KB if necessary, for every turn. First, the wizard is asked to fill in appropriate slot values inferred from the current dialogue context. If the word uttered by the user is not clear to directly map to a specific slot value, the wizard should clarify the meaning of the word first and then fill the slot when the word has the same meaning as the value. The annotation of a dialogue state is an explicit action of understanding a user request to fully focus on providing the required information. Then, the wizard generates response either to request or to convey information. If the values of required slots are absent, the system side worker is allowed to ask for the missing values to the user. Otherwise, the system provides the user with the adequate information. We enable the system to query the external knowledge base, if needed. When there are more than three search results, the system worker could request more details or recommend one among them.

To support the wizard to perform such complicate work effectively and efficiently, we provide a graphical web interface with a newly introduced feature: dropdown components (Figure~\ref{fig:wos_gui}). Dropdown interface enables the system side worker can choose a value from a list of pre-populated candidates. We present most probable value candidates based on the goal instruction and domain-specific knowledge base, since a dropdown might become worthless when too many options are presented to workers. This procedure naturally prevents several type of annotation errors, such as \textit{multi-annotations}, \textit{mis-annotations}, \textit{typos}, and \textit{value canonicalization} reported in MutiWOZ 2.1~\cite{eric2019multiwoz}.

\paragraph{4. Dataset Construction}

We adapt `Self-dialog' scheme inspired by taskmaster-1~\cite{byrne2019taskmaster1} to efficiently collect diverse dialogue dataset while reducing the cost and time. Self-dialog is effective to collect various dialogue data. By having both roles, a worker freely controls a flow of dialogue such as the order of slot occurrence in user utterances and recommendation of system responses. This also leads workers to speak their own styles naturally such that different personas are included. However, we found annotation errors in pilot phase such as \textit{early-markup} (system pre-fills the values before receiving the values from the user) and \textit{delayed-markup} (the system fills the values behindhand the proper turn) errors. We further improve the scheme by utilizing explicit turn-switching between user and system roles with providing error correction interface. 

To elaborate, we train and select trustworthy workers to participate in the main collection process. Prior to the main phase, we conduct several pilot studies with crowdworkers to avoid aforementioned early-markups and delayed-markup errors, and to generate more realistic dialogues including some miscommunications. We also implement an explicit turn-switching between the two roles, in order to immerse themselves in a user/system role, which mitigates overly reduced miscommunications as well. Throughout the pilot, we finally employ 15 selected workers who can effectively handle such issues.

\paragraph{Final Dataset}

Table~\ref{table:wos_stat} shows statistics of our dataset. {\wizard} contains overall 10,000 dialogues with 146,692 turns across 5 domains. The evaluation (dev/test) set specifically includes dialogues with counterfactual goals and unseen KB instances against train set. The dev/test set contains 294 and 361 counterfactual goal-based dialogues, respectively. All the splits contain sufficient number of dialogues with domain transition.

\begin{table}[t!]
    \centering
    \caption{Statistics of Wizard-of-Seoul (\wizard{}).}
    \label{table:wos_stat}
    \begin{tabular}{lcccc}
        \toprule
        & \textbf{|Train|} & \textbf{|Dev|} & \textbf{|Test|} & \textbf{Total}  \\
        \midrule
         \# Dialogues & 8,000 & 1,000 & 1,000 & 10,000 \\
         \# Single Domain Dialogues & 1,806 & 263 & 226 & 2,295 \\
         \# Multi Domain Dialogues  & 6,194 & 737 & 774 & 7,705 \\
         \# Counterfactual Dialogues  & 0 & 294 & 361 & 655 \\ \midrule
         \# Total Turns & 117,584 & 14,448 & 14,660 & 146,692  \\
         \# Total Tokens  & 899,450 & 114,169 & 114,914 & 1,128,533 \\ \midrule
         Avg Turns per Dialogue & 14.70 & 14.45 & 14.66 & 14.67 \\
         Avg Tokens per Turn & 7.65 & 7.90 & 7.84 & 7.69 \\
        \bottomrule
  \end{tabular}
\end{table}

\subsubsection{Evaluation Metrics}
\label{sec:dst-metric}

For evaluation metrics for \wizard{} is 1) joint goal accuracy (JGA) and 2) slot micro F1 score. JGA measures the proportion of exactly matched dialogue state which consists of a set of slot-value pairs with the ground truth dialogue state among the total number of dialogue turns. Slot micro F1 score is an average of micro F1 scores in each turn. For each turn, micro F1 score is defined as the harmonic mean of precision and recall in terms of predicted slot-value pairs and ground-truth pairs. Note that the slot micro F1 score ignores when value of the ground truth is ``\texttt{None}''.

\subsubsection{Analysis}
When splitting the train and dev/test set based on the counterfactual goals and unseen KB instances, like \citet{li2020coco}, we observe performance drop as shown in Table~\ref{table:data_split}. This demonstrates that the counterfactual goal makes \wizard{} more challenging. 

\begin{table}[t!]
    \centering
    \caption{Comparison regarding data split strategy. Random is splitting train, dev and test set randomly. The CF-goal indicates they are included in dev and test set with unseen KB instances.}
    \label{table:data_split}
    \begin{tabular}{lc}
        \toprule
        \textbf{Domain Split} & \textbf{Joint Goal Accuracy} \\
        \midrule
        Random & 57.53 \\
        CF-goal & 47.38 \\
        \bottomrule
  \end{tabular}
\end{table}

\subsubsection{Related Work}

Wizard-of-Oz (WOZ) \cite{kelley1984iterative} is a popular scheme in dialogue collection. In fact, conventional WOZ setting allows to collect various type of dialogues by employing role-playing of two humans. Each human should choose a role between the two: \textit{user} and \textit{system}. As taking a role, dialogues are collected by turn-taking generation of utterances with background information provided in advance. In the case of building a task-oriented dialogue, \textit{goal} is given to a \textit{user} while \textit{knowledge base} is allowed to be accessed to \textit{system}. The \textit{system} can use its \textit{knowledge base} when responding to \textit{user}’s request.

Many dialogue datasets closely follow WOZ settings \cite{wen2016network, asri2017frames, eric2017key}, however, it costs a lot of time and money because two crowdworkers must be matched at the same time and successfully play each role, which prevents collecting dialogues at scale. We refer this limitation to `worker coexistence constraints’. To overcome the limitation, MultiWOZ \cite{budzianowski2018multiwoz} slightly change this conventional WOZ to \textit{asynchronously} collect dialogues turn-by-turn from crowdworkers, which allows different workers to play the same \textit{user} or \textit{system} in a single dialogue. This approach costs less but error-prone because every worker must adapt to dialogue already progressed so that their response might be incoherent to the previous context \cite{eric2019multiwoz}. Recently, CrossWOZ and RiSAWOZ thus pair only selected trustworthy workers to collect dialogues in \textit{synchronous} manner as suggested in the WOZ settings to keep the annotation quality despite the high construction cost \cite{zhu2020crosswoz, quan2020risawoz}.


\citet{byrne2019taskmaster1} also argue that conventional WOZ settings are time consuming, complex and expensive, requiring considerable technical implementation as well as administrative procedures to train and manage both agents and crowdsourced workers, accordingly suggesting Self-dialog as an alternative. Self-dialog is a collection scheme in which workers write the entire dialogue playing both user and system roles. To demonstrate their idea, \citet{byrne2019taskmaster1} build a large-scale dialogue datase, Taskmaster-1, which is built upon the Self-dialog which stands on the WOZ schemes: 1) two people playing user and system roles (conventional WOZ setting) and 2) one person playing both roles (Self-dialog). As a result, Self-dialog can remedy the cost of `worker coexistence constraint’ effectively with avoiding incoherent dialogue generation caused by asynchronous dialogue collection. However, there is a tendency that little miscommunication occurs in the dialogue which compared to the real world conversations because the same person produces utterances in both roles, which might lead to a gap from reality.

Some researchers further tries to employ only machines to create such dialogues to maximize cost-efficiency. It builds the dialogues on top of a simulator which is able to create utterances by turn automatically from elaborately designed rules and given task schema. The simulator first generates psuedo-dialogue, then adopts crowdsourcing to paraphrase them to the natural utterances \cite{shah2018building, rastogi2020towards}. It requires much less human effort, however, heavily relies on the simulator.

Meanwhile, previous works exists addressing robust evaluation of DST models. According to CoCo \cite{li2020coco}, state-of-the-art DST models are not robust to realistic scenarios since they scarcely appear in the train data. As the name CoCo (controllable counterfactuals) suggests, it generates infrequent but realistic dialogues based on the predefined slot-value pairs. They show that even a state-of-the-art DST model’s performance drops significantly when they are evaluated on such dialogues including counterfactual goals. It means the current TOD benchmarks should be improved in terms of robustness to unseen but realistic scenarios.

As for Korean, there is a task-oriented dialog dataset is provided by National Information Society Agency (NIA). It covers about 10 domains related to the civil complaints and consists of more than 500k dialogues. The utterances are divided into four types: 1) a main question that a user asks, 2) a sub question that a system could ask for clarification, 3) a user answer, and 4) a system answer. Additionally, user intents are annotated and entities are extracted from each utterance. We find that this dataset does not follow any of the aforementioned settings; there is no dialogue state represented as slot-value pairs, only regarding single turn judgement. It also lacks information about task schema and redistribution is restricted which does not satisfy our accessibility principle, which motivates us to newly create a DST benchmark.

\subsubsection{Conclusion}
We introduce Wizard-of-Seoul ({\wizard})\index{WoS (Wizard of Seoul, KLUE-DST)}, the first large-scale Korean multi-domain task-oriented dialogue dataset that simulates conversations between Seoul tourists and travel agents. We adapt `Self-dialog' for efficiently scaling up of dialogue collection scheme. In addition, consideration on annotation interfaces (drop-down menu and turn-switching) mitigates erroneous cases and diverse goal instructions including counterfactual ones promote each conversation to be more natural and challenging. We hope that {\wizard} sparks various future dialogue research in Korean and also offers valuable insights to pushing forward end-to-end dialogue modeling.


\clearpage
\section{Pretrained Language Models}
In order to facilitate further research using KLUE, we provide strong baselines for all the benchmark tasks within it. As a part of this effort, we pretrain and release large-scale language models for Korean, which we hope would reduce the burden of retraining large-scale language models from individual researchers. 
More specifically, We pretrain language models (PLM), including BERT \cite{devlin2019bert} and RoBERTa \cite{liu2019roberta}, from scratch.

\subsection{Language Models}

We pretrain multiple Korean language models while varying training configuration. This enables us to explore effective settings for pretraining Korean models and further establish simple yet effective baseline models for KLUE. We train \kluebert{}\index{KLUE-BERT} and \klueroberta{}\index{KLUE-RoBERTa}. We vary the choice of a pretraining corpus, preprocessing procedure, tokenization strategy, and other training configurations.

\begin{table}[h]
    \centering
    \caption{Statistics of the pretraining corpus.}
    \begin{adjustbox}{width=0.9\textwidth}
    \begin{tabular}{@{}lcccccc@{}}
    \toprule
     & \textbf{\modu{}} & \textbf{\CommonCrawl{}} & \textbf{\namuwiki{}} & \textbf{\newscrawl{}} & \textbf{\petition{}} & \textbf{Total} \\ \midrule
    \# Sentences & 167M & 103M & 14M & 183M & 5.2M & \textbf{473M} \\
    \# Words & 1,892,814,395 & 1,593,887,022 & 265,203,602 & 2,716,968,038 & 50,631,183 & \textbf{6,519,504,240} \\ \midrule
    size (GB) & 18.27 & 15.46 & 2.52 & 25.87 & 0.53 & \textbf{62.65} \\ \bottomrule
    \end{tabular}
    \label{table:pretraining_corpus}
    \end{adjustbox}
\end{table}

\paragraph{Pretraining Corpora}
\label{sec:pretraining-corpora}

We gather the following five publicly available Korean corpora from diverse sources to cover a broad set of topics and many different styles. We combine these corpora to build the final pretraining corpus of size approximately 62GB. See Table~\ref{table:pretraining_corpus} for overall statistics:

\begin{itemize}[leftmargin=*]

    \item \textbf{{\modu}} : \textit{Modu}\footnote{
    A transliteration of a Korean word `모두' which means `Everyone'.
    } 
    Corpus \cite{nikl2020corpora}\index{MODU (Modu Corpus)} is a collection of Korean corpora distributed by National Institute of Korean Languages.\footnote{\url{https://corpus.korean.go.kr/}} It includes both formal articles (news and books) and colloquial text (dialogues).

    \item \textbf{{\CommonCrawl}} : CC-100\footnote{
    \url{http://data.statmt.org/cc-100/}
    }
    is the large-scale multilingual web crawled corpora by using CC-Net \cite{wenzek2020ccnet}\index{CC-100-Kor}. This is used for training XLM-R \cite{conneau2019unsupervised}. We use the Korean portion from this corpora.
    
    \item \textbf{{\namuwiki}} : \namuwiki{} is a Korean web-based encyclopedia, similar to Wikipedia, but known to be less formal. Specifically, we download the dump created on March 2nd, 2020.\footnote{
    \url{http://dump.thewiki.kr}
    }\index{NAMUWIKI}
    
    \item \textbf{{\newscrawl}} : \newscrawl{} consists of 12,800,000 news articles published from 2011 to 2020, collected from a news aggregation platform.\index{NEWSCRAWL}
    
    \item \textbf{{\petition}} : Petition is a collection of public petitions posted to the Blue House asking for administrative actions on social issues. We use the articles in the Blue House National Petition\footnote{
    \url{https://www1.president.go.kr/petitions}
    } 
    published from August 2017 to March 2019.\footnote{
    \url{https://ko-nlp.github.io/Korpora/en-docs/corpuslist/korean_petitions.html}
    }\index{PETITION}

\end{itemize}

\paragraph{Preprocessing}
\label{sec:plm-corpora-prep}

We filter noisy text and non-Korean text using the same methods from Section~\ref{sec:corpus-preprocessing}. Each document in the corpus is split into sentences using C++ implementation (v1.3.1.) of rule-based Korean Sentence Splitter (KSS).\footnote{
\url{https://github.com/likejazz/korean-sentence-splitter}
} 
For \CommonCrawl{} and \newscrawl{}, we keep sentences of length greater than equal to 200 characters, as a heuristics to keep well-formed sentences. We then remove sentences included in our benchmark task datasets, using BM25 as a sentence similarity metric \cite{robertson1995okapi}.

\paragraph{Ethical Considerations}

Because we collect and use as much publicly available data as possible for pretraining, these corpora often contain undesirable social biases. Furthermore, we noticed earlier quite a bit of PII in these corpora, although they were all publicly available. Both of these are problematic. Social biases in the corpus may result in a language model that learns such biases. PII in the corpus may be memorized by a language model and can subsequently be retrieved by adversarial attacks~\cite{carlini2020extracting}. 

We do not filter out socially biased contents nor hate speech for three reasons. 
First, manual inspection is infeasible for this large-scale pretraining corpora. Second, it is a challenging problem on its own to automatically detect socially biased contents or hate space, as both of these highly depend on the context in which they appear \cite{NEURIPS2020_1b84c4ce}. Lastly, being blind to such harmful contents prevents the future use of a language model for detecting and correcting these harmful contents, such as using it as an anti-expert~\cite{liu2021onthefly}. We expect future research on the pretrained language models we release to focus on how to detect and correct biases encoded in these models and on how to debias them, as has been recently demonstrated by  \citet{cheng2021fairfil}.

In contrast, we pseudonymize PII in our corpora as much as possible. We detect 16 personal data types using regular expressions based on the guideline from the Korea Internet and Security Agency (KISA).\footnote{
\url{https://www.kisa.or.kr/public/laws/laws2_View.jsp?cPage=1&mode=view&p_No=282&b_No=282&d_No=3}
} It is relatively easy to pseudonymize PII while keeping linguistic patterns, since the selected PII has standardized pattern. 
We then replace the original information, using either the \url{faker} library\footnote{
\url{https://github.com/joke2k/faker}
} 
or random generation based on the pattern. As a result, we pseudonymize 1.2\% of the pretraining corpora. Details are illustrated in Table~\ref{tab:pseudonymization}. 

\begin{table}[t!]
    \centering
    \caption{Our pseudonymization methods and examples. The examples are from \url{faker} library documentation or the public.}
    \begin{adjustbox}{width=0.75\textwidth}
    \begin{tabular}{@{}lll@{}}
    \toprule
    \textbf{Private Information} & \textbf{Pseudonymization} & \textbf{Pseudonymised Example} \\ \midrule \medskip
    \textbf{Telephone Number} & Faker & 055-604-8764 \\ \medskip
    \textbf{Social Security Number} & Faker & 600408-2764759 \\ \medskip
    \textbf{Foreign Registration Number} & Faker & 110527-1815659 \\ \medskip
    \textbf{Email Address} & Faker & agweon@example.org \\ \medskip
    \textbf{IP Address} & Faker & 166.186.169.69 \\ \medskip
    \textbf{MAC Address} & Faker & c5:d7:14:84:f8:cf \\ \medskip
    \textbf{Mention(@)} & Faker & @gildong \\ \medskip
    \textbf{Address} & Random Number Generation & 경상북도 성남시 서초대64가 \\ \medskip
    \textbf{Bank Account Number} & Random Number Generation & 110-245-124678 \\ \medskip
    \textbf{Passport Number} & Random Generation & M123A4567 \\ \medskip
    \textbf{Driver’s License} & Random Number Generation & 11-17-174133-01 \\ \medskip
    \textbf{Business Registration Number} & Random Number Generation & 123-45-67890 \\ \medskip
    \textbf{Health Insurance Information} & Random Number Generation & 1-2345678901 \\ \medskip
    \textbf{Credit or Debit Card Number} & Random Number Generation & 1234-5678-9012-3456 \\ \medskip
    \textbf{Vehicle Registration Place} & Random Generation & 55구 1601 \\
    \textbf{Homepage URL} & Random Generation & www.example.com \\ \bottomrule
    \end{tabular}
    \end{adjustbox}
    \label{tab:pseudonymization}
\end{table}

\paragraph{Tokenization}

We design and use a new tokenization method, \textit{morpheme-based subword} tokenization. When building a vocabulary, we pre-tokenize a raw text into morphemes using a morphological analyzer, and then we apply byte pair encoding (BPE) \cite{sennrich-etal-2016-neural} to get the final vocabulary. For morpheme segmentation, we use \texttt{Mecab-ko},\footnote{
\url{https://bitbucket.org/eunjeon/mecab-ko}
} 
MeCab \cite{kudo2006mecab} adapted for Korean, and for BPE segmentation, we use the wordpiece tokenizer from \texttt{Huggingface Tokenizers} library.\footnote{
\url{https://github.com/huggingface/tokenizers}
} 
We specify the vocabulary size to 32k.
After building the vocabulary, we only use the BPE model during inference, which allows us to tokenize a word sequence by reflecting morphemes without a morphological analyzer. This improves both usability and speed. Examples are presented in Table~\ref{table:qualitative_tokenzation}.

The motivation behind this method is that Korean is an agglutinative language, which is to say, a word is a constitution of morphemes - stems and affixes. The morphemes tend to remain unchange on different unions, and the boundary is generally clear. Although BPE has been widely used across many languages due to its effectiveness, it struggles to identify morphemes correctly as demonstrated in Table~\ref{table:qualitative_tokenzation}.

\begin{table}[h]
    \centering
    \caption{An input text ``조경현은 인공지능 분야의 저명한 연구자이다. (Kyunghyun Cho is a prominent AI researcher.)'' is segmented with various tokenization strategies. We denote slash (/) as a token separator. The mBERT tokenizer \cite{devlin2019bert} splits the input text into nearly in characters resulting in a longer sequence than monolingual tokenizers. BPE tokenizer generates  tokens spanning multiple morphemes (\#\#현은, \#\#명한). \textit{Morpheme-based subword} tokenizer, on the other hand, better splits text into morphemes (\#\#은, \#\#의).}
    \begin{adjustbox}{width=1\textwidth}
    \begin{tabular}{ll}
    \toprule
    \textbf{Tokenization} & \textbf{Tokenized Sequence} \\
    \midrule
    Raw Text & 조경현은 인공지능 분야의 저명한 연구자이다. \\
    \midrule
    BPE (Multilingual) & 조 / \#\#경 / \#\#현 / \#\#은 / 인 / \#\#공 / \#\#지 / \#\#능 / 분 / \#\#야 / \#\#의 / 저 / \#\#명한 / 연구 / \#\#자 / \#\#이다 / . \\
    \midrule
    BPE & 조경 / \#\#현은 / 인공지능 / 분야의 / 저 / \#\#명한 / 연구 / \#\#자이 / \#\#다 / . \\
    Morpheme & 조경현 / 은 / 인공지능 / 분야 / 의 / 저명 / 한 / 연구자 / 이 / 다 / . \\
    \textbf{Morpheme-based Subword} & 조경 / \#\#현 / \#\#은 / 인공지능 / 분야 / \#\#의 / 저명 / \#\#한 / 연구자 / \#\#이다 / . \\
    \bottomrule
    \end{tabular}
    \end{adjustbox}
    \label{table:qualitative_tokenzation}
\end{table}

\paragraph{Training Configurations}

We choose BERT \cite{devlin2019bert} and RoBERTa \cite{liu2019roberta} 
architectures for our language models. Table~\ref{table:pretraining_details} describes the implementation details. All models take sequences of at most 512 tokens long each and are pretrained with a static or dynamic masking strategy following the original training procedure. When masking tokens, we use whole word masking (WWM) which masks all of the the tokens that form a single word. BERT also performs next sentence prediction (NSP). Other hyperparameters not specified in Table~\ref{table:pretraining_details} nor in the pretraining procedure details are same as the original configurations from \cite{devlin2019bert,liu2019roberta}. Due to the resource constraints, we could increase batch size only up to 2,048, unlike \citet{liu2019roberta} who use the batch size of 8k. We decrease the learning rate accordingly. We fix the learning rate to $10^{-4}$ for both BERT and RoBERTa. 

\begin{table}[t!]
    \centering
    \caption{Implementation details of \kluebert{} and \klueroberta{}. WWM refers to the whole word masking strategy.}
    \begin{adjustbox}{width=1\textwidth}
    \begin{tabular}{lcccccc}
       \toprule
        \textbf{Model} & \textbf{\# Parameter} & \textbf{Masking} & \textbf{Training Steps} & \textbf{Batch Size} & \textbf{Learning Rate} & \textbf{Device} \\
        \midrule
        \textbf{\bertbase{}} & 110M & Static, WWM & 1M & 256 & $10^{-4}$ & TPU v3-8 \\
        \midrule
        \textbf{\robertasmall{}} & 68M & Dynamic, WWM & 1M & 2048 & $10^{-4}$ & 8$\times$ V100 GPUs \\
        \textbf{\robertabase{}} & 110M & Dynamic, WWM & 1M & 2048 & $10^{-4}$ & 8$\times$ V100 GPUs \\
        \textbf{\robertalarge{}} & 337M & Dynamic, WWM & 500k & 2048 & $10^{-4}$ & 8$\times$ V100 GPUs \\
        \bottomrule
    \end{tabular}
    \end{adjustbox}
    \label{table:pretraining_details}
\end{table}

\subsection{Existing Language Models}

In addition to our own language models, 
we evaluate the following two existing multilingual language models and two Korean monolingual language models on our benchmark:

\begin{itemize}[leftmargin=*]
\item{\textbf{mBERT}} \cite{devlin2019bert} : A multilingual BERT introduced and released by \citet{devlin2019bert}. It is trained with the MLM and NSP objectives on a multilingual corpus covering 104 languages including Korean.

\item{\textbf{XLM-R}} \cite{conneau2019unsupervised} : A RoBERTa \cite{liu2019roberta} trained on a large multilingual corpus by using the MLM objective. 

\item{\textbf{KR-BERT}} \cite{lee2020kr} : An open-sourced character-level Korean language model based on BERT. 
We use the \texttt{KR-BERT character WordPiece} which uses a vocabulary of 16,424 unique tokens.

\item{\textbf{KoELECTRA}} \cite{park2020koelectra} : An open-source Korean language model trained with the MLM and replaced token detection objectives, as was done by \citet{clark2020electra}. For training corpora, \citet{park2020koelectra} uses own crawled news data and {\modu} corpus \cite{nikl2020corpora}.

\end{itemize}


\section{Fine-tuning Language Models}
\subsection{Task-Specific Architectures}

8 main tasks of KLUE benchmark can be categorized into 4 types based on the fine-tuning strategy. 
Topic classification (TC) and relation extraction (RE) are single sentence classification tasks. Sentence textual similarity (STS) and natural language inference (NLI) are sentence pair classification / regression tasks. Dialogue state tracking (DST) is a multiple-sentence slot-value prediction task. Finally, named entity recognition (NER), dependency parsing (DP) and machine reading comprehension (MRC) are sequence tagging tasks.

We use $x_i$ to refer to the $i$-th token of input and $h_i \in \mathbb{R}^H$ to its corresponding final hidden state from a pretrained language models (PLM), where $H$ is the hidden dimensionality. Following the conventional fine-tuning setup \cite{devlin2019bert}, we use \texttt{[CLS]} as the first input token $x_0$ and \texttt{[SEP]} as a delimiter token to separate inputs (e.g., two sentences in STS and NLI, a passage and a question in the case of MRC, and dialogue turns for DST.)

\subsubsection{Single Sentence Classification}

In the single sentence classification task, such as TC and RE, a classifier classifies a single sentence into a set of predefined labels. Following the convention, the last hidden state of \texttt{[CLS]} token $h_0$ is linearly mapped to the number of labels ($K$) with $W \in \mathbb{R}^{K \times H}$ and the entire model is trained to minimize the cross entropy loss. 

\textbf{\ynat} is a single sentence classification task where $K$ is 7 for predefined topic labels, and does not require any special treatment of each input. \textbf{\kluere} on the other hand requires a special procedure to indicate entities within the input sentence.
We use \texttt{<subj>}, \texttt{</subj>}, \texttt{<obj>}, and \texttt{</obj>} to mark the beginnings and the ends of subject and object entities, respectively, following \citet{baldini-soares-etal-2019-matching}. We expand the embedding matrix to add these four extra tokens.

\subsubsection{Sentence Pair Classification / Regression}

In the sentence pair classification / regression task, a model is asked to determine the relationship between two sentences. A pair of input sentences are concatenated with a special separator token, often \texttt{[SEP]}, in-between.
 
In \textbf{\kluests}, each sentence pair is annotated with a real-valued similarity $[0,5]$. The model is thus trained to map from the final hidden state of \texttt{[CLS]} to a real number, by minimizing the mean squared error (MSE). In the case of \textbf{\kluenli}, each sentence pair, consisting of a premise and hypothesis, is coupled with one of three classes. The model thus maps the hidden state of \texttt{[CLS]} token to a three-dimensional real-valued vectors and is trained to minimize the cross-entropy loss.

\subsubsection{Multiple-Sentence Slot-Value Prediction}

\begin{figure}[t] 
\centering
\label{fig:wos_model_arch}
\includegraphics[width=0.9\textwidth]{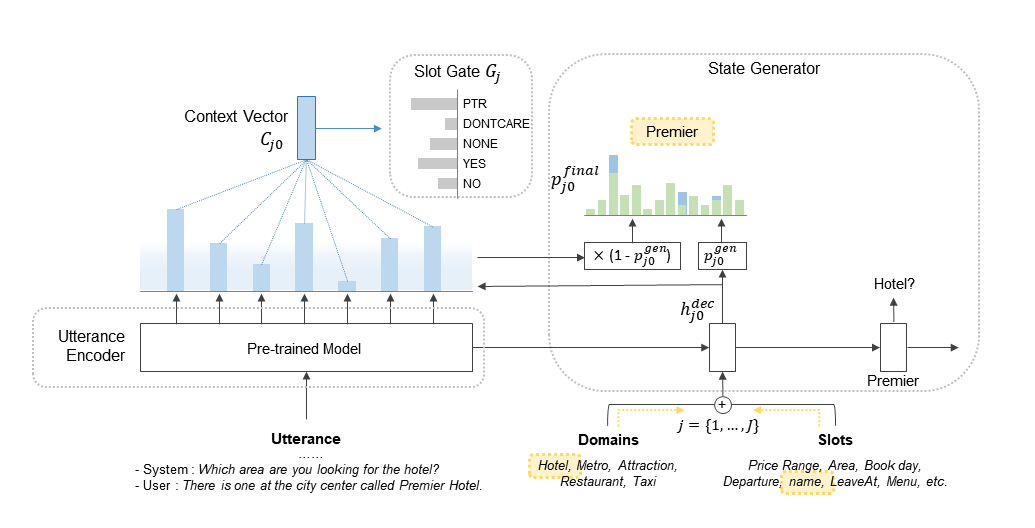}
\caption{A baseline architecture for \wizard{} based on TRADE \cite{wu2019transferable}.}
\end{figure}

\textbf{\wizard} is a slot-value prediction task for a given dialogue context, where the prediction should be considered across multiple turns instead of a single utterance. We employ an encoder-decoder model following the architecture of TRADE \cite{wu2019transferable}, which consists of an utterance encoder, a state generator, and a slot gate classifier (Figure~\ref{fig:wos_model_arch}). In our implementation, we change the utterance encoder from GRU \cite{cho-etal-2014-learning} to PLM to get better representations. Thus, the state generator takes the final hidden state of \texttt{[CLS]} token $h_0$ as the first decoder hidden state. We also modify the slot gate classifier to predict additional two slot gate labels (\textit{yes}, \textit{no}), since \wizard{} contains relatively more Boolean type slots than MultiWOZ \cite{budzianowski2018multiwoz}. We jointly minimize the cross-entropy loss of the state generator and slot gate classifier.

\subsubsection{Sequence Tagging}

\begin{figure}[t!] 
\centering
\includegraphics[width=0.95\textwidth]{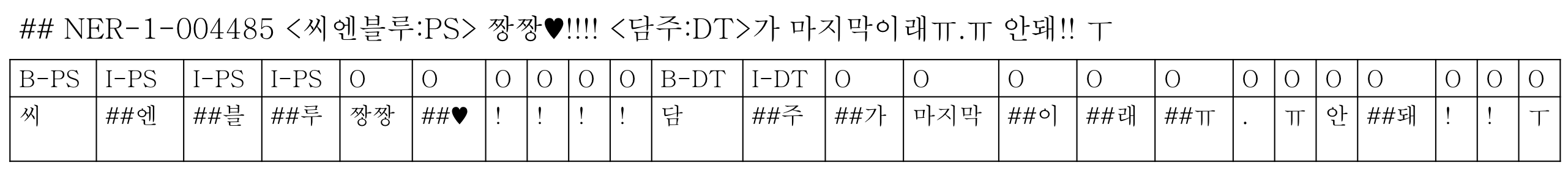}
\caption{An input and label example of \kluener{}. We realign original character-level label sequence of Figure~\ref{fig:ner_bio} for tokens from our Morpheme-based subword tokenization.}
\label{fig:ner_input_example}
\end{figure}

\textbf{\kluener} is a token-level tagging task, where each character is assigned a label. This requires a care in using tokenization, as the labels from the characters within each subword token must be aggregated, and the predicted label of each subword token must be properly distributed across the characters within it. See Figure~\ref{fig:ner_input_example} for an example. We linearly map each of the final hidden states from the encoder $h \in \mathbb{R}^{|x| \times H}$ into a 12-dimensional real-valued vectors, corresponding to the 12 named-entity categories. We then minimize the cross-entropy loss summed over all the tokens.

\textbf{\kluemrc} is a span prediction task in which a model tags the beginning and end tokens of the answer span within a passage, given a question. The input to the model is the concatenation of a tokenized passage and an associated question (separated by \texttt{[SEP]}). The final hidden state of each token in the passage is linearly projected to a 2-dimensional real-valued vector. The dimensions in this vector correspond to the logits of two binary classifiers, the start and end token classifiers. The special token \texttt{[CLS]} is considered as both the correct start and end tokens, when given question is unanswerable. We minimize the cross-entropy loss to train the model.

\begin{figure}[t!]
\centering
\includegraphics[scale=0.25]{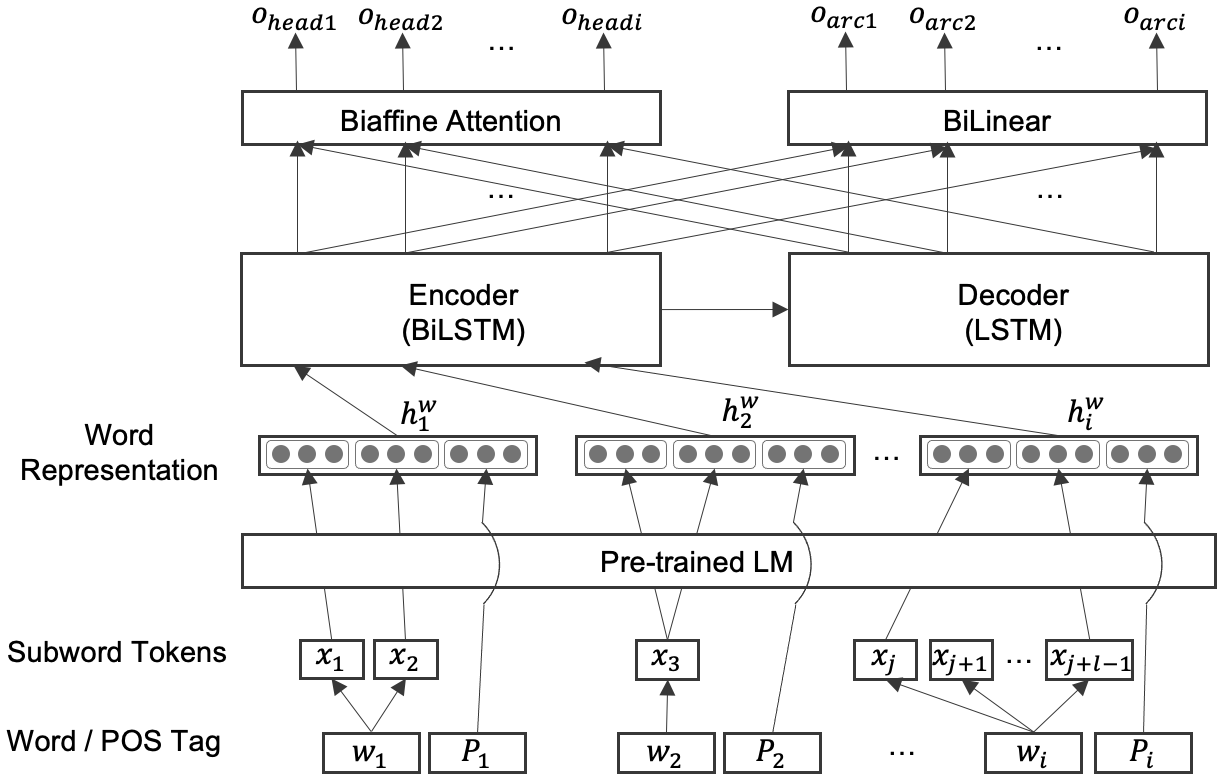}
\caption{An overview of {\klueposdp} baseline model architecture, which we take advantage of \citet{fernandez-gonzalez-gomez-rodriguez-2019-left}.}
\label{fig:dp_model}
\end{figure}

We frame \textbf{\klueposdp{}} as a sequence tagging problem. Each token within an input sentence is tagged twice, once with its head token and the other with the type of the arc connecting the head and the current token. Our baseline architecture follows the model proposed by \citet{fernandez-gonzalez-gomez-rodriguez-2019-left}, except word representation and attention mechanisms. Similarly to \kluener{}, we must be careful in handling subword tokens, as the annotation is done at the word level. In our approach, we use a pretrained language model (to be fine-tuned) to extract subword representations and concatenate the first and last subword token representations of each word, to form word vector representations. Each of these word representations is optionally concatenated with the part-of-speech embedding. For the attention layers, we use biaffine attention \cite{timo2017biattention} to predict HEAD, and bilinear attention \cite{kiperwasser-goldberg-2016-simple} to predict arc type (DEPREL) for each word. Just like \kluener{} and \kluemrc{}, we minimize the cross-entropy loss to fine-tune the entire model. See Figure~\ref{fig:dp_model} for a graphical illustration of the model architecture.

\subsection{Fine-Tuning Configurations} 

For all the experiments, we use \texttt{Huggingface Transformers} \cite{wolf-etal-2020-transformers} and \texttt{PyTorch-Lightning}.\footnote{
\url{https://github.com/PyTorchLightning/pytorch-lightning}
}  
We use AdamW optimizer \cite{loshchilov2018decoupled} with the learning rate selected from $\{10^{-5}, 2 \times 10^{-5}, 3 \times 10^{-5}, 5 \times 10^{-5}\}$, the warm-up ratio from $\{0., 0.1, 0.2, 0.6 \}$ and the weight decay coefficient from $\{0.0, 0.01\}$. We choose the batch size from $\{8, 16, 32\}$ and the number of epochs from $\{3, 4, 5, 10\}$. We use the maximum sequence length of 512 for {\kluemrc} and {\wizard}, and 128 for all the other tasks. We report the score obtained from the best hyperparameter configuration based on the dev set performance.

\subsection{Evaluation Results}
\begin{table}[t!]
\centering
\caption{Evaluation results of our pretrained LMs and other baselines on KLUE benchmark. The F1 refers to a macro-F1 score. The F1$^E$ and F1$^C$ of {\kluener} indicates entity-level and character-level macro-F1 score, respectively. The F1$^{mic}$ of {\kluere} is micro-averaged F1 score ignoring the \textit{no\_relation}. The F1$^S$ of {\wizard} is an average of slot-value pair level micro-F1 scores. The R$^P$ of {\kluests} denotes Pearson correlation. \textbf{Bold} shows the best performance across the models, and \underline{underline} indicates the best performance among \texttt{BASE} models.}
\begin{adjustbox}{width=1\textwidth}
\begin{tabular}{l c cc c cc cc cc cc cc}
\toprule
& \textbf{\ynat{}} & \multicolumn{2}{c}{\textbf{\kluests{}}} & \textbf{\kluenli{}} & \multicolumn{2}{c}{\textbf{\kluener{}}} & \multicolumn{2}{c}{\textbf{\kluere{}}} & \multicolumn{2}{c}{\textbf{\klueposdp{}}} & \multicolumn{2}{c}{\textbf{\kluemrc{}}} & \multicolumn{2}{c}{\textbf{\wizard{}}} \\ \cmidrule(lr){2-2} \cmidrule(lr){3-4} \cmidrule(lr){5-5} \cmidrule(lr){6-7} \cmidrule(lr){8-9} \cmidrule(lr){10-11} \cmidrule(lr){12-13} \cmidrule(lr){14-15}
        \textbf{Model} & F1 & R$^P$ & F1 & ACC & F1$^E$ & F1$^C$ & F1$^{mic}$ & AUC & UAS & LAS & EM & ROUGE & JGA & F1$^S$ \\ \midrule
        \textbf{\mbertbase{}} & 81.55 & 84.66 & 76.00 & 73.20 & 76.50 & 89.23 & 57.88 & 53.82 & 90.30 & 86.66 & 44.66 & 55.92 & 35.46 & 88.63 \\
        \textbf{\xlmrbase{}} & 83.52 & 89.16 & 82.01 & 77.33 & 80.37 & 92.12 & 57.46 & 54.98 & 89.20 & 87.69 & 27.48 & 53.93 & 39.82 & 89.61 \\
        \textbf{\xlmrlarge{}} & \textbf{86.06} & 92.97 & 85.86 & 85.93 & 82.27 & \textbf{93.22} & 58.39 & 61.15 & 92.71 & \textbf{88.70} & 35.99 & 66.77 & 41.20 & 89.80 \\
        \midrule
        \textbf{\krbertbase{}} & 84.58 & 88.61 & 81.07 & 77.17 & 74.58 & 90.13 & 62.74 & 60.94 & 89.92 & 87.48 & 48.28 & 58.54 & 45.33 & 90.70 \\
        \textbf{\koelectrabase{}} & 84.59 & \underline{92.46} & \underline{84.84} & \underline{85.63} & \underline{\textbf{86.11}} & \underline{92.56} & 62.85 & 58.94 & 92.90 & 87.77 & 59.82 & 66.05 & 41.58 & 89.60 \\
        \midrule
        \textbf{\bertbase{}} & \underline{85.73} & 90.85 & 82.84 & 81.63 & 83.97 & 91.39 & 66.44 & 66.17 & 89.96 & 88.05 & 62.32 & 68.51 & 46.64 & 91.61 \\
        \textbf{\robertasmall{}} & 84.98 & 91.54 & 85.16 & 79.33 & 83.65 & 91.14 & 60.89 & 58.96 & 90.04 & 88.14 & 57.32 & 62.70 & 46.62 & 91.44 \\
        \textbf{\robertabase{}} & 85.07 & 92.50 & 85.40 & 84.83 & 84.60 & 91.44 & \underline{67.65} & \underline{68.55} & \underline{93.04} & \underline{88.32} & \underline{68.67} & \underline{73.98} & \underline{47.49} & \underline{91.64} \\ 
        \textbf{\robertalarge{}} & 85.69 & \textbf{93.35} & \textbf{86.63} & \textbf{89.17} & 85.00 & 91.86 & \textbf{71.13} & \textbf{72.98} & \textbf{93.48} & 88.36 & \textbf{75.58} & \textbf{80.59} & \textbf{50.22} & \textbf{92.23} \\
        \bottomrule
\end{tabular}
\end{adjustbox}
\label{table:main_results}
\end{table}

In this section, we present the evaluation results including our KLUE-PLMs and existing PLMs on the KLUE benchmark, in Table~\ref{table:main_results}.\footnote{
See Appendix \ref{appendix:dev_set_results} for the corresponding table however computed on the development set.
} 
Different from other NLU benchmarks, we do not average the scores over tasks, since simple averaging of scores of different scales and interpretations could be highly misleading. 
Rather, we describe and discuss the result of each task separately. Within the Korean \texttt{BASE} models, \bertbase{} performs best for \ynat{} and \wizard{}, \robertabase{} for \kluere{} and \kluemrc{}, and \koelectrabase{} for \kluests{} and \kluenli{}.

We make two major observations. 
First, we see that \robertalarge{}, which is the largest model among the baseline PLM's we tested, outperforms all the other models across all the tasks except for \kluener{}. This observation agrees well with the recent trend which has demonstrated the correlation between the model size and task performance \cite{kaplan2020scaling, NEURIPS2020_1457c0d6}. This indicates that KLUE will be useful for the future investigation into how much gain we can expect by simply increasing the model size further. 
The second observation is that the monolingual models, which were specifically designed for and trained with a more carefully curated corpus in the target language (Korean), generally outperform the multilingual counterparts, especially when we compare models of similar sizes. We make this observation again across all the tasks, except for \kluener{}, where the \xlmrlarge{} performs similarly to the best performer, \koelectrabase{}, in terms of character-level F1 score. This observation re-iterates the importance of investing effort in understanding a target language and customizing data, models and learning algorithms for the target language.

\subsection{Analysis of Models}
There were two major decisions we made in preparing the pretraining corpus and preprocessing data. They were 1) whether to pseudonymize PIIs and 2) the tokenzation strategy. In this section, we analyze the impact of our choices, using  \robertabase{} by training it on the \modu{} corpus only.

\paragraph{Corpus Pseudonymization}

It can be expected that noise introduced in the process of pseudonymization may have detrimental effect on the downstream task performance. Our finding, presented in Table~\ref{table:pii_ablation}, however shows that there is some drop in a subset of the tasks, but such drop is quite minimal. This suggests that the minimal level of pseudonymization, just like what we have done, is already a good way to balance the task performance and the risk of leaking private information.

\begin{table}[t!]
\centering
\caption{Evaluation results according to whether the corpus pseudonymization is conducted in the preprocessing step.}
\begin{adjustbox}{width=1\textwidth}
\begin{tabular}{l c cc c cc cc cc cc cc}
\toprule
& \textbf{\ynat{}} & \multicolumn{2}{c}{\textbf{\kluests{}}} & \textbf{\kluenli{}} & \multicolumn{2}{c}{\textbf{\kluener{}}} & \multicolumn{2}{c}{\textbf{\kluere{}}} & \multicolumn{2}{c}{\textbf{\klueposdp{}}} & \multicolumn{2}{c}{\textbf{\kluemrc{}}} & \multicolumn{2}{c}{\textbf{\wizard{}}} \\ \cmidrule(lr){2-2} \cmidrule(lr){3-4} \cmidrule(lr){5-5} \cmidrule(lr){6-7} \cmidrule(lr){8-9} \cmidrule(lr){10-11} \cmidrule(lr){12-13} \cmidrule(lr){14-15}
        \textbf{Pretraining Corpus} & F1 & R$^P$ & F1 & ACC & F1$^E$ & F1$^C$ & F1$^{mic}$ & AUC & UAS & LAS & EM & ROUGE & JGA & F1$^S$ \\ \midrule
        Original & \textbf{83.40} & \textbf{92.06} & \textbf{84.70} & \textbf{81.60} & 84.84 & 91.03 & \textbf{65.25} & \textbf{64.79} & \textbf{92.17} & \textbf{88.34} & 62.13 & 67.46 & \textbf{47.14} & \textbf{91.60} \\ 
        Pseudonymized & 83.39 & 91.11 & 82.85 & 78.50 & \textbf{84.99} & \textbf{91.22} & 62.79 & 62.96 & 92.02 & 88.02 & \textbf{62.88} & \textbf{67.58} & 46.21 & 91.23 \\ \bottomrule
\end{tabular}
\end{adjustbox}
\label{table:pii_ablation}
\end{table}

\begin{table}[t!]
\centering
\caption{Comparison our tokenization strategy with other baselines.}
\begin{adjustbox}{width=1\textwidth}
\begin{tabular}{l c cc c cc cc cc cc cc}
\toprule
& \textbf{\ynat{}} & \multicolumn{2}{c}{\textbf{\kluests{}}} & \textbf{\kluenli{}} & \multicolumn{2}{c}{\textbf{\kluener{}}} & \multicolumn{2}{c}{\textbf{\kluere{}}} & \multicolumn{2}{c}{\textbf{\klueposdp{}}} & \multicolumn{2}{c}{\textbf{\kluemrc{}}} & \multicolumn{2}{c}{\textbf{\wizard{}}} \\ \cmidrule(lr){2-2} \cmidrule(lr){3-4} \cmidrule(lr){5-5} \cmidrule(lr){6-7} \cmidrule(lr){8-9} \cmidrule(lr){10-11} \cmidrule(lr){12-13} \cmidrule(lr){14-15}
        \textbf{Tokenization} & F1 & R$^P$ & F1 & ACC & F1$^E$ & F1$^C$ & F1$^{mic}$ & AUC & UAS & LAS & EM & ROUGE & JGA & F1$^S$ \\ \midrule
        BPE & \textbf{83.40} & 91.91 & \textbf{85.19} & \textbf{82.07} & 68.75 & 89.47 & 64.39 & \textbf{65.04} & 89.89 & \textbf{89.47} & 51.12 & 65.79 & 21.38 & 77.68 \\
        Morpheme-based Subword & \textbf{83.40} & \textbf{92.06} & 84.70 & 81.60 & \textbf{84.84} & \textbf{91.03} & \textbf{65.25} & 64.79 & \textbf{92.17} & 88.34 & \textbf{62.13} & \textbf{67.46} & \textbf{47.14} & \textbf{91.60} \\ \bottomrule
\end{tabular}
\end{adjustbox}
\label{table:tokenization_strategy}
\end{table}

\paragraph{Tokenization Strategy}

We contrast our tokenization scheme, \textit{morpheme-based subword} tokenization, against the standard byte pair encoding (BPE). First, we investigate the difference in how words are segmented into subword tokens. Following \citet{rust2020good}, we consider subword fertility, proportion of continued words, and \texttt{UNK} ratio. On the subword fertility, which measures the average number of subwords produced per word, the proposed tokenization scheme ends up slightly higher than BPE does. However, when we look at the proportion of continued words, which measures the number of words that were split into at least two subwords, we observe the opposite trend. This implies that our algorithm maintains the original words as much as it can, and only when it is necessary, it splits each word into potentially more subword pieces. The efficacy of the proposed scheme over BPE is evident from the \texttt{UNK} ratio, as it produces fewer \texttt{UNK} tokens compared to BPE when the vocabulary size was controlled to be 32k for both methods. See Table~\ref{table:tokenization_metrics}.

\begin{table}[!ht]
    \centering
    \caption{Overview of tokenization metrics. We build each vocabs using \modu{} corpus and compare them on \wikipedia{} corpus.}
    \begin{adjustbox}{width=0.65\textwidth}
    \begin{tabular*}{0.85\columnwidth}{lcccc}
        \toprule
        \textbf{Tokenization} & \textbf{\# Vocabs} & \textbf{Fertility $\downarrow$} & \textbf{\% Continued Word $\uparrow$} & \textbf{UNK Ratio $\downarrow$} \\
        \midrule
        BPE & 32k & \textbf{2.073} & 0.578 & 0.011 \\
        Morpheme-based Subword & 32k & 2.468 & \textbf{0.765} & \textbf{0.009} \\
        \bottomrule
    \end{tabular*}
    \end{adjustbox}
    \label{table:tokenization_metrics}
\end{table}

We find these qualitative differences between two schemes lead to significant differences in the task performance in the cases of \kluener{}, \kluemrc{} and \wizard{}. These tasks often involve tagging, detection and even generation at the morpheme level, and we suspect that morphologically consistent tokenization facilitates better prediction overall. On the other hand, the difference in the tokenization strategy does not manifest itself in the performance of classification or word-level tagging, likely as a corresponding NLU system can more readily overcome inconsistencies in subword segmentation when merging subword token representations into that of a larger unit. Overall, we recommend future researchers use the proposed tokenization strategy as a default option.


\clearpage
\section{Ethical Considerations}
In building KLUE and accompanying baseline models, we have incorporated various mechanisms to avoid any harmful and negative consequences from releasing both data and models. These mechanisms are described in detail wherever they were introduced and used, but in this section, we summarize these mechanisms, considerations and our principles behind them.

\subsection{Copyright and Accessibility}

Most NLP datasets are built upon the existing text sources. This raises a question on the terms of using such datasets, especially when the underlying source datasets are not well-specified nor carefully investigated. In order to avoid any such doubt on the terms of using KLUE and to accelerate NLP research in Korean, we fully adhere to the copyright act of Korea, which went effective on Dec. 8, 2020.\footnote{
\url{https://www.law.go.kr/\%EB\%B2\%95\%EB\%A0\%B9/\%EC\%A0\%80\%EC\%9E\%91\%EA\%B6\%8C\%EB\%B2\%95}
}
and include only text for which we know we can release under a license that permits both redistribution and re-mix without any restriction on the use.

\paragraph{Source Corpora}

Our goal is to secure and maximize the continued availability and usefulness of the benchmark. In other words, we must guarantee the possibility of derive new work and redistribute it freely, which comes together with CC BY-SA. To release KLUE under CC BY-SA, we have built a source corpus set by including only text that is either 1) not protected by copyright or 2) under CC0, CC BY, CC BY-SA or KOGL Type 1 license. In the case of news articles, which are copyrighted, we have signed contracts with the providers, Korean Economics Daily (KED) news media and Acrofan, that allow us to make \kluemrc{} and release them under CC BY-SA.

\paragraph{Task-Specific (Annotated) Datasets}

We subsample and annotate the source corpus for each KLUE benchmark task. We release each under CC BY-SA. This allows users of KLUE benchamrk to copy, redistribute, remix, transform and build upon it for both commercial and non-commercial purposes, as long as derivatives are distributed under the same license (CC BY-SA). We expect this to greatly facilitate future NLP research and development.

\paragraph{Pretraining Corpora and Language Models} 

As was discussed earlier~\ref{sec:pretraining-corpora}, we cannot guarantee that our pretraining corpus, built using \modu{}, \CommonCrawl{} and \newscrawl{}, does not contain any copyrighted work, although these are all created from publicly available text. 
Unfortunately without these corpora, it is not possible to find a sufficiently large resource to train large-scale language models for Korean. We thus use them for pretraining but do not publicly release the pretraining corpora in order to avoid any issues in the future, which is in contrast to KLUE. Instead, we openly release pretrained language models to facilitate future research. As the parameters of a language model does not \textit{[express] human thoughts and emotions}, they do not meet the requirement of being copyrighted.

\subsection{Toxic Content}

Although large-scale, accessible benchmark datasets advance machine learning and its applications to adjacent fields, such as natural language processing, toxic and unwanted contents within these datasets may be amplified via large-scale models we train on them. We have been aware of this issue from the beginning of the project, and here we describe how we have addressed these toxic contents in KLUE.

\paragraph{Task-Specific Datasets}

For each task-specific dataset, we apply three stages to minimize the introduction of toxic contents. First, we automatically detect hate speech and gender-biased sentences using toxicity classifiers and remove those even before sending these sentences for annotation (see Section~\ref{sec:corpus-preprocessing}). 
Second, we explicitly and clearly instruct annotators to mark any instance that exhibits social biases and/or is toxic (see Section~\ref{tasks}), after providing them with clear definitions of bias and hate speech. Finally, we manually examine these marked sentences and exclude them from the final dataset. 
This three-stage process may not catch all possible such instances, and we plan to use an online forum\footnote{\url{https://github.com/KLUE-benchmark/KLUE/issues}} to receive feedback and complaints from users of KLUE.

\paragraph{Pretrained Language Models}

We use our pretraining corpora (see Section~\ref{sec:pretraining-corpora}) as they are, for three reasons. First, manual inspection is simply not tractable due to the sheer scale. Second, it is challenging to build an automated tool to detect hate speech and biased sentences. This issue is made even more severe for Korean, because there is only one known hate speech dataset of limited size \cite{moon-etal-2020-beep}. 
Lastly, we envision the future in which these pretrained language models are used to build better tools for automatically detecting various toxic contents as well as undesirable social biases. In order for such pretrained models to be aware of these issues, they must have been trained with such toxic contents as well.

\subsection{Personally Identifiable Information}

It has recently been discovered by that a large-scale, pretrained language model memorizes a large amount of personally identifiable information (PII) and that an algorithm can be designed to retrieved those private information. We thus design two different approaches for pseudonymizing task-specific datasets and pretraining corpora, respectively.

\paragraph{Task-Specific Datasets}

In the case of task-specific datasets, we rely on manual inspection during annotation to detect PII. We discard any sentences that was reported to contain PII after manual inspection. In the case of \dst{}, which relies on simulated dialogues, we pseudonymize the database entries rather than actual text, using the \texttt{faker} library.\footnote{
\url{https://github.com/joke2k/faker}
}

\paragraph{Pretraining Corpora}

There is a trade-off between removing PII and the performance of a pretrained language model, as we will demonstrate later in this paper. We thus pseudonymize 16 PII types that are detectable purely by regular expressions. See Section~\ref{sec:plm-corpora-prep}

\section{Related Work}
\paragraph{General-Purpose NLU Benchmarks}

General Language Understanding Evaluation (GLUE) \cite{wang2018glue} benchmark, a collection of evaluation dataset for English, was the first general-purpose evaluation benchmarks for NLU. It is general-purpose in that it is not limited to a single task. It consists of 11 downstream tasks, including tasks that measures the capability of capturing semantic textual similarity (QQP, MRPC, STS) \cite{agirre-etal-2016-semeval,cer-etal-2017-semeval}, measures the capability of inference (MNLI, QNLI, RTE, WNLI) \cite{bowman2015large,williams2017broad} and that evaluates the capability of classifying a single sentence into a predefined set of categories (CoLA, SST) \cite{warstadt-etal-2019-neural,socher-etal-2013-recursive}. 
GLUE exclusively focused on English, and its variants in different languages have been built and released over the past couple of years, including CLUE in Chinese~\cite{xu2020clue}, a French version~\cite{le2020flaubert}, an Indonesian version~\cite{wilie2020indonlu}, a version for Indic languages~\cite{kakwani2020indicnlpsuite}, Russian SuperGLUE~\cite{shavrina2020russiansuperglue}, and Persian GLUE~\cite{2020parsiglue}. In all these cases, substantial efforts were carried out to follow the philosophy of the original GLUE, covering a broad spectrum of domains and tasks, while incorporating language-specific characteristics. 
On the other hand, there have been efforts to build a multilingual version of such benchmark, largely relying on automated methods, such as  XGLUE \cite{liang2020xglue} and XTREME \cite{hu2020xtreme}. Korean as a language has been included in subsets of these latter benchmarks, but there has not been a serious attempt at building a general-purpose language understanding evaluation suite for Korean, until this paper.

\paragraph{Absence of a Standard NLU Benchmark in Korean}

Until this paper, a number of task-specific benchmarks in Korean have been proposed and released. For example, \nsmc{} is used for sentiment classification, PAWS-X \cite{yang2019paws} for paraphrase detection, KorNLI and KorSTS \cite{ham-etal-2020-kornli} for NLI and STS, KorQuAD 1 and 2 \cite{lim2019korquad1,youngmin2020korquad2} for MRC, and BEEP! \cite{moon-etal-2020-beep} for hate speech detection. At this point, one may wonder whether it would have been easier and more convenient to simply aggregate these datasets to build KLUE. After all, this has been a popular strategy for constructing monolingual \cite{wang2018glue} as well as multilingual \cite{liang2020xglue,hu2020xtreme} benchmarks. Unfortunately this approach comes with two major issues that we directly address in this paper.

First, the existing datasets are constructed individually without considering other datasets and their properties. In other words, the aggregate of these individual datasets is unlikely to cover a broad spectrum of domains and writing styles, unlike KLUE for which we carefully curate the source corpora as well as subsets for downstream tasks to have broad coverage of domains and styles.
This goes beyond domains and styles, but also the coverage of linguistic phenomena under evaluation. Most of the existing benchmarks, listed above, focus on semantics rather than syntax, and it is difficult to find any widely-available benchmark that captures pragmatics. We address this issue by carefully selecting a set of downstream tasks.

Second, these existing datasets are not always publicly available,\footnote{
Some of these are publicly available in Korea but not internationally.
} 
and some are distributed with a highly restrictive license that prohibits redistribution nor the transformation of the original. These are often the ones published and released by government-affiliated institutes. In some cases, it is necessary to obtain a special permission to access datasets, which is often not easily accessible by non-Korean researchers. We address all these issues with KLUE by releaseing the entire benchmark data under CC BY-SA, both by careful curation of source corpora and by direct agreements with publishers.

\paragraph{Pretrained Language Models (PLMs)}

The recent trend of large-scale pretrained language models was sparked by the success of earlier models, such as ELMo \cite{peters-etal-2018-deep}, GPT-2 \cite{radford2019language} and BERT \cite{devlin2019bert}, on the GLUE and other similar NLU benchmarks. This earlier success has led to a series of advances in large-scale language models, including XLNet \cite{NEURIPS2019_dc6a7e65}, ALBERT \cite{lan2020ALBERT}, RoBERTa \cite{liu2019roberta}, ELECTRA \cite{clark2020electra}, and Deberta \cite{he2020deberta}, again largely driven by the availability of standardized benchmarks. 
This advance in language models, not only in terms of the model size but also in learning algorithms, in turn also sparked the interest in building and improving existing language understanding benchmarks. Some of the recently released, challenging benchmarks include SuperGLUE \cite{wang2019superglue} and KILT \cite{petroni2020kilt}. The availability of such a standard language understanding benchmark, such as KLUE from this paper, is expected to start such a virtuous cycle for Korean language understanding.

\paragraph{Pretrained Language Models for Korean}

Inspired by the development in other languages and multilingual models, PLM's for the Korean language have been trained and released by multiple research groups and individuals. SKT released KoBERT,\footnote{
\url{https://github.com/SKTBrain/KoBERT}
} 
followed by KorBERT\footnote{
\url{https://aiopen.etri.re.kr/service_dataset.php}
} 
from ETRI, 
HanBERT\footnote{
\url{https://github.com/tbai2019/HanBert-54k-N}
} 
from TwoBlock AI, KR-BERT \cite{lee2020kr} from Seoul National University. There are a few pretrained models released by individual researchers, such as KoELECTRA \cite{park2020koelectra} and KcBERT \cite{lee2020kcbert}. 

Unfortunately, it is unclear how we should compare this stream of pretrained language models in Korean, due to the lack of a standarded benchmark in Korean. Subsets of these models have been compared based on subsets of a few downtream NLP tasks in Korean above, but because these are not standardized, it is not easy to draw solid conclusions from these limited experiments. We expect the proposed KLUE benchmark will serve as a standard way to track the progress of research in language models for Korean.


\section{Discussion}
\paragraph{Open Access}

We distribute KLUE under CC BY-SA. The license allows everyone to freely copy and redistribute our benchmarks in any medium or format. In addition, one can improve our benchmark to build more challenging datasets after performance saturation. To function as a NLU \textit{benchmark}, open access is a must. If the original author does not allow derivative development of the benchmark, other researchers cannot improve it, for example by removing toxic content, or building a more challenging dataset to accelerate research for technical improvements. If commercial use is not allowed, researchers working at for-profit organizations would not be able to benefit from nor to (easily) contribute to the benchmark. Redistribution is another crucial factor because it significantly limits research if, for example, sharing the datasets with another researcher is prohibited. Another existing practice that limits research is transferring the responsibility of copyright infringement of related conflicts to researchers. To set a good precedent for open access of data, we allow using our datasets for 1) any purpose, 2) derivative work, and 3) redistribution, as long as the existing copyrights in our benchmark datasets are respected. We also open our pretrained Korean language models and the implementation of pretraining and fine-tuning pipelines. This enhances reproducibility of our work, and allows anyone to fix and improve our data and models. We hope to contribute to the Korean NLP research community as well the wider NLP community.

\paragraph{Facilitating Korean NLP Research} 

We developed KLUE with the aim of facilitating Korean NLP research, in response to the recent active development efforts of large Korean language models. The entire NLP community has seen BERT \cite{devlin2019bert} and its variants outperforming the previous NLU models for GLUE \cite{wang2018glue} and SuperGLUE \cite{wang2019superglue}, as well as the more recent GPT3 \cite{NEURIPS2020_1457c0d6} with outstanding performance without fine-tuning (and with \textit{in-context learning}) in natural language understanding and generation. Motivated by these models, many Korean researchers at various institutions rushed to pretrain large-scale Transformer-based Korean language models. Consequently, a number of nearly identical pretrained language models have been released to open-source communities. However, we could not systematically understand the behaviors and characteristics of these models because of the lack of well-designed general-purpose benchmarks like GLUE for Korean. KLUE will allow us to conduct controlled experiments to understand how and why various Korean LMs perform on certain tasks and thus obtain detailed insights into those models. Furthermore, since KLUE includes many representative NLU tasks that are also conducted in other languages, KLUE will function as a fundamental resource to NLP researchers who aim to conduct multilingual research with Korean and other languages.

\paragraph{Measuring Overall Performance of NLU models} 
We do not average all scores gained from each task in KLUE. The performance of all tasks are measured by different evaluation metrics. This is because we carefully choose the metric for each task with considering its own characteristics. Their granularity differs by tasks, for example, \kluemrc{} and \kluener{} employ character-level metrics because an entity can exist within a word in Korean whereas \kluests{} and \kluenli{} use sentence-level metrics. Furthermore, we use various metrics across tasks, such as F1 score, accuracy, area under the curve, UAS, LAS, ROUGE-W, joint goal accuracy, and Pearson's correlation. In this situation, simply computing the average of all tasks as in GLUE \cite{wang2018glue} results in misleading overall performance measure. The average will lose its interpretability as well as giving higher weights to a certain task in unintended ways. Accordingly, an alternative way to estimating a model's NLU capability is necessary. Recently, analyzing correctness of a model's prediction by using Item Response Theory (IRT) framework to estimate such capability is proposed \cite{lalor-yu-2020-dynamic}, however, we find that it is not clear how it should be applied precisely in our benchmark. As of now, we thus decide to evaluate a model for each task separately without any summarization of overall performance measure. This is our limitation, and we leave this problem for the future.

\paragraph{Rapid Saturation of KLUE} 

We expect fast saturation of KLUE based on our observations of for instance GLUE \cite{nangia-bowman-2019-human}. However, we do not artificially make our benchmark challenging by e.g. filtering out easy examples for models. Because the main purpose of KLUE is to properly evaluate models in terms of various aspects of NLU, we avoid focusing on enlarging headrooms for improvement over our baseline pretrained language models. We expect our license policy would positively affect the advancement of our benchmark after saturation by collectively developing more challenging tasks with other researchers, such as building the first open-domain question answering for Korean.

\paragraph{Analysis of Korean Language Models} 

We observe various patterns when comparing performances of the baseline models on each tasks, however, most of them are understudied to precisely explain the phenomena. With more thorough investigations, we hope to enhance understanding of the complex interaction of a model, corpus, linguistic properties of Korean and training mechanisms in future work.


\section{Conclusion}
We present KLUE, a suite of Korean NLU benchmarks that includes diverse tasks. We open KLUE to everyone, and we also provide Korean language models trained to outperform multilingual models and other existing open-sourced Korean language models. We set high standards from the outset, as we built the benchmark and trained the models from scratch. We designed the benchmark datasets and trained the annotators rigorously to consider potential ethical issues including private information and hate speech. We documented in detail all of the benchmark construction and testing processes. We also discussed broader impacts and limitations of KLUE and our models. Despite the limitations, KLUE and the accompanying language models will facilitate future Korean NLP research by setting a valuable precedent describing how datasets and language models should be created and spread to a wider community. 

\begin{ack}
Upstage sponsored annotation cost and built the leaderboard. NAVER CLOVA provided data annotation cost and GPU cloud computing infrastructure (NSML). We also thank to Google's TensorFlow Research Cloud (TFRC) and Kakao Enterprise's BrainCloud. The three computing resources were used to pretrain and fine-tune language models. Scatter Lab, SelectStar, Riiid!, DeepNatural and KAIST sponsored data annotation cost. In addition, we appreciate The Korea Economy Daily and Acrofan for supporting their news articles for MRC datasets.

The authors thank Cheoneum Park for discussions about task selection and DP task, Jinhyuk Lee and Minjoon Seo for discussions on MRC task, Sujeong Kim and DongYeon Kim for considerable efforts to manage annotators in MRC dataset, and Sangah Park for careful consideration of data construction in DP, NER, and RE. We appreciate Junyeop Lee, Geonhee Lee, Jiho Lee, Daehyun Nam, and Yongjin Cho for the leaderboard and evaluation system implementation.

This study is reviewed and approved by the KAIST Institutional Review Board (\#KH2020-173).
\end{ack}

\medskip

\clearpage  
\bibliographystyle{plainnat}
\bibliography{klue}

\small

\printindex
\clearpage
\section*{Contribution}

\textbf{Sungjoon Park} led the project as project manager, initiated the project, made decisions on overall progress of this project, secured financial resources, signed up with \acrofan{} for the articles, and organized IRB submission and research paper. 

\textbf{Jihyung Moon} led the project as project manager, managed overall datasets, models, and ethical concerns, signed up with \acrofan{} for the articles, prepared IRB, as well as contributed to \ner{}, \sts{}, \nli{}, \mrc{} dataset constructions, \airbnb{}, \policy{} corpora collection, and leaderboard design.  
 
\textbf{Sungdong Kim} managed overall fine-tuning of language models, served as a person in charge (PIC) of \dst{}, and contributed to the dataset construction of \tc{}, \sts{}, and \re{}.

\textbf{Won Ik Cho} managed the overall dataset construction of \tc{}, \sts{}, \nli{}, \re{}, \mrc{}, and \dst{}, provided the original corpus of \parakqc{}, and served as a PIC of \sts{}.

\textbf{Jiyoon Han} managed the overall dataset construction of \posdp{} and \ner{}, served as a PIC of \nli{}, contributed to the dataset construction of \sts{}, and took part in preparing IRB.

\textbf{Jangwon Park} served as a PIC of model pretraining, contributed to the text collection of \yna{}, collected and pre-processed \modu{}, \CommonCrawl{}, \namuwiki{}, \newscrawl{}, and \petition{}, and conducted the fine-tuning of \tc{}.

\textbf{Chisung Song} served as a PIC of \ner{} and contributed to the dataset construction of \posdp{} and \dst{}.

\textbf{Junseong Kim} served as a PIC of \mrc{}, conducted the fine-tuning of \mrc{}, and contributed to the text collection of \wikitree{} and \wikipedia{}.

\textbf{Youngsook Song} served as a PIC of \tc{} and contributed to the dataset construction of \ner{} and \dst{}.

\textbf{Taehwan Oh} served as a PIC of \posdp{}, contributed to the dataset construction of \ner{} and \nli{}, and took part in preparing IRB.

\textbf{Joohong Lee} served as a PIC of \re{}, and conducted the fine-tuning of \re{}.

\textbf{Juhyun Oh} contributed to the dataset construction of \nli{}, \ner{}, \posdp{} and \sts{}, took part in ethical considerations, IRB preparation and setup for model pretraining.

\textbf{Sungwon Lyu} contributed to the dataset construction of \sts{}, \nli{}, \re{}, and \mrc{}, taking part in the overall model pretraining and task-wise fine-tuning.

\textbf{Younghoon Jeong} contributed to the text collection of \wikinews{}, modeling of \posdp{}, and pre-processing of the pretraining corpus.

\textbf{Inkwon Lee} contributed to the text collection, modeling of \posdp{} and pre-processing of the pretraining corpus.

\textbf{Sangwoo Seo} contributed to the dataset construction of \re{}, and took part in preparing IRB.

\textbf{Dongjun Lee} contributed to the construction of the fine-tuning pipeline and the modeling of \sts{}.

\textbf{Hyunwoo Kim} contributed to the dataset construction of \mrc{}, and took part in preparing IRB.

\textbf{Myeonghwa Lee} contributed to the dataset construction of \sts{} and \tc.

\textbf{Seongbo Jang} contributed to the dataset construction of \re{}, and took part in preparing IRB.

\textbf{Seungwon Do} contributed to the dataset construction of \dst{} and text collection.

\textbf{Sunkyoung Kim} contributed to the dataset construction of \re{} and \mrc{}, and modeling of \mrc{}.

\textbf{Kyungtae Lim} contributed to the dataset construction of \posdp{}.

\textbf{Jongwon Lee} contributed to the dataset construction and modeling of \dst{}. 

\textbf{Kyumin Park} contributed to the dataset construction of \dst{}, and took part in preparing IRB.

\textbf{Jamin Shin} contributed to the dataset construction of \dst{}.

\textbf{Seonghyun Kim} contributed to the dataset construction and modeling of \ner.

\textbf{Lucy Park} contributed to the dataset construction of \mrc{} and provided the original corpus of \nsmc{}.

\textbf{Alice Oh} advised the project, sponsored the project via KAIST, provided feedback and suggested better way to improve the quality of our dataset and models, and helped with the final manuscript.

\textbf{Jung-Woo Ha} advised the project, sponsored annotation cost and computing cloud credits the project via NAVER, provided license-free news articles from \hankyung{}, and helped with the final manuscript. 

\textbf{Kyunghyun Cho} advised the project, provided critical feedback, suggested better way to improve the quality of our dataset and models, and helped a lot with polishing and rewriting the final manuscript. 

All participants contributed to this manuscript.

\clearpage
\appendix
\section{Dev Set Results}
\label{appendix:dev_set_results}

\begin{table}[ht!]
\centering
\caption{Performances of our pretrained LMs and other baselines on KLUE benchmark dev set. The notations are same with Table \ref{table:main_results}.}
\begin{adjustbox}{width=1\textwidth}
\begin{tabular}{l c cc c cc cc cc cc cc}
\toprule
& \textbf{\ynat{}} & \multicolumn{2}{c}{\textbf{\kluests{}}} & \textbf{\kluenli{}} & \multicolumn{2}{c}{\textbf{\kluener{}}} & \multicolumn{2}{c}{\textbf{\kluere{}}} & \multicolumn{2}{c}{\textbf{\klueposdp{}}} & \multicolumn{2}{c}{\textbf{\kluemrc{}}} & \multicolumn{2}{c}{\textbf{\wizard{}}} \\ \cmidrule(lr){2-2} \cmidrule(lr){3-4} \cmidrule(lr){5-5} \cmidrule(lr){6-7} \cmidrule(lr){8-9} \cmidrule(lr){10-11} \cmidrule(lr){12-13} \cmidrule(lr){14-15}
        \textbf{Model} & F1 & R$^P$ & F1 & ACC & F1$^E$ & F1$^C$ & F1$^{mic}$ & AUC & UAS & LAS & EM & ROUGE & JGA & F1$^S$ \\ \midrule
        \textbf{\mbertbase{}} & 82.64 & 82.97 & 75.93 & 72.90 & 75.56 & 88.81 & 58.39 & 56.41 & 88.53 & 86.04 & 49.96 & 55.57 & 35.27 & 88.60 \\
        \textbf{\xlmrbase{}} & 84.52 & 88.88 & 81.20 & 78.23 & 80.48 & 92.14 & 57.62 & 57.05 & 93.12 & 87.23 & 26.76 & 53.36 & 41.54 & 89.81 \\
        \textbf{\xlmrlarge{}} & \textbf{87.30} & 93.08 & \textbf{87.17} & 86.40 & 82.18 & \textbf{93.20} & 58.75 & 63.53 & 92.87 & 87.82 & 35.23 & 66.55 & 42.44 & 89.88 \\ \midrule
        \textbf{\krbertbase{}} & 85.36 & 87.50 & 77.92 & 77.10 & 74.97 & 90.46 & 62.83 & 65.42 & 92.87 & 87.13 & 48.95 & 58.38 & 45.60 & 90.82 \\
        \textbf{\koelectrabase{}} & 85.99 & \underline{93.14} & 85.89 & \underline{86.87} & \underline{\textbf{86.06}} & \underline{92.75} & 62.67 & 57.46 & 90.93 & 87.07 & 59.54 & 65.64 & 39.83 & 88.91 \\ \midrule
        \textbf{\bertbase{}} & \underline{86.95} & 91.01 & 83.44 & 79.87 & 83.71 & 91.17 & 65.58 & \underline{68.11} & 93.07 & 87.25 & 62.42 & 68.15 & 46.72 & 91.59 \\
        \textbf{\robertasmall{}} & 85.95 & 91.70 & 85.42 & 81.00 & 83.55 & 91.20 & 61.26 & 60.89 & 93.47 & 87.50 & 58.28 & 63.56 & 46.65 & 91.50 \\
        \textbf{\robertabase{}} & 86.19 & 92.91 & \underline{86.78} & 86.30 & 83.81 & 91.09 & \underline{66.73} & \underline{68.11} & \underline{93.75} & \underline{87.77} & \underline{69.56} & \underline{74.64} & \underline{47.41} & \underline{91.60} \\
        \textbf{\robertalarge{}} & 85.88 & \textbf{93.20} & 86.13 & \textbf{89.50} & 84.54 & 91.45 & \textbf{71.06} & \textbf{73.33} & \textbf{93.84} & \textbf{87.93} & \textbf{75.26} & \textbf{80.30} & \textbf{49.39} & \textbf{92.19} \\ \bottomrule
\end{tabular}
\end{adjustbox}
\label{table:dev_results}
\end{table}

In order to prevent early saturation of KLUE benchmark, we limit users to submit their models once per day. We thus present the dev set results to provide a reference for future work and local tests, in Table \ref{table:dev_results}. Models we used are same as in test set.

\end{document}